%% file: main.tex
\documentclass{article}

\usepackage[nonatbib,preprint]{neurips_2024}
\usepackage[subpreambles=true]{standalone}
\usepackage{import}

\usepackage[utf8]{inputenc} % allow utf-8 input
\usepackage[T1]{fontenc}    % use 8-bit T1 fonts
\usepackage{hyperref}       % hyperlinks
\usepackage{url}            % simple URL typesetting
\usepackage{booktabs}       % professional-quality tables
\usepackage{amsfonts}       % blackboard math symbols
\usepackage{nicefrac}       % compact symbols for 1/2, etc.
\usepackage{microtype}      % microtypography
\usepackage{xcolor}         % colors
\usepackage{amsmath}
\usepackage{arydshln}
\usepackage{blkarray}
\usepackage{graphicx}
\usepackage{multirow}
\usepackage{mathtools}
\def\block(#1,#2)#3{\multicolumn{#2}{c}{\multirow{#1}{*}{$ #3 $}}}

\input{env.tex}

\usepackage{amssymb,bm,amsthm}
\input{math_commands.tex}

\usepackage{floatrow}
\usepackage{multirow}
\usepackage{array}
\usepackage{color}
\usepackage{colortbl}
\usepackage{wrapfig}

\usepackage{bbding}
\usepackage[utf8]{inputenc} % allow utf-8 input
\usepackage[T1]{fontenc}    % use 8-bit T1 fonts
\usepackage{hyperref}       % hyperlinks
\usepackage{url}            % simple URL typesetting
\usepackage{booktabs}       % professional-quality tables
\usepackage{amsfonts}       % blackboard math symbols
\usepackage{nicefrac}       % compact symbols for 1/2, etc.
\usepackage{xcolor}         % colors
\usepackage{microtype}

\usepackage{caption}
\usepackage{subcaption}
\usepackage{floatrow}
\usepackage[subpreambles=true]{standalone}
\usepackage{import}
\usepackage{algorithm}
\usepackage{algpseudocode}

\usepackage{nccmath}

\newcommand{\mathbbm}[1]{\text{\usefont{U}{bbm}{m}{n}#1}}
\usepackage[normalem]{ulem}

\usepackage{enumitem}
\newlist{Properties}{enumerate}{2}
\setlist[Properties]{label=Property \arabic*., font=\textbf, itemindent=*}

\usepackage{amsmath}
\usepackage{amssymb}
\usepackage{mathtools}
\usepackage{amsthm}

\usepackage[capitalize,noabbrev]{cleveref}

\newtheorem{dfn}{Definition}

\let\b\boldsymbol

\title{What Is Missing In Homophily? Disentangling Graph Homophily For Graph Neural Networks}

\author{%
  Yilun Zheng \\
  Centre for Info. Sciences and Systems\\
  Nanyang Technological University\\
  Singapore 639798 \\
  \texttt{yilun001@e.ntu.edu.sg} \\
  \And
  Sitao Luan \\
  McGill University \\
  Mila - Quebec Artificial Intelligence Institute \\
  6666 Rue Saint-Urbain, Montréal \\
  \texttt{sitao.luan@mail.mcgill.ca} \\
  \And
  Lihui Chen \\
  Centre for Info. Sciences and Systems\\
  Nanyang Technological University\\
  Singapore 639798 \\
  \texttt{elhchen@ntu.edu.sg} \\
}

\begin{document}

\maketitle
\vspace{-0.5cm}

\begin{abstract}
\vspace{-0.3cm}
% Graph hom def, label aspect cannot align well; original not inferior; redefine from three aspects; CSBM-3H; theoretical results; synthetic&real-world datasets

Graph homophily refers to the phenomenon that connected nodes tend to share similar characteristics. Understanding this concept and its related metrics is crucial for designing effective Graph Neural Networks (GNNs). The most widely used homophily metrics, such as edge or node homophily, quantify such "similarity" as label consistency across the graph topology. These metrics are believed to be able to reflect the performance of GNNs, especially on node-level tasks. However, many recent studies have empirically demonstrated that the performance of GNNs does not always align with homophily metrics, and how homophily influences GNNs still remains unclear and controversial. Then, a crucial question arises: What is missing in our current understanding of homophily? To figure out the missing part, in this paper, we disentangle the graph homophily into three aspects: label, structural, and feature homophily, which are derived from the three basic element of graph data. We argue that the synergy of the three homophily can provide a more comprehensive understanding of GNN performance. Our new proposed structural and feature homophily consider the neighborhood consistency and feature dependencies among nodes, addressing the previously overlooked structural and feature aspects in graph homophily. To investigate their synergy, we propose a Contextual Stochastic Block Model with three types of Homophily (CSBM-3H), where the topology and feature generation are controlled by the three metrics. Based on the theoretical analysis of CSBM-3H, we derive a new composite metric, named Tri-Hom, that considers all three aspects and overcomes the limitations of conventional homophily metrics. The theoretical conclusions and the effectiveness of Tri-Hom have been verified through synthetic experiments on CSBM-3H. In addition, we conduct experiments on $31$ real-world benchmark datasets and calculate the correlations between homophily metrics and model performance. Tri-Hom has significantly higher correlation values than $17$ existing metrics that only focus on a single homophily aspect, demonstrating its superiority and the importance of homophily synergy. Our code is available at \url{https://github.com/zylMozart/Disentangle_GraphHom}.

\end{abstract}
\vspace{-0.5cm}
% \sitao{Based on the main content of this paper, I think the right question is " What is missing in graph homophily?" instead of "Should we completely discard the conventional definition of graph homophily?". And we answer this question in this paper by studying the 3 types of homophily derived from the 3 basic elements of graph data.}
% \yilun{Actually, previous metrics touched label, structural, feature aspect, and only label homophily is widely used and criticized. I feel "if graph homophily is bad?" is fine from this perspective.}

\section{Introduction}\vspace{-0.3cm}
\import{sections}{Introduction}\vspace{-0.3cm}

\section{Preliminary}\label{sec:preliminary}\vspace{-0.3cm}
\import{sections}{Preliminary}\vspace{-0.3cm}

\section{Disentangled Graph Homophily}\vspace{-0.3cm}
\import{sections}{GraphHomDef}
\vspace{-0.1cm}

\section{Impact of Disentangled Graph Homophily}\vspace{-0.3cm}
\import{sections}{GraphHomImpact}\vspace{-0.3cm}
% \vspace{-0.3cm}

\section{Experimental Results}\vspace{-0.3cm}
\import{sections}{Experiments}\vspace{-0.3cm}
\label{sec:experiments}
\section{Conclusions and Future Works}\label{sec:conclusion}\vspace{-0.3cm}
\import{sections}{Conclusion}\vspace{-0.3cm}

% \section{Limitation and Future Work}
% \import{sections}{LimitationFuture}

\newpage
\bibliography{references.bib}
\bibliographystyle{abbrv}

\newpage
\appendix
\import{sections}{Appendix}
%%%%%%%%%%%%%%%%%%%%%%%%%%%%%%%%%%%%%%%%%%%%%%%%%%%%%%%%%%%%

\newpage
\section*{NeurIPS Paper Checklist}

\begin{enumerate}

\item {\bf Claims}
    \item[] Question: Do the main claims made in the abstract and introduction accurately reflect the paper's contributions and scope?
    \item[] Answer: \answerYes{} % Replace by \answerYes{}, \answerNo{}, or \answerNA{}.
    \item[] Justification: We accurately answered the claims and contributions in the abstract and introduction, including why we shouldn't discard label homophily and the superiority of disentangled homophily.

\item {\bf Limitations}
    \item[] Question: Does the paper discuss the limitations of the work performed by the authors?
    \item[] Answer: \answerYes{} % Replace by \answerYes{}, \answerNo{}, or \answerNA{}.
    \item[] Justification: We discussed the limitations in Section \ref{sec:conclusion}

\item {\bf Theory Assumptions and Proofs}
    \item[] Question: For each theoretical result, does the paper provide the full set of assumptions and a complete (and correct) proof?
    \item[] Answer: \answerYes{} % Replace by \answerYes{}, \answerNo{}, or \answerNA{}.
    \item[] Justification: We provided the necessary assumptions for our proofs.

    \item {\bf Experimental Result Reproducibility}
    \item[] Question: Does the paper fully disclose all the information needed to reproduce the main experimental results of the paper to the extent that it affects the main claims and/or conclusions of the paper (regardless of whether the code and data are provided or not)?
    \item[] Answer: \answerYes{} % Replace by \answerYes{}, \answerNo{}, or \answerNA{}.
    \item[] Justification: We provide the reproducible code of our experiments, including node classification and homophily metrics.

\item {\bf Open access to data and code}
    \item[] Question: Does the paper provide open access to the data and code, with sufficient instructions to faithfully reproduce the main experimental results, as described in supplemental material?
    \item[] Answer: \answerYes{} % Replace by \answerYes{}, \answerNo{}, or \answerNA{}.
    \item[] Justification: All the datasets used in this paper are public. Our provided codes well reproduce the main experimental results.

\item {\bf Experimental Setting/Details}
    \item[] Question: Does the paper specify all the training and test details (e.g., data splits, hyperparameters, how they were chosen, type of optimizer, etc.) necessary to understand the results?
    \item[] Answer: \answerYes{} % Replace by \answerYes{}, \answerNo{}, or \answerNA{}.
    \item[] Justification: The experimental settings are shown in Appendix \ref{apd:exp_setting}.

\item {\bf Experiment Statistical Significance}
    \item[] Question: Does the paper report error bars suitably and correctly defined or other appropriate information about the statistical significance of the experiments?
    \item[] Answer: \answerYes{} % Replace by \answerYes{}, \answerNo{}, or \answerNA{}.
    \item[] Justification: We report the accuracy and standard deviation of the node classification performance with 10 runs with random splits. We report the p-value for the Pearson correlation.

\item {\bf Experiments Compute Resources}
    \item[] Question: For each experiment, does the paper provide sufficient information on the computer resources (type of compute workers, memory, time of execution) needed to reproduce the experiments?
    \item[] Answer: \answerNo{} % Replace by \answerYes{}, \answerNo{}, or \answerNA{}.
    \item[] Justification: We only provide computer resources used in our experiments. We didn't report the memory and time of execution because the main focus of this paper is to disentangle graph homophily instead of proposing a new method.
    
\item {\bf Code Of Ethics}
    \item[] Question: Does the research conducted in the paper conform, in every respect, with the NeurIPS Code of Ethics \url{https://neurips.cc/public/EthicsGuidelines}?
    \item[] Answer: \answerYes{} % Replace by \answerYes{}, \answerNo{}, or \answerNA{}.
    \item[] Justification: All the data used in this paper is public and the research doesn't violate the NeurIPS Code of Ethics.

\item {\bf Broader Impacts}
    \item[] Question: Does the paper discuss both potential positive societal impacts and negative societal impacts of the work performed?
    \item[] Answer: \answerNA{} % Replace by \answerYes{}, \answerNo{}, or \answerNA{}.
    \item[] Justification: This paper focuses on improving homophily measurement, which is irrelevant to societal impacts.
    
\item {\bf Safeguards}
    \item[] Question: Does the paper describe safeguards that have been put in place for responsible release of data or models that have a high risk for misuse (e.g., pretrained language models, image generators, or scraped datasets)?
    \item[] Answer: \answerNA{} % Replace by \answerYes{}, \answerNo{}, or \answerNA{}.
    \item[] Justification: This paper focuses on disentangling graph homophily, which is more conceptual and has no such risks.

\item {\bf Licenses for existing assets}
    \item[] Question: Are the creators or original owners of assets (e.g., code, data, models), used in the paper, properly credited and are the license and terms of use explicitly mentioned and properly respected?
    \item[] Answer: \answerYes{} % Replace by \answerYes{}, \answerNo{}, or \answerNA{}.
    \item[] Justification: We provided the necessary references for all the data used in this paper.

\item {\bf New Assets}
    \item[] Question: Are new assets introduced in the paper well documented and is the documentation provided alongside the assets?
    \item[] Answer: \answerNA{} % Replace by \answerYes{}, \answerNo{}, or \answerNA{}.
    \item[] Justification: We didn't release new assets

\item {\bf Crowdsourcing and Research with Human Subjects}
    \item[] Question: For crowdsourcing experiments and research with human subjects, does the paper include the full text of instructions given to participants and screenshots, if applicable, as well as details about compensation (if any)? 
    \item[] Answer: \answerNA{} % Replace by \answerYes{}, \answerNo{}, or \answerNA{}.
    \item[] Justification: The paper doesn't involve crowdsourcing nor research with human subjects.

\item {\bf Institutional Review Board (IRB) Approvals or Equivalent for Research with Human Subjects}
    \item[] Question: Does the paper describe potential risks incurred by study participants, whether such risks were disclosed to the subjects, and whether Institutional Review Board (IRB) approvals (or an equivalent approval/review based on the requirements of your country or institution) were obtained?
    \item[] Answer: \answerNA{} % Replace by \answerYes{}, \answerNo{}, or \answerNA{}.
    \item[] Justification: The paper doesn't involve crowdsourcing nor research with human subjects.

\end{enumerate}

\end{document}

%% file: env.tex
\newcommand\etal{\textit{et al.}}
\newcommand\ie{\textit{i.e.,}}
\newcommand\eg{\textit{e.g.,}}

\newcommand\iid{\textit{i.i.d.}}

\newcommand{\norm}[1]{\left\lVert#1\right\rVert}

\newcommand{\beq}{\begin{equation}}
\newcommand{\eeq}{\end{equation}}
\newcommand{\beqnn}{\begin{equation*}}
\newcommand{\eeqnn}{\end{equation*}}
\newcommand{\beqy}{\begin{eqnarray}}
\newcommand{\eeqy}{\end{eqnarray}}
\newcommand{\beqynn}{\begin{eqnarray*}}
\newcommand{\eeqynn}{\end{eqnarray*}}
\newcommand{\bit}{\begin{itemize}}
\newcommand{\eit}{\end{itemize}}
\newcommand{\ben}{\begin{enumerate}}
\newcommand{\een}{\end{enumerate}}
\newcommand{\bex}{\begin{example}}
\newcommand{\eex}{\end{example}}

\newcommand{\balg}[1]{\begin{algorithm} \caption{#1}}
\newcommand{\ealg}{\end{algorithm}}

\newcommand{\balgc}{\begin{algorithmic}[1]}
\newcommand{\ealgc}{\end{algorithmic}}

\newcommand{\bary}{\begin{array}}
\newcommand{\eary}{\end{array}}
\newcommand{\bmx}{\begin{bmatrix}}
\newcommand{\emx}{\end{bmatrix}}
\newcommand{\bsmx}{\left[\begin{smallmatrix}}
\newcommand{\esmx}{\end{smallmatrix}\right]}
\newcommand{\bmxc}[1]{\left[\begin{array}{@{}#1@{}}}
\newcommand{\emxc}{\end{array}\right]}
\newcommand{\bcn}{\begin{center}}
\newcommand{\ecn}{\end{center}}

\newcommand{\X}{{\mathbf{X}}}

\renewcommand{\b}{\mathbf{b}}

\providecommand{\norm}[1]{\lVert#1\rVert}

\providecommand{\abs}[1]{\left| #1 \right|}

\newenvironment{theorem}[2][Theorem]{\begin{trivlist}
		\item[\hskip \labelsep {\bfseries #1}\hskip \labelsep {\bfseries #2.}]}{\end{trivlist}}

%% file: math_commands.tex
\usepackage{amsmath,amsfonts,bm}

\def\eqref#1{equation~\ref{#1}}
\def\1{\bm{1}}

\DeclareMathAlphabet{\mathsfit}{\encodingdefault}{\sfdefault}{m}{sl}
\SetMathAlphabet{\mathsfit}{bold}{\encodingdefault}{\sfdefault}{bx}{n}

\DeclareMathOperator*{\argmin}{arg\,min}

%% file: sections/Introduction.tex
% Graph homophily is crucial in GNNs,Conventional homophily metrics related with GNN performance
% But inconsistency of Conventional metrics , question: abandon ?
% No! we need to Disentangle homophily: label, structural, feature aspects
% Theoretical&Emperical Analysis
% Contributions

Graph Neural Networks (GNNs) have been widely used in processing non-Euclidean data due to their superiority in extracting topological relations~\cite{defferrard2016fast, gilmer2017neural, GCN, GAT, GraphSage, luan2019break, lu2024representation}. They have achieved great success on numerous real-world tasks, \eg{} node classification~\cite{GCN}, link prediction~\cite{zhang2018link}, and graph clustering~\cite{tsitsulin2023graph}. It is found that their success, especially on node-level tasks, is closely related to the homophily assumption~\cite{hom_assump, def_hom_edge, ACM_GCN, luan2022complete}, \ie{} similar nodes tend to be connected~\cite{hom_conn_sim}. On the other hand, when dissimilar nodes are more likely to be connected, which is known as the non-homophily/heterophily scenario, GNNs fail to capture the useful neighbor information and even underperform Multilayer perceptrons (MLPs)~\cite{luan2023we}. Several homophily metrics, such as edge homophily~\cite{def_hom_edge,mixhop} and node homophily~\cite{Geom-GNN} were proposed, which were believed to be able to measure the performance of GNNs~\cite{when_do_graph_help} and recognize the difficult datasets~\cite{luan2022complete}.

However, recent studies~\cite{is_hom_necessary,when_do_graph_help,ana_label_info,ACM_GCN} show that the conventional homophily metrics~\cite{def_hom_edge,mixhop,Geom-GNN} are insufficient to measure the performance of GNNs: Luan \etal~\cite{ACM_GCN} show that the homophily metrics cannot tell if GNNs work well under heterophily. Ma \etal~\cite{is_hom_necessary} reveals that homophily is not a necessary assumption for effective GNNs and they propose to identify "good" and "bad" heterophily to explain why GNNs still work well under heterophily. Luan \etal~\cite{when_do_graph_help} discovers mid-homophily pitfall, showing the performance of GNNs reaches the worst in a medium level of homophily instead of the lowest. Then, a crucial question arises based on the above studies: What is missing in our current understanding of homophily?

In this paper, we fill the missing parts by investigating different perspectives of the "node similarity". Conventional homophily metrics quantify the "similarity" as an indicator function of whether connected nodes share the same label while ignoring the co-existence of three basic elements in graph data: label, structural, and node feature information. The ignorance of structural and feature information leads to insufficient understanding and unsatisfactory alignment between homophily metrics and GNN performance.

A complete understanding of graph homophily should include all the above three basic elements. To this end, we disentangle graph homophily into three corresponding aspects: label, structural, and feature homophily. Specifically, our new proposed structural homophily quantifies the "similarity" by considering the neighborhood structure consistency, and feature homophily measures the dependencies of node features across the topology. To investigate how their synergy affects GNN performance, we propose a Stochastic Block Model controlled with three types of Homophily (CSBM-3H). The node feature generation process in CSBM-3H breaks the \iid assumption in previous studies~\cite{is_hom_necessary,when_do_graph_help}, which is closer to real-world scenarios~\cite{feat_dpd_social_1,feat_dpd_social_2, feat_dpd_recm_2}. With the three metrics, CSBM-3H enables a more comprehensive study on the impact of graph homophily than previous analysis~\cite{is_hom_necessary, when_do_graph_help, ana_feat_dist, ana_nei_noise}.

%Compared with previous analysis of the impact of graph homophily\cite{is_hom_necessary, when_do_graph_help, ana_feat_dist, ana_nei_noise}, CSBM-3H enables us to study the impact of graph homophily on GNNs in a more complete view.
%Specifically, the graph topology generation is controlled by label and structural homophily, and the node feature generation is controlled by feature homophily. Compared with other random graph models used in previous studies~\cite{is_hom_necessary,when_do_graph_help}, CSBM-3H extends the graph topology generation by incorporating both the label and structural homophily and refines the node feature generation by introducing feature homophily.

From the theoretical study of CSBM-3H, we derive a new composite metric named Tri-Hom to measure the synergy, which includes all the three homophily aspects. Through CSBM-3H, our theoretical analysis and simulation results both show that the performance of GNNs is highly influenced by Tri-Hom. It can help explain how the three types of homophily influence GNN behavior individually or collectively. In addition, Our theoretical findings can explain some interesting phenomena observed in previous literature, such as "good" or "bad" heterophily~\cite{is_hom_necessary,ACM_GCN} and the impact of feature shuffling on GNNs~\cite{ana_feat_dist}. To verify the effectiveness of Tri-Hom, we conduct experiments on $31$ real-world datasets. The results show that GNN performance is significantly better aligned with Tri-Hom than the other $17$ existing metrics that focus only on a single homophily aspect. This implies that Tri-Hom can complete the absent parts in existing homophily metrics.

%% file: sections/Preliminary.tex
% Definition of Graph Basics, Labels, Neighbors, Features
We denote $\mathcal{G} = (\mathcal{V},\mathcal{E})$ as an undirected graph, where $\mathcal{V}$ is the node set and $\mathcal{E}$ is the edge set. The graph has $N$ nodes with $C$ classes. The adjacency matrix of the graph is denoted as $\b{A}\in \mathbb{R}^{N\times N}$. We use $\b{A}_{uv}=1$ or $e_{uv}\in \mathcal{E}$ to denote the existence of an edge between node $u$ and $v$, otherwise $\b{A}_{uv}=0$ or $e_{uv} \notin \mathcal{E}$. Node degree vector is denoted as $\b{D}\in\mathbb{R}^N$ where $\b{D}_u$ is the degree of node $u$. Node label vector is denoted as $\b{Y}\in\mathbb{R}^{N}$ and its one-hot encoding matrix is $\b{Z}\in \mathbb{R}^{N\times C}$. The number of nodes in class $c$ is denoted as $N_c=\abs{\{u|\b{Y}_{u}=c,u\in\mathcal{V}\}}$. The neighbor set of node $u$ is denoted as $\mathcal{N}_u = \{v|e_{uv}\in\mathcal{E}\}$. The features of all the nodes is denoted as $\b{X}\in\mathbb{R}^{N\times M}$, where $\b{X}_{v,:}$ are the features of node $v$ with $M$ dimensions. We use $\b{I}_E\in\mathbb{R}^{E}$ and $\b{1}_E\in\mathbb{R}^{E\times E}$ to denote identify matrix and all-ones matrix with size $E$, respectively.
%The $\b{A}$ describes one-hop neighbors and we get derive k-hop neighbors with $\b{A}^k$.

\textbf{Graph homophily metrics} are used to measure the similarity between connected nodes. Edge~\cite{mixhop,def_hom_edge} and node homophily~\cite{Geom-GNN} are $2$ most commonly used metrics and are defined as follows,
\begin{equation}
\begin{aligned}
\label{eq:definition_homophily_metrics}
& \resizebox{0.9\hsize}{!}{$h_{\text{edge}}(\mathcal{G},\b{Y}) = \frac{\big|\{e_{uv} \mid e_{uv}\in \mathcal{E}, Y_{u}=Y_{v}\}\big|}{|\mathcal{E}|}, \ 
h_{\text{node}}(\mathcal{G},\b{Y}) = \frac{1}{|\mathcal{V}|} \sum_{v \in \mathcal{V}}\frac{\big|\{u \mid u \in \mathcal{N}_v, Y_{u}=Y_{v}\}\big|}{\big|\mathcal{N}_v\bigl|}$} 
\end{aligned}
\end{equation}
These metrics qualify the ratio of whether the labels of two connected nodes are the same in a graph. However, this definition of graph homophily only considers a label aspect and neglects structural and feature aspects, resulting in a partial understanding of graph homophily. Therefore, we propose to disentangle the graph homophily as label, structural, and feature homophily in the next section.

\textbf{Graph-aware models $\mathcal{M}^{\mathcal{G}}$ and graph-agnostic models $\mathcal{M}^{\neg\mathcal{G}}$} refer to the models that either utilize structure information or do not, respectively. For example, baseline graph-aware models $\mathcal{M}^{\mathcal{G}}$, such as Graph Convolutional Network (GCN)~\cite{GCN}, Graph Attention Network (GAT)~\cite{GAT} and GraphSage~\cite{GraphSage}, encode both graph structure and node feature information in each layer; the corresponding graph-agnostic models $\mathcal{M}^{\neg\mathcal{G}}$ are the Multilayer Perceptrons (MLPs), which only encode node features~\cite{when_do_graph_help}. 

\textbf{Structural-agnostic features} refer to the node features $\b{X}$ that are conditionally independent of graph topology $\b{A}$ given $\b{Y}$, \ie{$(\b{X}\!\perp\!\!\!\perp\b{A}|\b{Y})$}; \textbf{structural-aware features} indicate $(\b{X}\not\!\perp\!\!\!\perp\b{A}|\b{Y})$.

%% file: sections/GraphHomDef.tex
% Conventional metrics of graph homophily only reflect a partial understanding of the GNN performance, therefore, we disentangle the graph homophily into three aspects label, structural, and feature homophily in this section. 
% \sitao{In this section, we are going to answer the previous question, \ie{} in which way the conventional homophily is still useful and in which part it is insufficient. And then we provide a more comprehensive and broader view to see this problem.}
In this section, we first introduce the definition of disentangled graph homophily from label, structural, and feature aspects to complete the missing part of the graph homophily. Then, in the next section, we will introduce how they collectively impact the performance of GNNs.

\vspace{-0.2cm}
\subsection{Label Homophily}
\vspace{-0.2cm}
\begin{dfn}
    \textbf{Label homophily} is defined as the consistency of node labels across the topology.
\end{dfn}
\vspace{-0.2cm}
Label homophily is the most widely used conventional metric of graph homophily and it qualifies the similarity between connected nodes $u$ and $v$ using an indicator function $\mathbbm{1}(Y_u=Y_v)$. Most of the conventional homophily metrics focus on label homophily, including edge homophily~\cite{mixhop, def_hom_edge}, node homophily~\cite{Geom-GNN}, class homophily~\cite{def_hom_class}, adjusted homophily~\cite{ana_label_info}, density-aware homophily~\cite{def_hom_den}, 2-hop neighbor class homophily~\cite{def_hom_2hop}, and neighbor homophily~\cite{def_hom_dom_nei}.

% Various forms of label homophily can be obtained by replacing the indicator function $\mathbbm{1}(\cdot)$~\cite{def_hom_den,def_hom_2hop,def_hom_dom_nei} or the grouping function $g(\cdot)$~\cite{def_hom_edge,Geom-GNN,def_hom_class}.

% Then we have label homophily for the whole graph on the edge set $g(\mathcal{E})$ grouped from different levels, which can be represented as
% \sitao{When you introduce or cite something in literature, you don't need to give detailed reason. For example, you don't need to explain why label homo is the most widely used one. That's not your job. When you introduce your proposed stuff, you need to give explanation. This is your job.}
% \begin{equation}
% h_L(\mathcal{G},\b{Y})=\frac{1}{\abs{g(\mathcal{E})}}\displaystyle\sum_{\mathcal{E}'\in g(\mathcal{E})} \frac{1}{\abs{\mathcal{E}'}}\displaystyle\sum_{(u,v) \in \mathcal{E}'} \mathbbm{1}(Y_u=Y_v)
% \end{equation}
% \sitao{Do you need this general form in your paper?}
%  \sitao{There is some redundancy and repetition with the previous section. You can put the introduction of the label-based homophily here.}

However, label homophily only focuses on the consistency of label information for connected nodes while neglecting structural and feature information, which are two indispensable components of graph data. Hence, it offers only a partial understanding of graph homophily, which cannot always align well with the performance of GNNs ~\cite{is_hom_necessary,ACM_GCN,when_do_graph_help}. To capture the missing structural and feature information and better understand graph homophily, we give the definitions of structural and feature homophily in the following $2$ subsections.

\vspace{-0.2cm}
\subsection{Structural homophily}\vspace{-0.2cm}
% The main difference of structure homophily with label homophily is, the "atom" information is the structure information 
% \sitao{You need to explain how you define the structural homophily. For example, "To capture structural similarity of nodes from the same class, we define the structural homophily as follows"}
For structural homophily, the "atom" information of a node is structural information instead of the label. It is meaningless to define the structural homophily using the consistency across the graph topology as in the label homophily because the structural information already contains the information from the graph topology. Therefore, we define the structural information as the consistency of structural information among the nodes from the same classes\footnote{There we do not consider the inter-class structural information because the structural homophily represents a property of a graph instead of the node distinguishability\cite{when_do_graph_help}. The detailed discussion of the node distinguishability is shown in Section \ref{sec:node_distinguish}.}, which better disentangles itself from the label homophily.

\begin{dfn}
    \textbf{Structural homophily} is defined as the consistency of structural information of nodes within the same class. The structural homophily in a graph is defined as:
    \vspace{-0.1cm}
    \begin{equation}\label{eq:def_structural_hom}
    \begin{aligned}
   h_{S}(\mathcal{G},\mathcal{S},\b{Y}) = \frac{1}{C}\sum_{c=1}^C h_{S,c}, \; \text{ where } h_{S,c}(\mathcal{G},\mathcal{S},\b{Y}) &= 1 - \frac{\sigma(\{\mathcal{S}(u)|u\in\mathcal{V},Y_{u}=c\})}{\sigma_{max}} \\
    \end{aligned}
    \end{equation}
    where $h_{S,c}$ is the class-wise structural homophily for class $c$, function $\mathcal{S}(\cdot)$ measures structural information, $\sigma$ denotes standard deviation of structural information, and $\sigma_{max}$ denotes the maximum value of $\sigma$.
\end{dfn}
  
In this paper, we quantify the structural information for node $u$ through neighbor distribution (the class distribution of local neighbors) $\b{D}^{\mathcal{N}}_u = [p_{u,1}, p_{u,2}, \ldots, p_{u,c}]$, where $p_{u,k} = \frac{|\{Y_{v}=k|v\in \mathcal{N}_u\}|}{|\mathcal{N}_u|}$ is the proportion of neighbors of node $u$ that belong to class $c$. A high structural homophily indicates that the graph-aware models leveraging structural encoding will have similar embeddings for intra-class nodes after aggregation, which are expected to outperform graph-agnostic models, irrespective of a low label homophily. There are also some homophily metrics that focus on the structural aspect in previous studies, including label informativeness~\cite{ana_label_info}, neighborhood similarity~\cite{is_hom_necessary}, and aggregation homophily~\cite{ACM_GCN}, which is similar as the structural homophily defined there. 

\vspace{-0.2cm}
\subsection{Feature Homophily}\label{sec:def_h_f}\vspace{-0.2cm}

Previous feature-based graph homophily metrics, such as generalized edge homophily~\cite{def_hom_GE}, local similarity~\cite{def_hom_localsim}, attribute homophily~\cite{def_hom_attr}, and class-controlled feature homophily~\cite{ana_feat_dist}, mainly focus on the consistency of node features across the graph topology, which is similar as the definition of Dirichlet energy in graphs. However, these homophily metrics on feature consistency cannot fully disentangle itself from label homophily: Since the features of nodes in a graph are supposed to depend on their classes, when the graph shows a high/low label homophily, the connected nodes are more likely to share the same/different labels, resulting in a high/low feature similarity. Therefore, these feature-based homophily metrics are dependent on label homophily. Such dependency contains redundancy, which decreases the useful information inside feature-based homophily and impedes our understanding of the relationship between node features and GNN performance.

To disentangle the feature effect from label and graph structure, we define the feature homophily as the dependencies of node features across the graph topology, thereby dissociating it from label homophily and structural homophily. Inspired by graph diffusion~\cite{GRAND} and interactive particle systems~\cite{diff_particle_sys, diff_particle_sys2}, we have the structural-agnostic unobserved feature $\b{X}(0)$ and the observed structural-aware feature $\b{X}$ that satisfy the following relation

\vspace{-0.3cm}
\begin{equation}\label{eq:node_feat_sampling}
    \b{X} = \Bigr[ \sum_{t=0}^\infty (\omega\b{A})^t \Bigr] \b{X}(0) = (\b{I}-\omega\b{A})^{-1}\b{X}(0)
\end{equation}
\vspace{-0.4cm}

The detailed process of this relation is given in Appendix \ref{apd:diff_particle}. Here $\omega \in (-\frac{1}{\rho(\b{A})},\frac{1}{\rho(\b{A})})$ is a parameter that controls the feature dependencies, where a positive, negative, or zero value corresponds to an attractive relation, repulsive relation, or independence of the nodes with their neighbors in graphs~\cite{diff_particle_sys, diff_particle_sys2}. The feature dependencies $(\omega\b{A})^t$ of $t$-order neighbors are introduced to structural-agnostic features $\b{X}(0)$. Finally, the state of all the nodes will converge to an equilibrium with structure-aware feature $\b{X}$. The $\omega$ in Eq. (\ref{eq:node_feat_sampling}) is independent of the graph topology because no matter how the label homophily or structural homophily changes, $\omega$ will remain unaffected. To disentangle feature homophily from label homophily and structural homophily, we define the feature homophily based on $\omega$ as follows.

% Compared with the previous feature sampling strategy~\cite{ana_nei_noise, ana_feat_dist, when_do_graph_help} where node features are independently sampled from their class-specific multivariate Gaussian distributions $p(\b{X}|\b{Y})$ \sitao{Is this feature modeling process related to CSBM modeling? If yes, you should put it in later section.}, our feature modelling further considers the feature dependencies among neighbors in graphs $p(\b{X}|\b{Y},\b{A})$, which are closer to real-world scenarios~\cite{feat_dpd_social_1,feat_dpd_social_2,feat_dpd_recm_2}. 

\begin{dfn}
    \textbf{Feature homophily} is defined as the degree of feature dependencies of nodes across the topology. For the linear case of the graph diffusion process with feature dependencies, the feature homophily for feature $m$ satisfies
    \begin{equation}\label{eq:feat_hom_def_implicit}
        \b{X}_{:,m} = \left(\b{I}-\frac{h_{F,m}}{\rho(\b{A})}\b{A}\right)^{-1}\b{X}_{:,m}(0)
    \end{equation}
    where $\rho(\b{A})$ is the spectral radius of $\b{A}$, $\b{X}(0)\sim p(\b{X}|\b{Y})$ are the unseen structural-agnostic node features, and $\b{X}\sim p(\b{X}|\b{Y},\b{A})$ is the observed structural-aware node features. The feature homophily for the whole graph is the averaged feature homophily for all the features
    % \sitao{Why did you put $h_{F,m}$ in the numerator? To disentangle the effect of label?}
    \vspace{-0.1cm}
    \begin{equation}
    h_F(\mathcal{G},\b{X},\b{Y}) = \frac{1}{M}\sum_{m=1}^M h_{F,m}
    \end{equation}
\end{dfn}
\vspace{-0.3cm}
% \sitao{This should be put in later section, after the synthetic part.}

% \sitao{Why do we use this criterion to solve $h_{F,m}$? Is this closer to reality? If yes, give references.}

\paragraph{Remark} It is easy to control the feature homophily in the generation of synthetic graphs. However, since both $h_{F,m}$ and $\b{X}_{:,m}(0)$ are unknown in Eq.~\ref{eq:feat_hom_def_implicit}, one more condition is required to estimate feature homophily in real-world datasets. To address this issue, we consider the case where the intra-class distances of $\b{X}_{:,m}(0)$ are small. This case holds in lots of real-world scenarios~\cite{ACM_GCN, when_do_graph_help} and we can utilize this property to estimate $h_{F,m}$ without solving $\b{X}_{:,m}(0)$. Specifically, the feature homophily $h_{F,m}$ for feature $m$ can be estimated with the following optimization process %():
\vspace{-0.1cm}
\begin{equation}
\label{eq:feature_homophily}
\begin{split}
 \resizebox{0.94\hsize}{!}{$h_{F,m}^{*}(\mathcal{G},\b{X}_{:,m},\b{Y}) = \argmin_{h_{F,m}} \sum_{\substack{u,v\in\mathcal{V},\\ Y_u=Y_v}} \norm{X_{u,m}(0)-X_{v,m}(0)}^2, \; \text{ where } \b{X}_{:,m}(0) = \left(\b{I}-\frac{h_{F,m}}{\rho(\b{A})}\b{A}\right)\b{X}_{:,m}$}
\end{split}
\end{equation}
\vspace{-0.3cm}
It will be used in Section~\ref{sec:experiments} for calculation.

%% file: sections/GraphHomImpact.tex
% \sitao{You should first state why you use a modified CSBM to study the effect, what is the advantages of proposed CSBM? See my paper\cite{when_do_graph_help} for reference.}

To study the model performance in a graph, the Contextual Stochastic Block Model (CSBM) has been widely used to study the performance of GNNs with controlled graph topology and node features. Previous studies~\cite{ana_label_info,ana_feat_dist,when_do_graph_help,is_hom_necessary} on graph homophily generally adopt a modified CSBM to control the label homophily through assigning nodes with different probabilities that connect to the nodes from other classes. Then the node features are sampled solely based on the classes. However, this graph modeling, that only considers label homophily, has two drawbacks: First, the probabilities of nodes from the same class connecting to the nodes with different classes are uniform, which lacks diversity. Second, the sampled node features are independent with their structures \ie{$(\b{X}\!\perp\!\!\!\perp\b{A}|\b{Y})$}, which is uncommon in real-world scenarios where interactions influence the attributes of connected nodes~\cite{feat_dpd_recm_2,feat_dpd_social_1,feat_dpd_social_2}. Therefore, we propose a Contextual Stochastic Block Model with three types of Homophily (CSBM-3H), a random graph generative model that integrates the three types of homophily (Section~\ref{sec:CSBM-3H}), where the newly proposed structural homophily $h_S$ and feature homophily $h_F$ can well address the aforementioned two drawbacks and fills the missing part of graph homophily. Then, Based on CSBM-3H, we theoretically study how the graph-agnostic and graph-aware models are affected by label, structural, and feature homophily metrics to explore their relationship and verify the effectiveness of proposed metrics (Section~\ref{sec:node_distinguish}).

% To investigate how three types of homophily influence the performance on graph-aware models $\mathcal{M^G}$ and graph-agnostic models $\mathcal{M}^{\neg\mathcal{G}}$, we introduce Contextual Stochastic Block Model with three types of Homophily (CSBM-3H), a random graph generative model that integrates the three types of homophily (Section~\ref{sec:CSBM-3H}). Based on CSBM-3H, we theoretically and numerically study how the graph-agnostic and graph-aware models are affected by label, structural and feature homophily metrics to explore their relationship and verify the effectiveness of proposed metrics (Section~\ref{sec:node_distinguish}).

\vspace{-0.2cm}
\subsection{CSBM-3H}\label{sec:CSBM-3H}\vspace{-0.2cm}

\paragraph{Graph Topology Generation} To be consistent with existing literature and without loss of generality~\cite{when_do_graph_help}, we assume all the nodes are class-balanced and share the same node degree $d$. We use node homophily $h_L$ to control the label consistency across the graph topology and $h_S$ to control the consistency of neighbor distribution $\b{D}^{\mathcal{N}}$ of nodes within the same classes. Then, the neighbor distribution can be expressed as:

\vspace{-0.3cm}
\begin{equation}\label{eq:sampling_hl_hs}
\begin{split}
    \b{D}^{\mathcal{N}} = \mathbb{E}_{\epsilon}[\b{Z}\b{S}], \ 
    \text{where}\ \b{S}
    % \begin{bmatrix}
    %     h_L & \frac{1-h_L}{c-1} & \cdots & \frac{1-h_L}{c-1} \\
    %     \frac{1-h_L}{c-1} & h_L & \cdots & \frac{1-h_L}{c-1} \\
    %     \vdots & \vdots & \ddots & \vdots \\
    %     \frac{1-h_L}{c-1} & \frac{1-h_L}{c-1} & \cdots & h_L \\
    % \end{bmatrix} + \b{\epsilon} 
    = \frac{1-h_L}{c-1}\b{1}_C + (h_L-\frac{1-h_L}{c-1})\b{I}_C + \b{\epsilon}
\end{split}
\end{equation}
\vspace{-0.3cm}

% each row of $\b{D}^{\mathcal{N}}$ satisfies the constraint $\sum_{c} {D_{\mathcal{N}}_{u,c}}=1$ and $D_{\mathcal{N}}_{u,c}\in [0,1] ,\ \forall u = 1,\dots , N, c = 1, \dots, C$;

where $\b{D}^{\mathcal{N}}\in\mathbb{R}^{N\times C}$ is the neighbor distribution for all the nodes, $\b{S}\in\mathbb{R}^{C\times C}$ is a class-sampling matrix, and $\b{\epsilon}\in\mathbb{R}^{C\times C}$ is a noise matrix. Each entry of $\b{\epsilon}$ is a noise of neighbor sampling that follows a Gaussian distribution $N(0,\frac{(1-h_S)^2}{c-1})$. The class-sampling matrix $\b{S}$ should be legal in practice~\cite{ana_nei_noise} \ie{} $\b{S}_{u,v}>0$ and $\sum_v \b{S}_{u,v}=1$. Then an adjacency matrix $\b{A}$ can be sampled from a neighborhood sampling matrix $\b{A_p} = \frac{Cd}{N} \b{D}^{\mathcal{N}} \b{Z}^T$, where $\mathbb{E}[A_{uv}=1]=(\b{A}_{p})_{uv}$ for each pair of nodes $u,v$. In this way, we control the label homophily $h_L$ and structural homophily $h_S$ in a graph.

%In this way, we can get the neighbor sampling matrix as $\b{A_p} = \frac{Cd}{N}\b{D}^{\mathcal{N}}\b{Y}^T$. Then, we can derive the final adjacency matrix $\b{A}$ by sampling $\b{A_p}$, where $\mathbb{E}[A_{uv}=1]=A_{p,uv}, \forall u,v \in \mathcal{V}$.

% \begin{equation}
%     \b{A} = ( Sym \circ Bernoulli ) (\b{A_p}) 
% \end{equation}
% \sitao{Such Compound probability distribution makes things complex. See \url{https://en.wikipedia.org/wiki/Compound_probability_distribution}}
% where the $bernoulli$ function converts each value into 1 or 0 with given probability and the $Sym$ function turns the non-symmetric matrix to symmetric matrix.

% As shown in Eq. (\ref{eq:sampling_hl_hs}), the label homophily $h_L$ controls the ratio of the neighbors that share the same labels as the ego nodes and the structural homophily $h_S$ controls the standard deviation of the neighbor distribution of each class. The $h_L$ and $h_S$ control different aspects of graph, enabling us to study the how the model performance will be affected by the topology.

\textbf{Node Feature Generation}. For any node $u$ in a graph, we first sample its structural-agnostic features $\b{X}_u (0)\in\mathbb{R}^{F}$ from a class-wised Gaussian distribution $\b{X}_u (0) \sim \b{N}_{Y_u}(\b{\mu}_{Y_u},\b{\Sigma}_{Y_u})$ with $\b{\mu}_{Y_u}\in\mathbb{R}^F$ and $\b{\Sigma}_{Y_u}\in\mathbb{R}^{F\times F}$. We also assume each dimension of feature vector is independent from each other, thereby $\b{\Sigma}_{Y_u}\in\mathbb{R}^{F\times F}$ is a diagonal matrix. Then the observed structural-aware features can be generated by the unseen structure-agnostic feature as described in Eq. (\ref{eq:feat_hom_def_implicit}).

\vspace{-0.2cm}
\subsection{Node distinguishability}\label{sec:node_distinguish}\vspace{-0.2cm}
Suppose we have the representations of node $u$ as $\b{H}_{u:} = \frac{1}{d}\sum_{v\in \mathcal{N}_u} \b{X}_{v:}$ for graph-aware models $\mathcal{M}^{\mathcal{G}}$ and  $\b{H}_{u:} = \b{X}_{u:}$ for graph-agnostic models $\mathcal{M}^{\neg\mathcal{G}}$. To quantify the impact of the aforementioned homophily metrics on both the graph-aware models $\mathcal{M}^{\mathcal{G}}$ and graph-agnostic models $\mathcal{M}^{\neg\mathcal{G}}$, we follow the node distinguishability of node embeddings $\b{H}$, which measures the ratio of intra-class node distance to inter-class distance~\cite{when_do_graph_help}. To ideally distinguish nodes from different classes, a smaller intra-class distance $D_\text{intra}(\b{H})$ and larger inter-class distance $D_\text{intra}(\b{H})$  are preferred, because this will reduce boundary nodes and increase the margins among classes. The metric is defined as follows,
%Therefore, the ratio of $D_\text{intra}(\b{H})$ to $D_\text{inter}(\b{H})$ as a metric to measure the node distinguishability of $\b{H}$: 

\vspace{-0.3cm}
\begin{equation}
    \begin{split}
    \mathcal{J} &= \frac{D_\text{intra}(\b{H})}
                        {D_\text{inter}(\b{H})} 
        = \frac {\mathbb{E}_{y_u=y_v,\epsilon}\left[\norm{\b{H}_u-\b{H}_v}^2\right]}
                {\mathbb{E}_{y_u\neq y_v,\epsilon}\left[\norm{\b{H}_u-\b{H}_v}^2\right]}
    \end{split}
\end{equation}
\vspace{-0.3cm}

% \yilun{how to present the variable in this expectation comfortably?}

A smaller $\mathcal{J}$ indicates better node embeddings for the model performance and vice versa, which has been proved in~\cite{when_do_graph_help}. With the proposed CSBM-3H, we can analyze the impacts of $h_L$, $h_S$, and $h_F$ on $\mathcal{M}^{\neg\mathcal{G}}$ and $\mathcal{M}^{\mathcal{G}}$ by studying their relations with $\mathcal{J}$, which will be derived in the following theorems.

\begin{theorem}{1}\label{theo:ratio}
    In CSBM-3H, the ratio of the expectation of intra-class distance to the expectation of inter-class distance of node representations for graph-agnostic models $\mathcal{M}^{\neg\mathcal{G}}$ and graph-aware models $\mathcal{M}^{\mathcal{G}}$ is:
    \begin{equation}
        \begin{split}
            \mathcal{J}^{\neg\mathcal{G}} = (1+\mathcal{J}_{\b{N}}\mathcal{J}_h^{\neg\mathcal{G}})^{-1}
            \;\text{and}\;
            \mathcal{J}^{\mathcal{G}} = (1+\mathcal{J}_{\b{N}}\mathcal{J}_h^{\mathcal{G}})^{-1}
        \end{split}
    \end{equation}
    where $\mathcal{J}_{\b{N}}= \frac{\sum_{Y_u\neq Y_v}[2C(C-1)]^{-1}\norm{\b{\mu}_{Y_u}-\b{\mu}_{Y_v}}^2}{C^{-1}\abs{\b{\sigma}^2}}$, $\mathcal{J}_h^{\neg\mathcal{G}}=\frac{1-(\frac{h_F}{\rho(\b{A})})^2(C(\frac{1-h_L}{C-1})^2+C\frac{(1-h_S)^2}{C-1}+(\frac{h_L C-1}{C-1})^2)}{\left[1- (\frac{h_F}{\rho(\b{A})}) (\frac{h_L C-1}{C-1})\right]^2}$, and $\mathcal{J}_h^{\mathcal{G}}=\frac{(\frac{h_L C-1}{C-1})^2}{C(\frac{1-h_L}{C-1})^2+C\frac{(1-h_S)^2}{C-1}+(\frac{h_L C-1}{C-1})^2}\mathcal{J}_h^{\neg\mathcal{G}}$. (See the proof in Appendix~\ref{apd:proof_theorem1}.)
\end{theorem}
From Theorem 1 we can see that, $\mathcal{J}_{\b{N}}$ is a normalized distance term which is a constant given the distribution of structural-agnostic features and is irrelevant with graph information; $\mathcal{J}_h^{\neg\mathcal{G}}$ and $\mathcal{J}_h^{\mathcal{G}}$ are controlled by the three types of homophily, which can reflect the influence of graph homophily on model performance. We name $\mathcal{J}_h^{\neg\mathcal{G}}$ and $\mathcal{J}_h^{\mathcal{G}}$ as \textbf{Tri-Hom} for $\mathcal{M}^{\neg\mathcal{G}}$ and $\mathcal{M}^{\mathcal{G}}$. To study the effect of Tri-Hom in more detail, we take the partial derivative of $\mathcal{J}_h^{\mathcal{G}}$ with respect to $h_S$, $h_F$, and $h_L$ to show the analytical results of their influences\footnote{We also calculate the partial derivative of $\mathcal{J}_h^{\neg\mathcal{G}}$ and discuss the impact of three types of homophily on $\mathcal{M}^{\neg\mathcal{G}}$ in Appendix~\ref{apd:hom_impact_mlp}.}. (See calculation in Appendix~\ref{apd:proof_theorem21}, \ref{apd:proof_theorem22}, and \ref{apd:proof_theorem23}.)

\begin{theorem} {2.1}
    The partial derivative of ${\mathcal{J}}_h^{\mathcal{G}}$ with respect to label homophily $h_L$ satisfies,
    \begin{equation}
        \begin{split}
        \begin{cases}
        \frac{\partial\mathcal{J}_h^{\mathcal{G}}}{\partial h_L}<0, & \text{if}\ h_L\in[0,\frac{1}{C})\\
        \frac{\partial\mathcal{J}_h^{\mathcal{G}}}{\partial h_L} \geq 0, & \text{if}\ h_L\in[\frac{1}{C},1]
        \end{cases}
        \end{split}
    \end{equation}
\end{theorem}
\vspace{-0.2cm}
From Theorem 2.1 we can see that, the worst performance of $\mathcal{M}^{\mathcal{G}}$ is reached when $h_L = \frac{1}{C}$, which corresponds to the scenario with the highest number of unpredictable boundary nodes among classes. This result is close to the previous finding with the worst $h_L = \frac{1}{C+1}$~\cite{ACM_GCN} and aligns with the phenomenon of mid-homophily pitfall in~\cite{when_do_graph_help}. 
\begin{theorem} {2.2}
    The partial derivative of ${\mathcal{J}}_h^{\mathcal{G}}$ with respect to structural homophily $h_S$ satisfies,
    \begin{equation}
        \begin{split}
            \frac{\partial\mathcal{J}_h^{\mathcal{G}}}{\partial h_S}\ge0
        \end{split}
    \end{equation}
\end{theorem}
\vspace{-0.2cm}

From Theorem 2.2 we can see that, a larger $h_S$ consistently improves the performance of $\mathcal{M}^{\mathcal{G}}$. This is intuitive for $\mathcal{M}^{\mathcal{G}}$ because a consistent intra-class neighbor distribution will lead to closer intra-class node representations after feature aggregation. This conclusion is also shown in Wang \etal~\cite{ana_nei_noise}, where the topological noise (which is inversely proportional to $h_S$) has a detrimental impact on separability.
\begin{theorem} {2.3}
    The partial derivative of ${\mathcal{J}}_h^{\mathcal{G}}$ with respect to feature homophily $h_F$ satisfies,
    \begin{equation}
        \begin{split}
            \begin{cases} 
            \frac{\partial\mathcal{J}_h^{\mathcal{G}}}{\partial h_F}< 0, & \text{if}\ h_L\in(0 ,h_L^-);\ h_L \in( h_L^-,h_L^+)\ \text{ and } \ h_F \in (\hat{h}_F,1)\\
            \frac{\partial\mathcal{J}_h^{\mathcal{G}}}{\partial h_F}> 0, & \text{if}\ h_L\in(h_L^+,1];\ h_L\in(h_L^-,h_L^+)\ \text{ and }\ h_F\in(-1,\hat{h}_F) \\
            \frac{\partial\mathcal{J}_h^{\mathcal{G}}}{\partial h_F} = 0, & \text{if}\ h_L = \frac{1}{C};\ h_L\in(h_L^-,h_L^+)\ \text{ and }\ h_F=\hat{h}_F
            \end{cases}
        \end{split}
    \end{equation}
    where $0<h_L^- < h_L^+<1$ and $-1<\hat{h}_F<1$. The expressions and detailed calculation of $h_L^-$, $h_L^+$, and $\hat{h}_F$ are shown in Appendix \ref{apd:proof_theorem23}.
\end{theorem}
\vspace{-0.2cm}

From Theorem 2.3 we can see that, when $h_L$ is high in a graph, nodes with the same labels tend to be connected, thereby a larger $h_F$ makes the intra-class nodes share more similar representations and positively affect $\mathcal{M^G}$; when $h_L$ is low in a graph, nodes with the different labels tend to be connected, thereby a larger $h_F$ makes the inter-class nodes share more similar representations and negatively affect $\mathcal{M^G}$; when $h_L$ is on a medium level \ie{} $h_L\in(h_L^-,h_L^+)$, an increase of $h_F$ will first improves and then reduces the performance of $\mathcal{M^G}$ with the cut-off point at $h_F=\hat{h}_F$. There the $\hat{h}_F$ is influenced by $h_L$, $h_S$, $C$, and $\rho(\b{A})$, and the $h_L^-$ and $h_L^+$ are influenced by $h_S$, $C$, and $\rho(\b{A})$.

\vspace{-0.2cm}
\paragraph{Remark} Apart from the new findings mentioned above, Tri-Hom can help explain other interesting but under-explored phenomena of the graph homophily in previous studies, \eg{} 1) "good" or "bad" heterophily~\cite{is_hom_necessary}, which states that GNN can still perform well in some heterophily cases; and 2) feature shuffling~\cite{ana_feat_dist}, which states that shuffling the node features randomly within the same class can improves the performance of GNNs on node classification. Our explanations with Tri-Hom are: 1) The occurrence of "good" or "bad" heterophily is due to that the model performance is influenced by a combination of $h_L$, $h_F$, and $h_S$, instead of $h_L$ alone. When the $h_L$ is low, the graph-aware models can still achieve good performance with a high $h_S$ or a low $h_F$; 2)  feature shuffling is due to the existence of the structural-aware features of nodes. When $h_F>0$, nodes are positively dependent on their neighbors. In this case, the nodes at the class boundaries or class centers are the hardest or easiest ones to predict because of the feature dependencies. If we randomly shuffle the nodes inside their classes, the nodes at the class boundaries will be easier to be classified because their features are more likely to be replaced by the nodes from the center, which are more distinguishable. For the nodes close to the class centers, it will be compensated by their neighbors. 
\vspace{0.3cm}
%Therefore, CSBM-3H explains the overall improvement of the accuracy of node classification under the feature shuffle. In conclusion, our proposed CSBM-3H and Tri-Hom can well explain these interesting phenomenon, which have more generality compared with previous analysis.

% \yilun{writing this by reasoning, without proof.}

% \yilun{add total derivative?}

% \begin{figure}
%     \centering
%     \includegraphics[width=1.0\linewidth]{imgs/theory/proof_2d.pdf}
%     \text{and}tion{Numerical results of $\mathcal{J}_h$ with respect to label homophily, feature homophily, and structural homophily.}
%     \label{fig:numerical_hom_obj}
% \end{figure}

% \begin{figure}
%     \centering
%     \includegraphics[width=1.0\linewidth]{imgs/experiment/all_acc_0426.pdf}
%     \text{and}tion{Model Performance with respect to label homophily, feature homophily, and structural homophily.}
%     \label{fig:syn_hom}
% \end{figure}

% \begin{figure} 
%     \centering
%   \subfloat[Numerical results of $\mathcal{J}_h$\label{fig:numerical_hom_obj}]{%
%        \includegraphics[width=0.5\textwidth]{imgs/theory/proof_2d.pdf}}
%     \hfill
%   \subfloat[Model performance on synthetic graphs\label{fig:syn_hom}]{%
%        \includegraphics[width=0.5\textwidth]{imgs/experiment/all_acc_0426.pdf}}
%     \hfill
%     \text{and}tion{aaaaaaa}
% \end{figure}

%% file: sections/Experiments.tex
In Section \ref{exp:synthetic}, we conduct experiments on synthetic data generated by CSBM-3H to verify the conclusions from Theorem 2.1, 2.2, and 2.3, demonstrate the synergy of label $h_L$, structural $h_S$, and feature homophily $h_F$ and test whether Tri-Hom $\mathcal{J}_h^{\mathcal{G}}$ can reflect GNN performance. In Section \ref{exp:realworld}, we evaluate the effectiveness of Tri-Hom on real-world benchmark datasets to test how well it can predict the model performance in real-world scenarios. In addition, we calculate the correlation between Tri-Hom value and model performance and compare them with the results of other $17$ existing metrics. The results show that Tri-Hom has a significantly higher correlation with the model performance, demonstrating the necessity of filling the missing part by disentangling graph homophily from three aspects.

\vspace{-0.2cm}
\subsection{Experiments on Synthetic Datasets}\label{exp:synthetic}
\vspace{-0.2cm}
To verify our theoretical results in a more general case, we measure the performance of GCN on synthetic datasets, where we can easily control $h_L$, $h_S$, and $h_F$. Specifically, 
\vspace{-0.3cm}
\paragraph{Synthetic Data Generation} We generate the synthetic graphs using CSBM-3H with a given tuple  $(h_L,h_S,h_F)$, where $h_L \in \{0, 0.1, \dots, 0.9, 1\}, \; h_S \in \{ 0, 0.1, \dots, 0.9, 1\}$ and $h_F\in \{-0.8, 0.6, \dots, 0.6, 0.8\}$. For each $(h_L,h_S,h_F)$, we generate $1000$ nodes with three balanced classes and the node degrees are sampled from a uniform distribution $[1,10]$. For node features, we first sample the structural-agnostic features from class-specific Gaussian distributions, then we use Eq. (\ref{eq:feat_hom_def_implicit}) to propagate these features across the graph topology with feature homophily $h_F$ to generate the observed structural-aware features as the final node features. Then, we evaluate the node classification performance of GCN~\cite{GCN} on these synthetic graphs. To get a robust evaluation and mitigate the numerical instability, we generate $10$ graphs with $10$ random seeds for each $(h_L,h_S,h_F)$ and report the average and standard deviation of classification accuracy on validation sets. The detailed process of topology and feature generation is shown in Appendix \ref{apd:dataset_syn}.

\vspace{-0.3cm}
\paragraph{Numerical Results of Tri-Hom} To verify whether Tri-Hom $\mathcal{J}_h^{\mathcal{G}}$ can reflect the behavior of GCN, we calculate its numerical results with the same setting as the synthetic graphs and make a comparison. Specifically, we set $C=3$ and $\rho(\b{A})=10$ to mitigate the influences of varying numbers of classes and spectral radius. 
% \sitao{Any other special settings that are worth mentioning here?}
% \yilun{nothing}

\begin{figure}
     \centering
     \begin{subfigure}{1\textwidth}
         \centering
         \includegraphics[width=\textwidth]{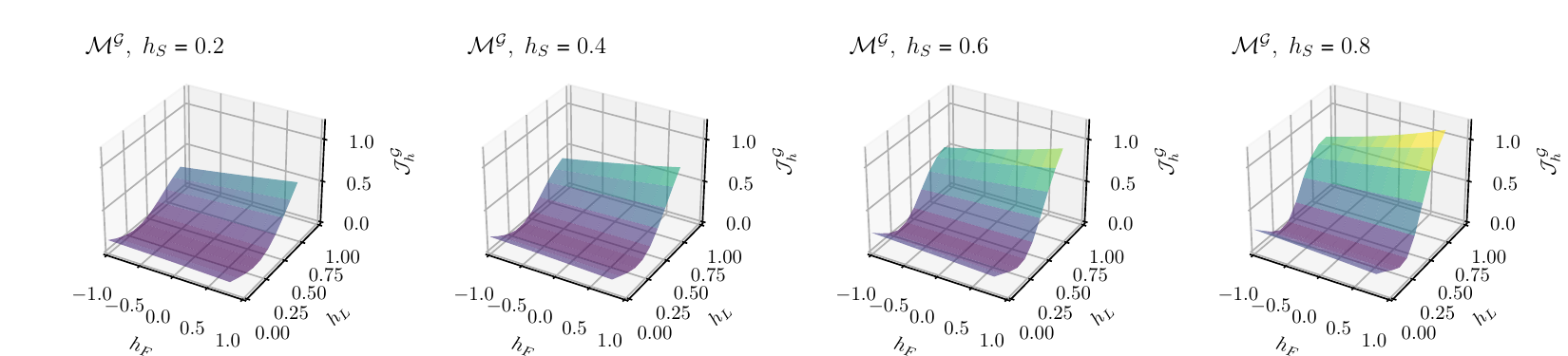}
         \caption{Numerical results of Tri-Hom}
         \label{fig:numerical_hom_obj_gnn}
     \end{subfigure}
     \hfill
     \begin{subfigure}{1\textwidth}
         \centering
         \includegraphics[width=\textwidth]{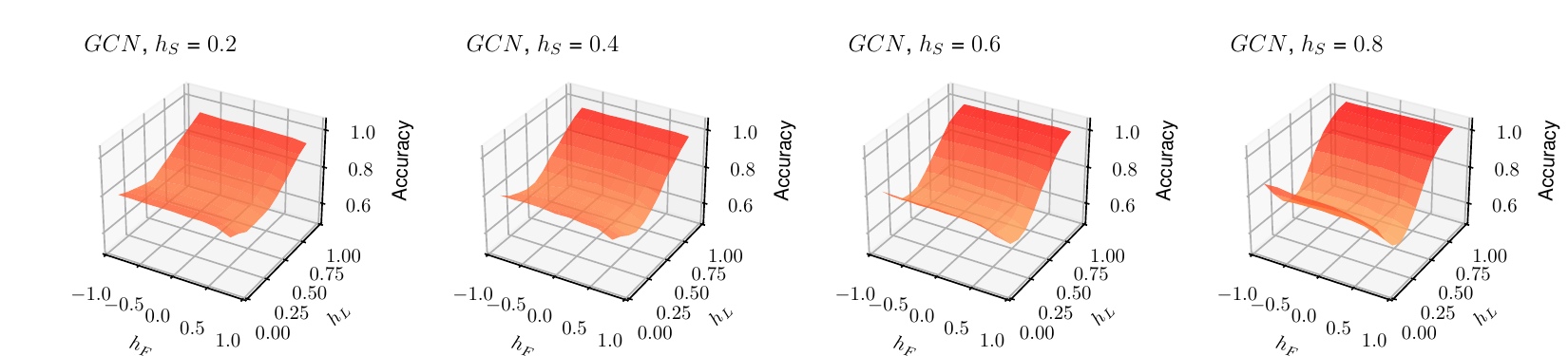}
         \caption{Model Performance on synthetic datasets}
         \label{fig:syn_hom_gnn}
     \end{subfigure}
     \hfill
    \vspace{-0.3cm}
    \caption{We measure the impact of label homophily $h_L$, feature homophily $h_F$, and structural homophily $h_S$ through numerical results of Tri-Hom $\mathcal{J}_h^{\mathcal{G}}$ and simulation results of the node classification accuracy with GCN on synthetic datasets.}\label{fig:impact_2d_plot_gnn}
\end{figure}

\vspace{-0.3cm}
\paragraph{Comparison and Analysis} The results are shown in Figure~\ref{fig:impact_2d_plot_gnn}. For a better demonstration, we only show the results for $h_S = \{0.2,0.4,0.6,0.8\}$ and each subfigure is a slice of $h_S$ that visualizes the impact of $h_L$ and $h_F$ on GCN and Tri-Hom. From the comparison of Figure~\ref{fig:numerical_hom_obj_gnn} and ~\ref{fig:syn_hom_gnn}, we have two main observations: 1) Overall, the impact of $h_L$, $h_S$, and $h_F$ on synthetic datasets aligns well with the numerical results of Tri-Hom $\mathcal{J}_h^{\mathcal{G}}$. The only difference is that $h_F$ seems to have less impact on GCN. We speculate that this is because the parameters in the graph filter in GCN are optimized during the training process, while these parameters are simplified as a fixed value in the theoretical analysis, leading to the difference\footnote{We also show the detailed influences of $h_L$, $h_S$, and $h_F$ individually to MLP or GCN in Appendix \ref{apd:exp_syn_single}, which correlates well with our theoretical results of the homophily impact.}. 2) Our theoretical analysis of the impact of $h_L, h_F$ and $h_S$ on Tri-Hom is consistent with GCN's behavior in Figure~\ref{fig:syn_hom_gnn}: Theorem 2.1 shows the worst performance of $\mathcal{M}^\mathcal{G}$ is reached when $h_L=\frac{1}{C}$, corresponding to the ravine in Figure~\ref{fig:syn_hom_gnn} for $h_L$; Theorem 2.2 shows an increase of $h_S$ consistently improves the performance of $\mathcal{M}^\mathcal{G}$, corresponding to the overall increases of the accuracy in Figure~\ref{fig:syn_hom_gnn} from the left to the right; Theorem 2.3 shows the influences of $h_F$ to $\mathcal{M}^\mathcal{G}$ is determined by the $h_L$. Even if this is not obvious in Figure~\ref{fig:syn_hom_gnn}, we show more detailed figures of the individual impact of $h_L$, $h_S$, and $h_F$ in Appendix \ref{apd:exp_syn_single}, confirming the impact of $h_F$ in Theorem 2.3.
% \sitao{Give some examples, \eg{} your theorem 2.1 shows when $H_L>1/C$, the performance of GNN will improve as $H_L$ get bigger, and this is consistent with the trained results in Figure~\ref{fig:syn_hom_gnn}, which verifies the effectiveness of your theoretical results.}

\vspace{-0.2cm}
\subsection{Experiments on Real-world Datasets}\label{exp:realworld}
\vspace{-0.2cm}
In this subsection, we show the superiority of our proposed Tri-Hom over the existing metrics by studying the correlation between the estimated metric values and model performance on real-world graph data.
% \sitao{Present the motivation. For example, "to verify the effectiveness of our proposed metrics, we test them on real-world datasets and compare with existing metrics."}
% \yilun{include all homophily metrics and acc in main paper?}
% settings, dataset statistics&hom, gcn&mlp performance

\vspace{-0.3cm}
\paragraph{Experimental Settings} To verify the effectiveness of our proposed Tri-Hom and compare with existing metrics, we train baseline models, MLP\footnote{We measure the performance of MLP to verify the effectiveness our proposed $\mathcal{J}_h^{\neg\mathcal{G}}$ and compare the performance gap between GNNs and MLP.}, GCN~\cite{GCN}, GraphSage~\cite{GraphSage}, GAT~\cite{GAT} and estimate metrics, $\mathcal{J}_h^{\neg\mathcal{G}}$, $\mathcal{J}_h^{\mathcal{G}}$\footnote{In the estimation of $\mathcal{J}_h^{\neg\mathcal{G}}$, $\mathcal{J}_h^{\mathcal{G}}$, we consider $h_L$, $h_S$, and $h_F$ as variables and let $C=3$ and $\rho(\b{A})=10$ to mitigate the influences of varying number of classes  and spectral radius.}, $h_F$, $h_S$, and $h_L$\footnote{We use $h_{node}$ as label homophily $h_L$ in our experiments.}, on 31 real-world heterophilic and homophilic datasets. These datasets include \textit{Roman-Empire, Amazon-Ratings, Mineweeper, Tolokers}, and \textit{Questions} from ~\cite{dataset_roman}; \textit{Squirrel, Chameleon, Actor, Texas, Cornell, Wisconsin} originally from ~\cite{Geom-GNN,musae} and refined by ~\cite{dataset_roman}; \textit{Cora, PubMed}, and \textit{CiteSeer} from ~\cite{dataset_cora}; \textit{CoraFull, Amazon-Photo, Amazon-Computer, Coauthor-CS}, and \textit{Coauther-Physics} from ~\cite{dataset_amazon}; \textit{Flickr} from ~\cite{dataset_flickr}; \textit{WikiCS} from ~\cite{dataset_wikics}; \textit{Blog-Catalog} from ~\cite{dataset_blog}; \textit{Ogbn-Arxiv} from ~\cite{dataset_arxiv}; \textit{Genius, Twitch-DE, Twitch-ENGB, Twitch-ES, Twitch-FR, Twitch-PTBR, Twitch-RU}, and \textit{Twitch-TW} from ~\cite{dataset_genius_twitch} \footnote{See the dataset statistics, training details and model performance on node classification in Appendix \ref{apd:dataset_realworld}, \ref{apd:train_detail} and \ref{apd:acc_node_cls}, respectively.}. Besides, we also calculate other graph homophily and performance metrics on the benchmark datasets, the metrics include label-based homophily: $h_\text{edge}$~\cite{def_hom_edge}, $h_\text{node}$~\cite{Geom-GNN}, $h_\text{class}$~\cite{def_hom_class}, $h_\text{adj}$~\cite{ana_label_info},  $h_\text{den}$~\cite{def_hom_den}, $h_\text{2hop}$~\cite{def_hom_2hop},  $h_\text{nei}$~\cite{def_hom_dom_nei}; structural-based homophily: $LI$~\cite{ana_label_info}, $h_\text{NS}$~\cite{is_hom_necessary}, $h_\text{agg}$~\cite{ACM_GCN}; feature-based homophily: $h_\text{GE}$~\cite{def_hom_GE}, $h_\text{LS-cos}$~\cite{def_hom_localsim}, $h_\text{LS-euc}$~\cite{def_hom_localsim}, $h_\text{attr}$~\cite{def_hom_attr}, $h_\text{CF}$~\cite{ana_feat_dist}; and classifier-based homophily metrics~\cite{when_do_graph_help}: $h_\text{KR}$, $h_\text{GNB}$, $h_\text{SVM}$ on these datasets. The detailed definitions of these metrics are summarized in Appendix \ref{apd:hom_measurement} and the details of all estimations are shown in Appendix \ref{apd:dataset_realworld}.

\input{tables/correlation_pearson_model}

\vspace{-0.3cm}
\paragraph{Correlations with Model Performance} To find out which metric can better align with the behavior of GNNs on graphs with different properties, in Table \ref{tab:pearson_acc_model}, we show Pearson correlation between all the metrics and model performance on the $31$ real-world datasets\footnote{In addition to the Pearson correlation, we show Kendall's Tau rank correlation~\cite{cor_kendall} in Appendix \ref{apd:exp_other_cor}. Besides, we show the correlations between the metrics with performance gaps between GNNs and MLP in Appendix \ref{apd:sec_cor_acc_diff}.}. For each model, the homophily metrics with the best, second, and third highest correlation values are highlighted in \textcolor{red}{red}, \textcolor{blue}{blue} and \textcolor{purple}{purple}, respectively. To get more robust comparison results, we rank the metrics for each model and report the average rank in the last column.
% In addition to the Pearson correlation, we also show Kendall's Tau rank correlation~\cite{cor_kendall} in Appendix \ref{apd:exp_other_cor}, which leads to the similar results as the Pearson correlation. We also measure the correlations between the metrics with performance gaps between GNNs and MLP in Appendix \ref{apd:sec_cor_acc_diff}, where $h_S$ exhibits a strong correlation with the performance gaps.

\vspace{-0.3cm}
\paragraph{Comparison and Analysis} The results show that: 1) $\mathcal{J}_h^{\mathcal{G}}$ achieves the highest correlation values with all models, much better than the other metrics that only consider a single aspect of graph data. This confirms the effectiveness of Tri-Hom in filling the missing part of graph homophily by taking all the three types of homophily $h_L,h_S,h_F$ into account, providing a comprehensive understanding of graph homophily; 2) $\mathcal{J}_h^{\neg\mathcal{G}}$, the Tri-Hom for $\mathcal{M}^{\neg\mathcal{G}}$, also shows high correlation values. This indicates the existence of structural-aware features of nodes in graphs, which justifies our modeling of feature homophily in CSBM-3H. Compared with other variants of CSBM~\cite{is_hom_necessary,when_do_graph_help}, which assume node features are conditionally independent of graph topology given node labels, the setting of CSBM-3H is closer to real-world scenarios. This is one of the important reasons that the metrics derived from CSBM-3H are better than the existing metrics.

%% file: tables/correlation_pearson_model.tex
\begin{table}[htbp]
  \centering
  \caption{Pearson correlation with p-value of all the metrics with model performance of node classification on 31 real-world datasets.}
  \resizebox{1\hsize}{!}{
    \begin{tabular}{lccccccccr}
    \toprule
     \multirow{2}{*}{Metric} & \multicolumn{2}{c}{MLP} & \multicolumn{2}{c}{GCN} & \multicolumn{2}{c}{GraphSage} & \multicolumn{2}{c}{GAT} & \multirow{2}{*}{Rank} \\
    \cmidrule{2-9}
    & Cor. & p-value& Cor. & p-value& Cor. & p-value& Cor. & p-value & \\
    \toprule
$h_\text{edge}$& 0.4441& 0.0123& 0.5663& 0.0009& 0.4737& 0.0071& 0.5344& 0.0020& 6.25\\
$h_\text{node}$& 0.4232& 0.0177& 0.5457& 0.0015& 0.4524& 0.0106& 0.5257& 0.0024& 8.00\\
$h_\text{class}$&\textcolor{red}{0.6078}& 0.0003&\textcolor{purple}{0.6120}& 0.0003& 0.5790& 0.0006&\textcolor{blue}{0.6169}& 0.0002&\textcolor{blue}{2.50}\\
$h_\text{adj}$& 0.4972& 0.0044& 0.5486& 0.0014& 0.4932& 0.0048& 0.5396& 0.0017& 5.25\\
$h_\text{den}$& 0.0038& 0.9839& 0.1483& 0.4258& 0.0525& 0.7791& 0.1258& 0.5001& 19.00\\
$h_\text{2hop}$& 0.4517& 0.0107& 0.5182& 0.0028& 0.4692& 0.0078& 0.4870& 0.0055& 7.50\\
$h_\text{nei}$& 0.3961& 0.0274& 0.4793& 0.0064& 0.4473& 0.0116& 0.4535& 0.0104& 10.75\\
$LI$& 0.4502& 0.0110& 0.4992& 0.0043& 0.4270& 0.0166& 0.4731& 0.0072& 9.75\\
$h_\text{NS}$& 0.2898& 0.1139& 0.3603& 0.0465& 0.3452& 0.0572& 0.3671& 0.0422& 14.00\\
$h_\text{agg}$& 0.5201& 0.0027& 0.5617& 0.0010&\textcolor{blue}{0.6040}& 0.0003& 0.5832& 0.0006& 3.75\\
$h_S$& 0.0981& 0.5994& 0.2345& 0.2042& 0.1981& 0.2854& 0.2886& 0.1153& 17.50\\
$h_\text{GE}$& 0.3641& 0.0440& 0.4501& 0.0111& 0.4347& 0.0145& 0.4094& 0.0222& 11.75\\
$h_\text{LS-cos}$& 0.3511& 0.0528& 0.4389& 0.0135& 0.4254& 0.0170& 0.4061& 0.0234& 13.00\\
$h_\text{LS-euc}$& 0.1272& 0.4953& 0.1101& 0.5555& 0.1117& 0.5498& 0.1168& 0.5313& 18.50\\
$h_\text{attr}$& 0.2022& 0.2754& 0.0990& 0.5963& 0.0735& 0.6945& 0.1121& 0.5482& 19.00\\
$h_\text{CF}$& 0.2549& 0.1664& 0.2890& 0.1149& 0.3154& 0.0840& 0.3167& 0.0825& 15.25\\
$h_F$& 0.4035& 0.0244& 0.4994& 0.0042& 0.4814& 0.0061& 0.4767& 0.0067& 8.50\\
$h_\text{KR}$& -0.5318& 0.0021& -0.3536& 0.0510& -0.3854& 0.0323& -0.3599& 0.0468& 22.00\\
$h_\text{GNB}$& -0.3796& 0.0352& -0.2440& 0.1858& -0.2828& 0.1232& -0.2421& 0.1894& 21.00\\
$h_\text{SVM}$& 0.2430& 0.1878& 0.2741& 0.1356& 0.3320& 0.0681& 0.2961& 0.1058& 15.75\\
$\mathcal{J}_h^{\neg\mathcal{G}}$&\textcolor{blue}{0.5800}& 0.0006&\textcolor{blue}{0.6286}& 0.0002&\textcolor{purple}{0.5978}& 0.0004&\textcolor{purple}{0.6136}& 0.0002&\textcolor{blue}{2.50}\\
$\mathcal{J}_h^{\mathcal{G}}$&\textcolor{purple}{0.5471}& 0.0014&\textcolor{red}{0.6650}& 0.0000&\textcolor{red}{0.6223}& 0.0002&\textcolor{red}{0.6731}& 0.0000&\textcolor{red}{1.50}\\
    \bottomrule
    \end{tabular}%
}
  \label{tab:pearson_acc_model}%
\end{table}%

%% file: sections/Conclusion.tex
In this paper, we study the missing components of graph homophily by disentangling it from label, structural, and feature perspectives. Compared with previous homophily metrics, the combination of the three homophily metrics provides a unique and comprehensive understanding of graph homophily. Notably, our proposed feature homophily can measure the feature dependencies among nodes, fully disentangling itself from the label and structural homophily, which helps us analyze the disentangled impact. The theoretical study on CSBM-3H leads us to Tri-Hom, a combination of the three types of homophily. The synthetic experiments on CSBM-3H verify the theoretical results and the effectiveness of Tri-Hom. The high correlation with GNN behaviors on $31$ real-world benchmark datasets confirms the superiority of Tri-Hom over $17$ existing metrics.

In the future, it will be interesting to investigate the disentangled graph homophily in more general CSBM settings without some assumptions, \eg{} uniform node degrees, balanced class, and linear feature dependencies. Additionally, the estimation of three types of homophily in unsupervised or weak-supervised scenarios would be important for label-scarcity cases. Furthermore, the theoretical results in the paper reveal a nuanced understanding of graph homophily, potentially inspiring future studies on model designs that address heterophily issues in GNNs.

%To investigate how three types of homophily influence model performance, we propose a CSBM-3H that incorporates three types of homophily. Based on our theoretical and numerical results, we find that the performance of graph-aware models is closely related to a new metric named Tri-Hom, a combination of three types of homophily. Our experimental results on synthetic datasets verify our theoretical findings, while real-world dataset results show a strong correlation between Tri-Hom and GNNs. This confirms the effectiveness of Tri-Hom as well as the necessity of disentangling graph homophily. However, we acknowledge that our assumptions, including uniform node degrees, class balance, and the simplification of feature dependencies using linear relationships may limit the generality of our results. In the future, it will be interesting to investigate the disentangled graph homophily in a more general case with fewer assumptions. Additionally, it will also be interesting to estimate three types of homophily in unsupervised or weak-supervised approaches. Furthermore, the theoretical results in the paper reveal a nuanced understanding of graph homophily, potentially inspiring future studies on model designs that address heterophily issues in GNNs.

%% file: sections/Appendix.tex
\section{Related Work on Homophily Measurements}\label{apd:hom_measurement}
% GNNs
% Graph homophily
% Graph homophily metrics&analysis

In general, graph homophily metrics can be categorized as either statistic-based metrics or classifier-based metrics. The former can be further classified into the homophily on label, structural, or feature aspect. The definitions of these metrics are introduced as follows.

% \textbf{Label homophily} measures graph homophily based on labels across the whole graph structure. Since labels are the highly compressed information of nodes, label homophily is the most widely used metric for graph homophily. \textbf{Structural homophily} measures the graph homophily by considering the node structural information across the intra-class nodes. Generally, structural homophily can well explain the phenomenon in the cases of "good" heterophily where the performance of GNNs remains good in low heterophily. \textbf{Feature homophily} utilizes node features to measure graph homophily. In a graph, node features are not independent and identically distributed, thus the feature homophily could reflect a different aspect as label homophily.

\subsection{Homophily on Label Aspect}

\textbf{Edge homophily}~\cite{def_hom_edge} measures the graph homophily at the edge level, which is defined as the fraction of edges in a graph that connects nodes with the same labels:
\begin{equation}
h_{edge}(\mathcal{G},\b{Y}) = \frac{\big|\{e_{uv} \mid e_{uv}\in \mathcal{E}, Y_{u}=Y_{v}\}\big|}{|\mathcal{E}|}
\end{equation}

\textbf{Node homophily}~\cite{Geom-GNN} measures the graph homophily at the node level, where the homophily degree for each node is computed as the proportion of the neighbors sharing the same class. Then the node homophily for the whole graph is defined as the average homophily degree for all the nodes:
\begin{equation}
h_{node}(\mathcal{G},\b{Y}) = \frac{1}{|\mathcal{V}|} \sum_{u \in \mathcal{V}}\frac{\big|\{u \mid u \in \mathcal{N}_u, Y_{u}=Y_{v}\}\big|}{d_u}, \\
\end{equation}

\textbf{Class homophily}~\cite{def_hom_class} addresses class imbalance by treating all classes equally. This metric mitigates the sensitivity of edge homophily and node homophily to the number of classes and nodes in each class. The definition of class homophily is given by:
\begin{equation}
h_{class}(\mathcal{G},\b{Y}) = \frac{1}{C-1} \sum_{c=1}^{C}\bigg[\frac{\sum_{u \in \mathcal{V}, Y_{u} = c } \big|\{u \mid  u \in \mathcal{N}_v, Y_{u}=Y_{v}\}\big| }{\sum_{u \in \{u|Y_u=c\}} d_{u}} - \frac{N_c}{N}\bigg]_{+}
\end{equation}

\textbf{Adjusted homophily}~\cite{ana_label_info} considers the probability of an edge endpoint connecting to a node with a particular class and adjusts the edge homophily using node degrees, which is defined as:
\begin{equation}
h_{adj}(\mathcal{G},\b{Y}) = \frac{h_{edge}(\mathcal{G},\b{Y})-\sum_{c=1}^C\frac{D^2_c}{(2|\mathcal{E}|)^2}}{1-\sum_{c=1}^C\frac{D^2_c}{(2|\mathcal{E}|)^2}}
\end{equation}
where $D_c$ represents the total degree of class $c$, \ie{$D_k = \big|\{e_{uv}|Y_u=c\ \text{or}\ Y_v=c,e_{uv}\in\mathcal{E}\}\big|$}.

\textbf{Density-aware homophily}~\cite{def_hom_den} is introduced as an improvement over class homophily, which only captures relative edge proportions and disregards graph connectivity. This limitation results in inflated homophily scores for highly disconnected graphs~\cite{ana_label_info}. The proposed density-aware homophily aims to provide a more accurate measurement of edge density and is defined as:
\begin{equation}
h_{den}(\mathcal{G},\b{Y}) = \frac{1+\min{\{\zeta_c-\hat{\zeta}_c\}}_{c=1}^C}{2}
\end{equation}
where $\zeta_k$ is the edge density of the subgraph formed by intra-class edges of class $k$ and $\hat{\zeta}_k$ is the maximum intra-class edge density of class $k$.

\textbf{2-hop Neighbor Class Similarity}~\cite{def_hom_2hop}. Since the information of 1-hop neighbors might be less representative or even misleading~\cite{def_hom_2hop}, 2-hop Neighbor Class Similarity extends the concept of "neighbors" from 1-hop to 2-hop, which is defined as:
\begin{equation}
h_{2hop}(\mathcal{G},\b{Y}) = \frac{1}{|\mathcal{V}|} \sum_{u \in \mathcal{V}}\frac{\big|\{u \mid u \in \mathcal{N}_u^{(2)}, Y_{u}=Y_{v}\}\big|}{d_u}, \\
\end{equation}
where $\mathcal{N}_u^{(2)}=\{ \bigcup\limits_{v\in\mathcal{N}_u} \mathcal{N}_v \}\backslash\{u\} $ represents the two-hop neighbors of node $u$.

\textbf{Neighbor Homophily}~\cite{def_hom_dom_nei} is proposed to address the "good" and "bad" heterophily issue by considering the dominant neighbors. 
The homophily score for any given node $u$ is based on the number of nodes in which its class holds dominance among its k-hop neighbors. The definition of neighbor homophily in a graph is defined as:
% \footnote{This is originally called neighbor homophily~\cite{def_hom_dom_nei}. However, this homophily measurement doesn't contain all the neighbor information, instead, it is only partial of neighbor information by only considering the dominant neighbors according to their classes. Therefore, we call this dominant-neighbor homophily in this paper.}

\begin{equation}
h_{DN}(\mathcal{G},\b{Y},k) =  \frac{1}{|\mathcal{V}|} \sum_{u \in \mathcal{V}}\frac{\max{\big\{\big|v|v\in\mathcal{N}_u^{(k)}\big|\big\}}_{c=1}^C}{\big|\mathcal{N}_u^{(k)}\big|} \\
\end{equation}
where $\mathcal{N}_v^{(k)}$ represents the k-hop neighbors of node $u$.

\subsection{Homophily on Structure Aspect}

\textbf{Label informativeness}~\cite{ana_label_info} measures the informativeness of a neighbor's label for a node's label using condition entropy:
\begin{equation}
LI(\mathcal{G},\b{Y}) = -\frac{\sum_{c_1,c_2}\log\frac{p(c_1,c_2)}{\bar{p}(c_1),\bar{p}(c_2)}}{\sum_c \bar{p}(c)\log\bar{p}(c)} = 2-\frac{\sum_{c_1,c_2}\log p(c_1,c_2)}{\sum_c \bar{p}(c)\log\bar{p}(c)}
\end{equation}
where $p(c_1,c_2)=\sum_{(u,v)\in\mathcal{E}}\frac{\mathbbm{1}\{Y_u=c_1,Y_v=c_2\}}{2\abs{\mathcal{E}}}$ represents mutual distribution for a randomly sampled edge from class $c_1$ to class $c_2$ and $\bar{p}_c = \frac{D_c}{2\abs{\mathcal{E}}}$ represents the distribution of the node degree for class $c$.

\textbf{Neighborhood Similarity}~\cite{is_hom_necessary} measures the similarity of the neighbor distributions between two classes. A high similarity of intra-class neighbor distributions and a low similarity of inter-class neighbor distributions ensure the neighborhood patterns for nodes with different labels are distinguishable~\cite{is_hom_necessary}. Therefore, the ratio of inter-class to intra-class neighborhood similarity could reflect the performance of GNNs\footnote{This definition is not directly given in the original paper~\cite{is_hom_necessary}. We define this ratio based on the proposition
~\cite{is_hom_necessary} that a high intra-class neighborhood similarity and a low inter-class neighborhood similarity improves GNN performance.}
\begin{equation}
h_{NS}(\mathcal{G},\b{D^{\mathcal{N}}} ,\b{Y}) =  \frac{\mathbb{E}_{Y_u=Y_v}[\cos(D^\mathcal{N}_u,D^\mathcal{N}_v)]}{\mathbb{E}_{Y_u\neq Y_v}[\cos(D^\mathcal{N}_u,D^\mathcal{N}_v)]}
\end{equation}
where $\mathcal{N}_v^{(k)}$ is the k-hop neighbors of node $u$.

\textbf{Aggregation homophily}~\cite{ACM_GCN} measures the ratio of nodes in a graph that has a higher intra-class aggregation similarity than inter-class aggregation similarity. The aggregation similarity of node $u$ and $u$ is the multiplication of their neighbor distribution \ie{$D^\mathcal{N}_uD^\mathcal{N}_v$}. The definition of aggregation homophily is given as:

\begin{equation}
    h_{agg}(\mathcal{G},\b{D^{\mathcal{N}}},\b{Y}) = \frac{1}{\abs{\mathcal{V}}}\abs{\biggl\{u\biggl|\frac{\sum_{Y_u=Y_v} D^\mathcal{N}_uD^\mathcal{N}_v}{\abs{\{v|Y_u=Y_v\}}}\ge\frac{\sum_{Y_u\neq Y_v} D^\mathcal{N}_uD^\mathcal{N}_v}{\abs{\{v|Y_u\neq Y_v\}}},v\in\mathcal{V},u\in\mathcal{V}\biggl\}}
\end{equation}

\subsection{Homophily on Feature Aspect}

\textbf{Generalized edge homophily}~\cite{def_hom_GE} defines the feature homophily in graphs as the feature consistency across the graph topology:
\begin{equation}
h_{GE}(\mathcal{G},\b{X}) =\frac{1}{|\mathcal{E}|} \sum_{e_{uv}\in\mathcal{E}}\frac{\b{X}_u\b{X}_v}{\norm{\b{X}_u}\norm{\b{X}_u}}
\end{equation}
The difference between this homophily with edge homophily is that generalized edge homophily replaced the indicator function of two connected nodes in edge homophily, \ie{} $\mathbbm{1}\{Y_u=Y_v\}$, to a similarity measurement of node features, \ie{} $\text{sim}(\b{X}_u,\b{X}_v)$.

\textbf{Local Similarity}~\cite{def_hom_localsim} measures feature homophily at the node level based on the hypothesis that nodes with similar features are likely to belong to the same class. The definition is given based on either cosine similarity
\begin{equation}
h_{LS-cos}(\mathcal{G},\b{X}) = \frac{1}{|\mathcal{V}|} \sum_{u \in \mathcal{V}}\frac{1}{d_u}\sum_{v\in\mathcal{N}_u}\frac{\b{X}_u\b{X}_v}{\norm{\b{X}_u}\norm{\b{X}_u}} \\
\end{equation}
or Euclidean similarity.
\begin{equation}
h_{LS-euc}(\mathcal{G},\b{X}) = \frac{1}{|\mathcal{V}|} \sum_{u \in \mathcal{V}}\frac{1}{d_u}\sum_{v\in\mathcal{N}_u}(-\norm{\b{X}_u-\b{X}_v}_2) \\
\end{equation}

\textbf{Attribute homophily}~\cite{def_hom_attr} considers the homophily with respect to each feature, which is defined as:
\begin{equation}
    \begin{split}
    h_{attr,m}(\mathcal{G},\b{X}_{:,m}) &= \frac{1}{\sum_{u\in\mathcal{V}}X_{u,m}}\sum_{u\in\mathcal{V}} \left(X_{u,m}\frac{\sum_{v\in\mathcal{N}_u}X_{v,m}}{d_u}\right)\\
    h_{attr}(\mathcal{G},\b{X}) &= \sum_{m=1}^M h_{attr,m}(\mathcal{G},\b{X}_{:,m}),\\
    \end{split}
\end{equation}
where $h_{attr}$ represents the attribute homophily for the whole graph and $h_{attr,m}$ represents the attribute homophily for feature $m$.

\textbf{Class-controlled feature homophily}~\cite{ana_feat_dist} considers the interplay between graph topology and feature dependence through the disparity of nodes' expected distances to random nodes with their neighbors, which is defined as:
% cannot fully disentangle from label homophily; just the relative difference of class-wise variance.
\begin{equation}
\begin{split}
    h_{CF}(\mathcal{G},\b{X},\b{Y})&=\frac{1}{\abs{\mathcal{V}}}\sum_{u\in\mathcal{V}}\frac{1}{d_u}\sum_{v\in\mathcal{N}_u}\Bigl(\b{d}(v,\mathcal{V}\backslash\{u\})-\b{d}(v,\{u\})\Bigl)\\
    \b{d}(u,\mathcal{V}')&=\frac{1}{\abs{\mathcal{V}'}}\sum_{v\in\mathcal{V}'}\norm{(\b{X}_u|\b{Y})-(\b{X}_v|\b{Y})}\\
    \b{X}_u|\b{Y}&=\b{X}_u-\left(\frac{\sum_{Y_u=Y_v}\b{X}_v}{\abs{\{ v|Y_u=Y_v,v\in\mathcal{V} \}}}\right)
\end{split}
\end{equation}
where $\b{X}_u|\b{Y}$ represents class-controlled features and $\b{d}(\cdot)$ denotes a distance function.

\subsection{Classifier-based homophily}
\textbf{Classifier-based homophily}~\cite{when_do_graph_help} uses a classifier to capture the feature-based linear or non-linear information without iterative training. To determine when graph-aware models perform better than graph-agnostic models, a hypothesis test is conducted on the original feature $\b{X}$ and the aggregated features $\b{H}$, as shown below:
\begin{align*}
    &\resizebox{1\hsize}{!}{$\text{H}_0: \text{Prop}(\text{G-aware model}) \geq \text{Prop}(\text{G-agnostic model}); \ \text{H}_1: \text{Prop}(\text{G-aware model})<\text{Prop}(\text{G-agnostic model})$}
\end{align*}
The resulting p-value from this hypothesis test can indicate whether the performance of $\b{H}$ is superior to that of $\b{X}$. Three types of classifier, Gaussian Naive Bayes, Kernel Regression, and Support Vector Machine are used in ~\cite{when_do_graph_help}, which correspond to the metrics $h_{GNB}$, $h_{KR}$, and $h_{SVM}$ in this paper.

\section{Structural-Aware Node Features}\label{apd:diff_particle}
This section explores the impact of graph topology on node features. Unlike the data in Euclidean space, where samples are \iid, the data sampled from a graph for each node are structural-aware \ie{$(\b{X}\not\!\perp\!\!\!\perp\b{A}|\b{Y})$}. To model the feature dependencies of nodes in graphs, we follow \cite{diff_particle_sys} and adopt a graph diffusion process~\cite{GRAND}, which is shown as follows
\begin{equation}\label{eq:diffusion}
\begin{split}
    \b{X}(t) &= \b{X}(0) + \int_{0}^{T} \frac{\partial \b{X}(t)}{\partial t} \,dt, \\
    \text{where}\ \frac{\partial \b{X}(t)}{\partial t}&=(\mathcal{F}(\b{A})-\b{I})\b{X}(t-1)+\b{X}(0)
\end{split}
\end{equation}
Here the structural-agnostic node features $\b{X}(0)\sim p(\b{X}|\b{Y})$ are sampled from the distributions with respect to each class, while the structural-aware node features $\b{X}(t>0)\sim p(\b{X}|\b{Y},\b{A})$ are generated through the diffusion process. In $\frac{\partial \b{X}(t)}{\partial t}$, the first term describes how the dependencies are introduced with a feature dependency function $\mathcal{F}:(\b{A})\mapsto\mathbb{R}^{N\times N}$ and the second term $\b{X}(0)$ preserves the node distinguishability. Then, based on the Eq. (\ref{eq:diffusion}), we have
\begin{equation}\label{eq:diffusion_res}
    \b{X}(t) = \b{X}(0) + \mathcal{F}(\b{A})\b{X}(t-1)= \b{X}(0) + (\mathcal{F}(\b{A}))^1\b{X}(0) + \cdots + (\mathcal{F}(\b{A}))^t\b{X}(0)
\end{equation}

We can also interpret the process as an interactive particle system~\cite{diff_particle_sys} where all node features collapse into an equilibrium eventually. The equilibrium requires Eq. (\ref{eq:diffusion_res}) to have a closed form when $t\to \infty$, implying the spectral radius of the adjacency matrix $\rho(\mathcal{F}(\b{A}))<1$. This constraint also implies that $\abs{\mathcal{F}(\b{A})}^{k}\ge\abs{\mathcal{F}(\b{A})}^{k-1}$ for $k>0$, which aligns with the common relations in graphs that nodes are likely to have higher dependencies with their closer neighbors than with farther neighbors.

For simplicity and a better understanding of feature dependencies, we consider a linear case as in ~\cite{GRAND,diff_particle_sys} and use a parameter $\omega$ to control the feature dependencies \ie{$\mathcal{F}(\b{A})=\omega\b{A}$} with a range of $(-\frac{1}{\rho(\b{A})},\frac{1}{\rho(\b{A})})$. Then we can represent the structural-aware features as

\begin{equation}
    \b{X} = \Bigr[ \sum_{t=0}^\infty (\omega\b{A})^t \Bigr] \b{X} = (\b{I}-\omega\b{A})^{-1}\b{X}(0)
\end{equation}

\section{Impact of Graph Homophily on Graph-agnostic Models}\label{apd:hom_impact_mlp}
In addition to the impact of label homophily $h_L$, structural homophily $h_S$, and feature homophily $h_F$ on Graph-aware models as discussed in Section \ref{sec:node_distinguish}, we further discuss the impact on Graph-agnostic models through ${\mathcal{J}}_h^{\neg\mathcal{G}}$ in this section. Specifically, we compute the partial derivative of $\mathcal{J}_h^{\neg\mathcal{G}}$ with respect to $h_S$, $h_F$, and $h_L$ to reveal the analytical influences. (See calculation in Appendix~\ref{apd:proof_theorem31}, \ref{apd:proof_theorem32}, and \ref{apd:proof_theorem33}.)
\begin{theorem} {3.1}
    The partial derivative of ${\mathcal{J}}_h^{\neg\mathcal{G}}$ with respect to label homophily $h_L$ satisfies,
    \begin{equation}
        \begin{split}
        \frac{\partial\mathcal{J}_h^{\neg\mathcal{G}}}{\partial h_L}
        \begin{cases}
        <0, & \text{if}\ h_F\in(-1,0)\\
        \geq 0, & \text{if}\ h_F\in[0,1)
        \end{cases}
        \end{split}
    \end{equation}
\end{theorem}
From Theorem 3.1 we can see that, under a positive $h_F$ \ie{the features of connected nodes become similar}, the increase of $h_L$ makes the features of intra-class nodes more distinguishable, thereby improving the performance of $\mathcal{M}^{\neg\mathcal{G}}$. Conversely, under a negative $h_F$ \ie{the features of connected nodes become dissimilar}, the increase of $h_L$ makes the features of intra-class nodes more indistinguishable, resulting in a degradation of the performance of $\mathcal{M}^{\neg\mathcal{G}}$.
\begin{theorem} {3.2}
    The partial derivative of ${\mathcal{J}}_h^{\neg\mathcal{G}}$ with respect to structural homophily $h_S$ satisfies,
    \begin{equation}
        \begin{split}
            \frac{\partial\mathcal{J}_h^{\neg\mathcal{G}}}{\partial h_S}\ge0
        \end{split}
    \end{equation}
\end{theorem}

From Theorem 3.2 we can see that, since the existence of feature dependencies, a larger $h_S$ makes the structural-aware features more distinguishable among different classes, improving the performance of $\mathcal{M}^{\neg\mathcal{G}}$.
\begin{theorem} {3.3}
    The partial derivative of ${\mathcal{J}}_h^{\neg\mathcal{G}}$ with respect to feature homophily $h_F$ satisfies,
    \begin{equation}
        \begin{split}
        \frac{\partial\mathcal{J}_h^{\neg\mathcal{G}}}{\partial h_F}
            \begin{cases} 
            < 0, & \text{if}\ h_L\in(0 ,h_L^-); h_L \in( h_L^-,h_L^+)\ \text{and} \ h_F \in (\hat{h}_F,1)\\
            > 0, & \text{if}\ h_L\in(h_L^+,1]; h_L\in(h_L^-,h_L^+)\ \text{and}\ h_F\in(-1,\hat{h}_F) \\
            = 0, & \text{if}\ h_L\in(h_L^-,h_L^+)\ \text{and}\ h_F=\hat{h}_F
            \end{cases}
        \end{split}
    \end{equation}
    where $0<h_L^{*,-} < h_L^{*,+}<1$ and $-1<h_F^*<1$.
\end{theorem}

From Theorem 3.3 we can see, how the impact of $h_F$ on $\mathcal{M}^{\neg\mathcal{G}}$ is determined by $h_L$. This result is similar to the Theorem 2.3 in the case of $\mathcal{M}^{\mathcal{G}}$. The increase of $h_F$ makes node features more similar to their neighbors. As a result, under a high $h_L$, features of intra-class nodes become more similar, thereby improving the performance of $\mathcal{M}^{\neg\mathcal{G}}$; under a low $h_L$, features of intra-class nodes becomes more similar, thereby reducing the performance of $\mathcal{M}^{\neg\mathcal{G}}$.

\include{sections/appendix_sections/ExperimentalDetails}

% For both the $\mathcal{M}^{\neg\mathcal{G}}$ and $\mathcal{M}^{\mathcal{G}}$, we can see that increasing or decreasing $h_L$ and $h_F$ simultaneously leads to a higher $\mathcal{J}_h$. This resulting in two extremums, $(h_N=1, h_L=1, h_F=1)$ and $(h_N=1, h_L=0, h_F=-1)$, where the $\mathcal{J}_h$ of the former is higher than the latter. As for the $h_N$, it consistently increases the $\mathcal{J}_h$. This numerical results align with our analytical results as shown in Theorem 2.1, 2.2, and 2.3.

\include{sections/appendix_sections/ProofofTheorems}

%% file: sections/appendix_sections/ExperimentalDetails.tex
\section{Experimental Details}\label{apd:exp_setting}

\subsection{Datasets}

\subsubsection{Synthetic Datasets}\label{apd:dataset_syn}
We show the detailed process of constructing CSBM-3H in Algorithm \ref{alg:CSBM-3H}. First, graph topology is constructed with label homophily $h_L$ and structural homophily $h_S$. Then, structural-aware node features are constructed with feature homophily $h_F$. To investigate how three types of homophily influence the model performance, we generate graphs with $h_L\in[0,0.1,\dots,1]$, $h_S\in[0,0.1,\dots,1]$, and $h_F\in[-0.8,-0.6,\dots,0.8]$. For each given $h_L$, $h_S$, and $h_F$, we generate the graphs with $10$ seeds from $[0,1,\dots,9]$ to mitigate random deviations. Each graph contains 1000 nodes distributed across 3 classes, with node degrees sampled from a uniform distribution in the range

\begin{algorithm}
\caption{Stochastic Block Model controlled with 3 types of Homophily(CSBM-3H)}\label{alg:CSBM-3H}
\begin{algorithmic}
\Require Label homophily $h_L$, structural homophily $h_S$, and feature homophily $h_F$, \\
number of nodes $N$, number of classes $C$, node degree $\b{D}$, \\
dimension of node features $M$, random seed $\phi$, \\
class-wised Gaussian mean $\b{\mu}=[\b{\mu_1},\dots,\b{\mu_C}]^T$, covariance $\b{\Sigma}=[\b{\Sigma_1},\dots,\b{\Sigma_C}]^T$ with $\b{\mu_c}\in\mathcal{R}^{M}$ and $\b{\Sigma_c}\in\mathcal{R}^{M\times M}$ for class $c$.
\Ensure Adjacency matrix $\b{A}$, node features $\b{X}$ and node labels $\b{Y}$ of a synthetic graph.
\State Initialize node features $\b{X}$, node labels $\b{Y}$, with $\b{X}\in\mathbb{R}^{N\times M}$, $\b{Y}\in\mathbb{R}^{N}$. Set random seed $\phi$.

\For{$n=0$ to $N$} \Comment{Sample node labels}
\State $k \sim \lfloor \text{Uniform}(0, C) \rfloor$ 
\State $Y_{n} \gets k$
\EndFor

\State $\b{Z}\gets \text{One-hot}(\b{Y})$ \Comment{One-hot encoding of node labels}

\State $\b{S}\gets h_L\b{I}_C+ \frac{1-h_L}{C-1}(\b{1}_C-\b{I}_C)$
\Comment{Introduce label homophily $h_L$}
\State $\b{D}^\mathcal{N}\gets \b{ZS}$
\Comment{Construct neighbor distribution}
\For{$D^\mathcal{N}_{u,c}\in\b{D}^\mathcal{N}$}
\State $\epsilon\sim \text{Normal}(0,\frac{(1-h_S)^2}{C-1})$
\Comment{Introduce structural homophily $h_S$}
\State $D_{u,c}^\mathcal{N} \gets D_{u,c}^\mathcal{N}+\epsilon$ 
\EndFor
\State $\b{\hat{D}} \gets \text{diag}(\b{D})$
\State $\b{A_p}\sim\frac{C}{N}\b{\hat{D}}^{-\frac{1}{2}}\b{D}^\mathcal{N}\b{\hat{D}}^{-\frac{1}{2}} \b{Z}^T$
\Comment{Construct neighbor sampling matrix with node degrees}
\State $\b{A_p}\gets \max(0,\min(1,\b{A_p}))$
\Comment{Bound with $[0,1]$ before sampling}
\State $\b{A} \gets <\text{Sym}\circ \text{Binarize} >(\b{A_p})$
\Comment{Binarization sampling and symmetrize adjacency matrix}
\For{$n=0$ to $N$}
\Comment{Sample Structural-agnostic node features}
\State $\b{X_{n,:}}\sim \b{\text{Normal}}(\b{\mu}_{Y_n},\b{\Sigma}_{Y_n})$
\EndFor

\State $\b{X} \gets (\b{I}_C-\frac{h_F}{\sigma(\b{A})}\b{A})^{-2}\b{X}$
\Comment{Introduce feature homophily $h_F$}
\State
\Return $\b{A},\b{X},\b{Y}$

\end{algorithmic}
\end{algorithm}

For the random graph generative models with homophily, current studies~\cite{ana_label_info,ana_feat_dist,when_do_graph_help,is_hom_necessary} generally adopt a Contextual Stochastic Block Model with Homophily(CSBM-H) to control the label homophily $h_L$ through assigning nodes with different probabilities that connect to the nodes from other classes. Then the node features are sampled solely based on the classes. However, these random graph generative models have two drawbacks: First, the probabilities of nodes connecting to the nodes with different classes are uniform, which lacks diversity. Second, the sampled node features are independent with their structures \ie{$(\b{X}\!\perp\!\!\!\perp\b{A}|\b{Y})$}, which is uncommon in real-world scenarios where interactions influence the attributes of connected nodes~\cite{feat_dpd_recm_2,feat_dpd_social_1,feat_dpd_social_2}. Our proposed CSBM-3H well address these drawbacks by considering $h_S$ and $h_F$, thereby providing a more comprehensive and realistic model.

\subsubsection{Real-World Datasets}\label{apd:dataset_realworld}
We conduct our experiments on 31 real-world datasets: Roman-Empire, Amazon-Ratings, Mineweeper, Tolokers, and Questions from ~\cite{dataset_roman}; Squirrel, Chameleon, Actor, Texas, Cornell, Wisconsin originally from ~\cite{Geom-GNN} and refined by ~\cite{dataset_roman}; Cora, PubMed, and CiteSeer from ~\cite{dataset_cora}; CoraFull, Amazon-Photo, Amazon-Computer, Coauthor-CS, and Coauther-Physics from ~\cite{dataset_amazon}; Flickr from ~\cite{dataset_flickr}; WikiCS from ~\cite{dataset_wikics}; Blog-Catalog from ~\cite{dataset_blog}; Ogbn-Arxiv from ~\cite{dataset_arxiv}; Genius, Twitch-DE, Twitch-ENGB, Twitch-ES, Twitch-FR, Twitch-PTBR, Twitch-RU, and Twitch-TW from ~\cite{dataset_genius_twitch}. These datasets contain both the homophilic and heterophilic graphs that come from citation networks, webpage networks, purchase networks, image description networks, coauthor networks, actor networks, and social networks. The diversity of these datasets enables us to evaluate the model performance in a general case. For these datasets, we show basic statistics in Table \ref{tab:real_world_datasets_statistics} and the graph homophily metrics or model performance metrics in Table \ref{tab:real_world_datasets_metrics}.
\input{tables/dataset_statistics_basics}
\input{tables/dataset_statistics_hom}

\subsection{Training Detail}\label{apd:train_detail}
For all the datasets, we randomly split the train, validation, and test set as 50\%:25\%:25\% for 10 runs. We use the Adam optimizer~\cite{adam} with a learning rate of 0.001. The maximum training epoch is set to 1000 with a patience of 40 for early stopping. To enhance performance, we incorporated skip connections~\cite{residual} and layer normalization~\cite{layer_norm} in each layer. All models are trained on a single NVIDIA RTX A5000 GPU with 24GB memory. For hyperparameter tuning, we perform a grid search on the validation set. The search space included the following hyperparameters:
\begin{itemize}
    \item Number of layers: \{1, 2\},
    \item Hidden dimension: \{64, 128, 256\},
    \item Dropout rate: \{0.2, 0.4, 0.6, 0.8\},
    \item Weight decay: \{1e-3, 1e-4, 1e-5\}.
\end{itemize}

\subsection{Node Classification Performance}\label{apd:acc_node_cls}
\input{tables/node_classification_acc}

Table \ref{tab:node_cls_acc_realworld} shows the averaged mean and stand deviation accuracy of node classification performance for MLP, GCN, GraphSage, and GAT across 10 runs. Notably, graph-aware models $\mathcal{M}^{\mathcal{G}}$ outperform graph-agnostic models $\mathcal{M}^{\neg\mathcal{G}}$ in most datasets. This phenomenon holds true for both homophilic datasets, such as Cora, Citeseer, and PubMed, and heterophilic datasets, such as Roman-empire, Chameleon-filtered, and Flickr. Consequently, relying solely on label homophily is insufficient to determine the performance of $\mathcal{M}^{\mathcal{G}}$, which aligns with previous studies~\cite{is_hom_necessary,when_do_graph_help}

\subsection{Tri-Hom for Graph-agnostic Models}
\begin{figure}
     \centering
     \begin{subfigure}{1\textwidth}
         \centering
         \includegraphics[width=\textwidth]{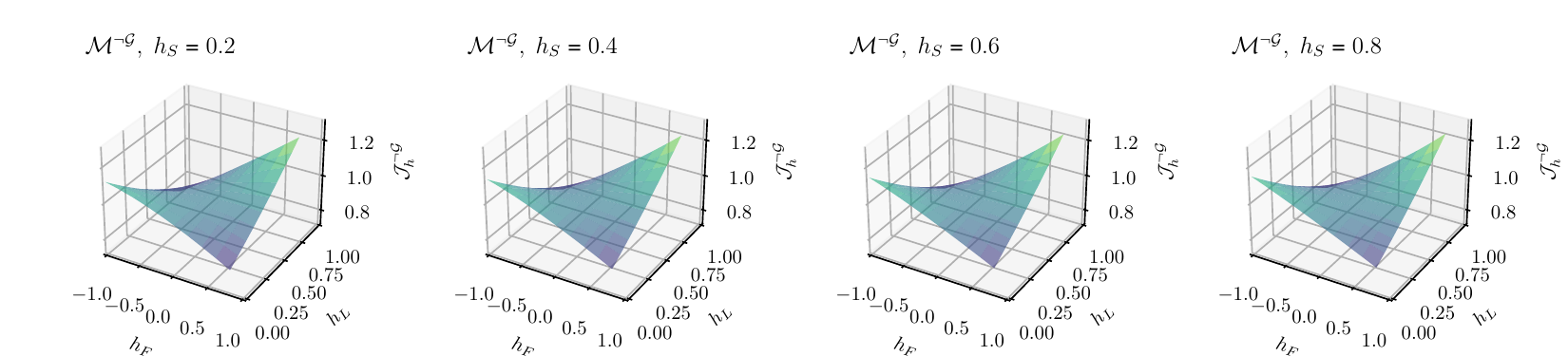}
         \caption{Numerical results of Tri-Hom}
         \label{fig:numerical_hom_obj_mlp}
     \end{subfigure}
     \hfill
     \begin{subfigure}{1\textwidth}
         \centering
         \includegraphics[width=\textwidth]{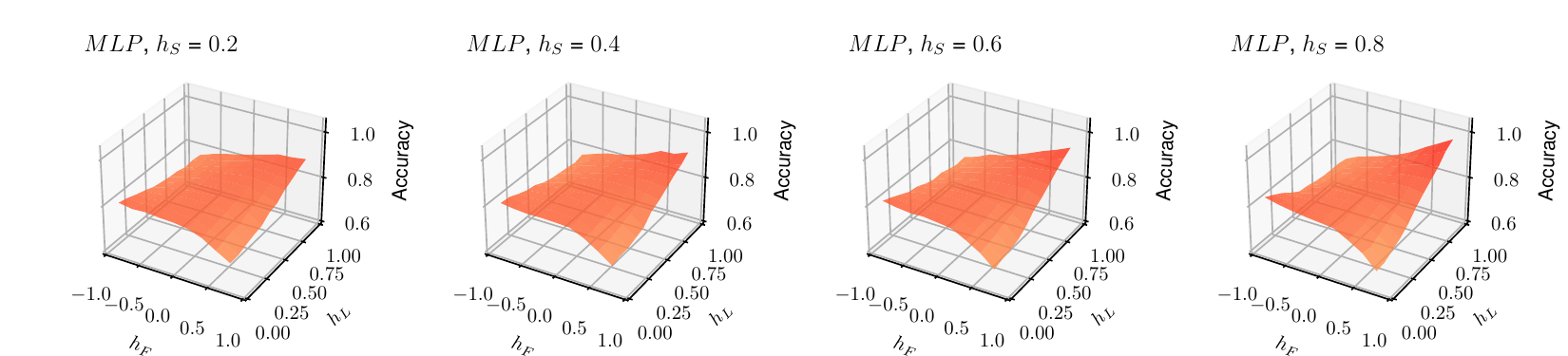}
         \caption{Model Performance on synthetic datasets}
         \label{fig:syn_hom_mlp}
     \end{subfigure}
     \hfill
    \caption{We measure the impact of label homophily $h_L$, feature homophily $h_F$, and structural homophily $h_S$ through numerical results of Tri-Hom $\mathcal{J}_h^{\neg\mathcal{G}}$ and simulation results of the node classification accuracy with MLP on synthetic datasets.}\label{fig:impact_2d_plot_mlp}
\end{figure}
We show the numerical results of Tri-Hom $\mathcal{J}_h^{\neg\mathcal{G}}$ for graph-agnostic models $\mathcal{M}^{\neg\mathcal{G}}$ in Figure \ref{fig:numerical_hom_obj_mlp}, where each subfigure is a slicer of $h_S$ that visualizes the influences of $h_L$ and $h_F$ on $\mathcal{J}_h^{\mathcal{G}}$. For the impact of $h_L$, $h_F$, and $h_S$ individually, we can get the same conclusions as in Theorem 3.1, 3.2, and 3.3. To validate our theoretical results of $\mathcal{M}^{\neg\mathcal{G}}$ in a more general case, we further show the impact of three types of homophily with MLP on synthetic datasets in Figure \ref{fig:syn_hom_mlp}. The results correlate well with our numerical results, showing the effectiveness of $\mathcal{J}_h^{\neg\mathcal{G}}$ in measuring the performance of $\mathcal{M}^{\neg\mathcal{G}}$ under the influences of three types of homophily.

\subsection{Numerical Results of Tri-Hom with Three types of Homophily}\label{apd:hom_numerical_3d}

\begin{figure}[h]
    \centering
  \subfloat[Graph-agnostic Models]{%
       \includegraphics[width=0.5\textwidth]{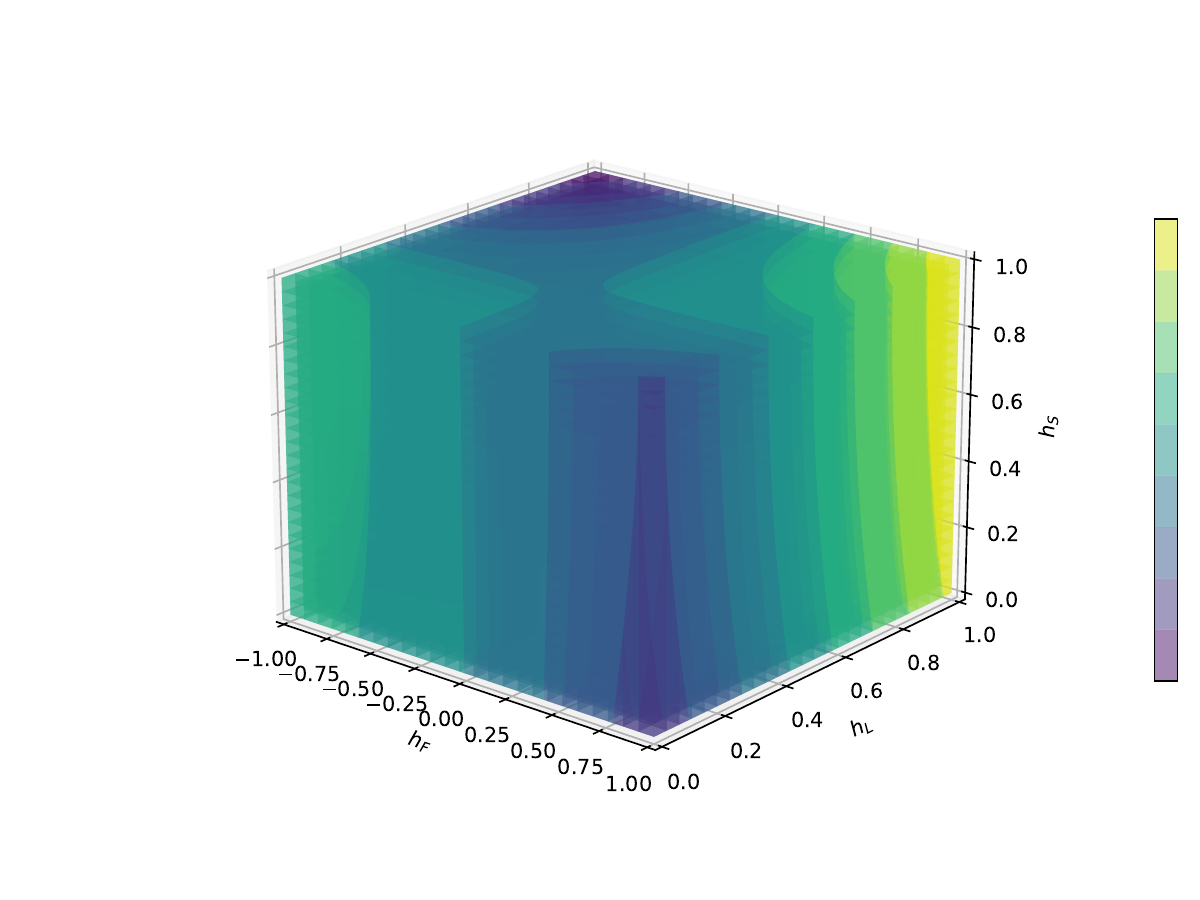}}
    \hfill
  \subfloat[Graph-aware Models]{%
        \includegraphics[width=0.5\textwidth]{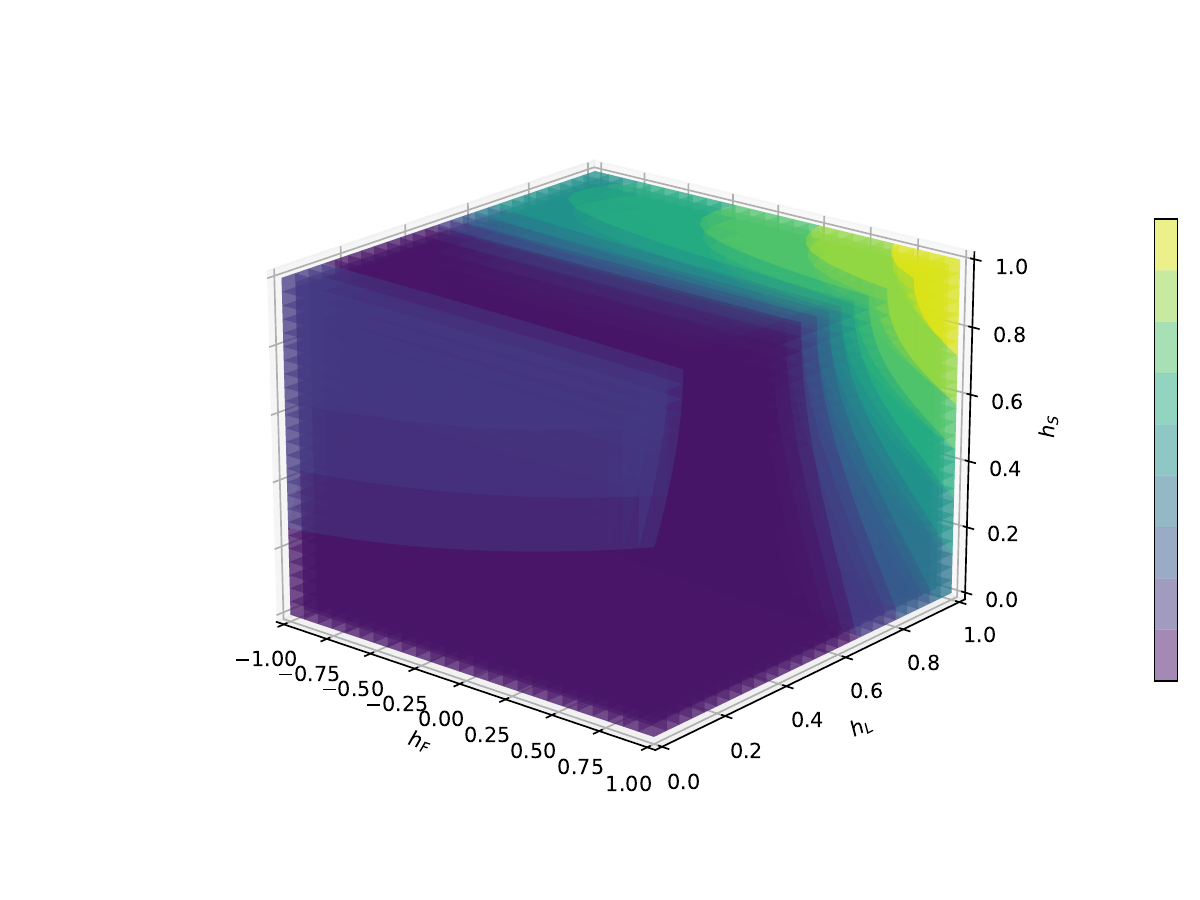}}
    \hfill
    \caption{The Influences of label homophily $h_L$, structural homophily $h_S$, and feature homophily $h_F$ to graph-agnostic models and graph-aware models}\label{fig:hom_numerical_3d}
\end{figure}

We show how the label homophily $h_L$, structural homophily $h_S$, and feature homophily $h_F$ collectively influence the $\mathcal{J}_h^{\neg\mathcal{G}}$ and $\mathcal{J}_h^{\mathcal{G}}$ in Figure \ref{fig:hom_numerical_3d} in a more general case, where the x-axis, y-axis, and z-axis denotes $h_L$, $h_F$, and $h_N$. A lighter color (close to yellow) indicates a larger $\mathcal{J}_h^{\neg\mathcal{G}}$ or $\mathcal{J}_h^{\mathcal{G}}$, which also implies a better model performance. These numerical results align well with our theoretical results of the individual impact on $\mathcal{J}_h^{\neg\mathcal{G}}$ or $\mathcal{J}_h^{\mathcal{G}}$ with respect to $h_L$, $h_S$, or $h_F$. Furthermore, by observing the gradients in Figure \ref{fig:hom_numerical_3d}, we can know how these three homophily metrics influence the model performance collectively. 

\subsection{Influences of A Single Type of Homophily on Synthetic Datasets}\label{apd:exp_syn_single}
\begin{figure}
\centering
\includegraphics[width=.3\linewidth]{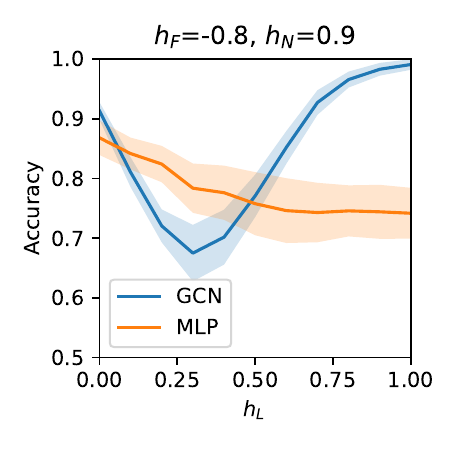}
\includegraphics[width=.3\linewidth]{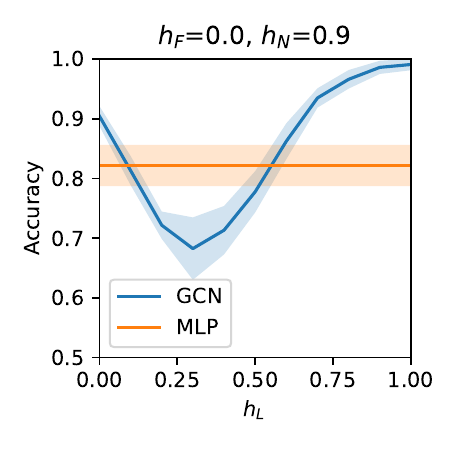}
\includegraphics[width=.3\linewidth]{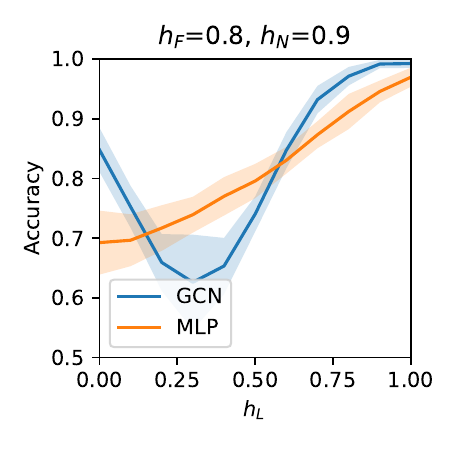}
\includegraphics[width=.3\linewidth]{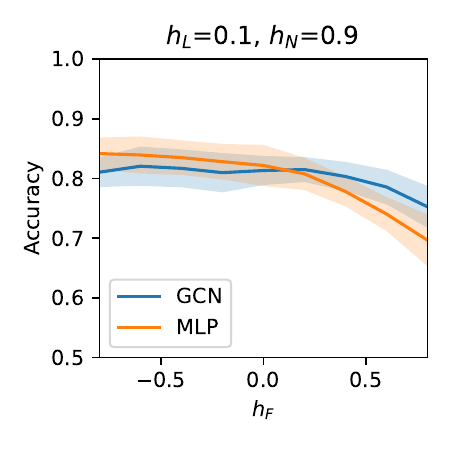}
\includegraphics[width=.3\linewidth]{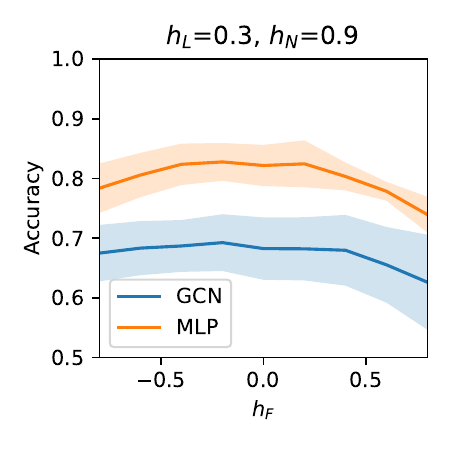}
\includegraphics[width=.3\linewidth]{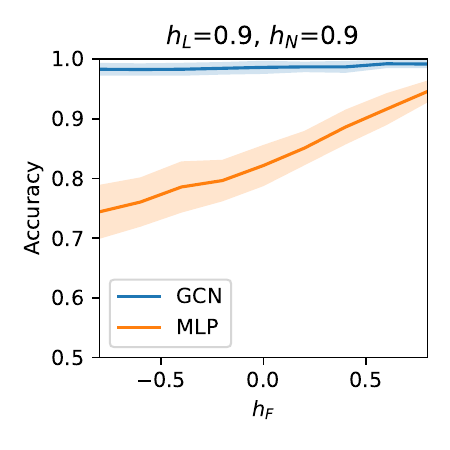}
\includegraphics[width=.3\linewidth]{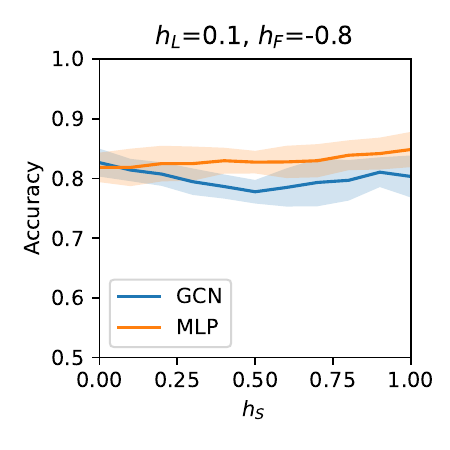}
\includegraphics[width=.3\linewidth]{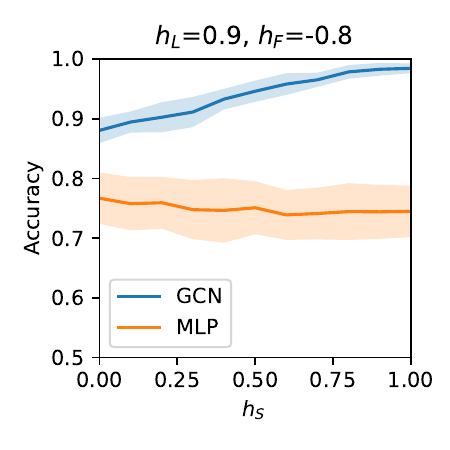}
\includegraphics[width=.3\linewidth]{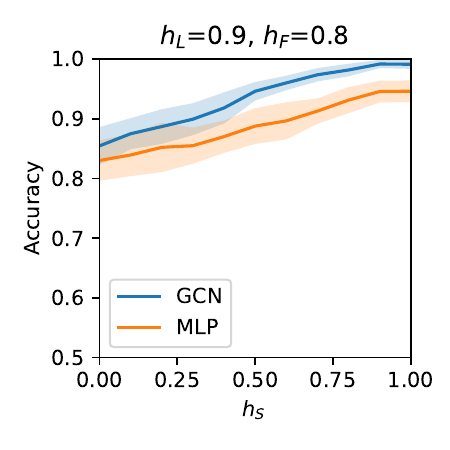}
\caption{The impact of label homophily $h_L$, feature homophily $h_F$, and structural homophily $h_S$ on the accuracy of node classification using MLP and GCN.}
\label{fig:hom_acc_cor_single}
\end{figure}

To verify our theoretical results, we show how the performances of GCN and MLP are affected by three types of homophily individually in Figure \ref{fig:hom_acc_cor_single}, which shows the influences of label homophily $h_L$, structural homophily $h_S$, and feature homophily $h_F$ on the node classification accuracy of GCN and MLP on synthetic datasets. First, the top 3 subfigures show the influences of $h_L$ under different $h_F$ and $h_S$. With the increase of $h_L$, the accuracy of GCN first decreases and then increases with a pitfall on a medium-level of homophily, and the accuracy of MLP is dependent on the sign of $h_F$, leading to the same results as in Theorem 2.1 and 3.1. Second, the medium 3 subfigures show the influences of $h_F$. With the increases of $h_F$, the accuracy of both of the GCN and MLP will decreases with a low $h_L$, increases with a high $h_L$, and increases first and then decreases with a medium $h_L$. The result also aligns well with Theorem 2.3 and 3.3. Last, the bottom 3 subfigures show the influences of $h_S$. Even if the influences under different $h_L$ or $h_F$ would slightly vary, an increase of $h_S$ generally improves the performance of both the GCN and MLP, aligning well with Theorem 2.2 and 3.2.

\subsection{Visualization of Real-world Dataset with Three Types of Homophily}\label{apd:fig_visual_3h_gcn_realworld}
\begin{figure}
    \centering
    \includegraphics[width=0.9\linewidth]{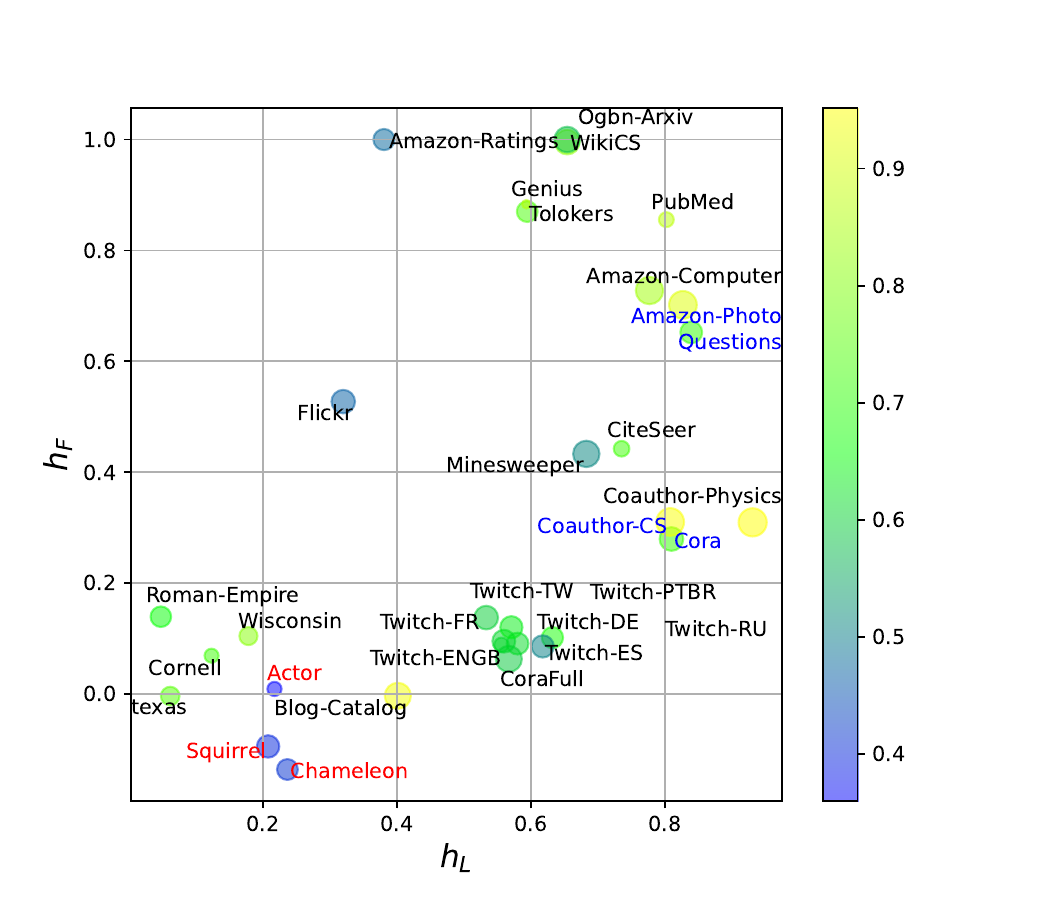}
    \caption{Label, feature, and structural homophily metrics on real-world datasets are shown as the x-axis, y-axis, and the size of the scatter respectively. The classification performance of GCN is denoted by the color of the scatters.}
    \label{fig:correlation_hom_acc}
\end{figure}

To better visualize the impact of $h_L$, $h_S$, and $h_F$ to graph-aware models $\mathcal{M}^\mathcal{G}$ on real-world datasets, we show the performance of GCN on all the datasets with these homophily metrics in Figure \ref{fig:correlation_hom_acc}, where $h_L$, $h_F$, and $h_S$ are shown as the x-axis, y-axis, and the size of the scatter respectively. Generally, the correlation of three homophily metrics with the performance of GCN is similar to our theoretical results in Figure \ref{fig:numerical_hom_obj_gnn}. With the increase of $h_L$, the performance of GCN decreases first and then increases, with a pitfall (as shown by datasets of \textcolor{red}{Actor}, \textcolor{red}{Squirrel}, and \textcolor{red}{Chameleon}). We also see that a higher $h_S$ leads to a better performance when we fix $h_F$ and $h_L$(as shown by datasets of \textcolor{blue}{Question} with \textcolor{blue}{Amazon-photo}, and \textcolor{blue}{Cora} with \textcolor{blue}{Coauthor-CS}). As for the $h_F$, since its influence is not so obvious compared with $h_L$ and $h_S$ as shown in Figure \ref{fig:hom_acc_cor_single} and the scarcity of real-world datasets, it is hard to see its influence in real-world datasets.

\subsection{Influences of Class-wised Structural Homophily on Real-world Datasets}\label{apd:fig_hs_classwise}
\begin{figure}
    \centering
    \includegraphics[width=1\linewidth]{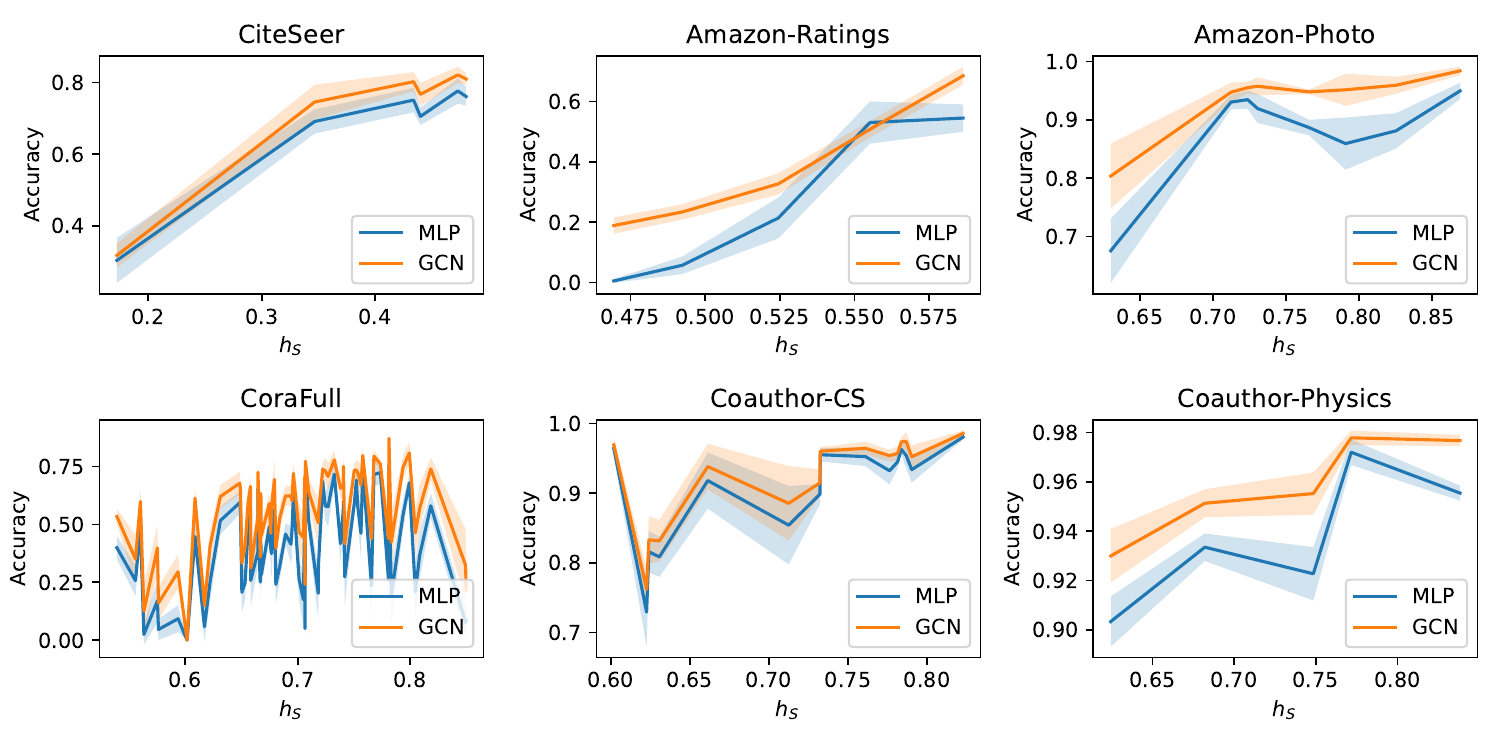}
    \caption{The influences of structural homophily to both the GCN and MLP for each class on real-world datasets.}
    \label{fig:correlation_h_N}
\end{figure}

As shown in Figure \ref{fig:correlation_h_N}, we investigate more nuanced influences of structural homophily $h_S$ without the interference of $h_L$ and $h_F$ on real-world datasets with respect to each class. Specifically, we calculate node classification accuracy with respect to each class inside one dataset and show the alignment of the class-wised accuracy and class-wised structural homophily $h_S$. The results show for some of the datasets such as Citeseer, Amazon-ratings, and Coauthor-physics, we can clearly observe the accuracy increases with $h_S$ while for datasets such as Corafull and Coauthor-CS, we can only see the general tendency of alignment with noises. We speculate that for Corafull and Coauthor-CS, the $h_S$ is not as uniform as other datasets for the nodes in one class, leading to the noises. 

\subsection{Correlation of Metrics with Performance Gap of GNNs and MLP}\label{apd:sec_cor_acc_diff}
\input{tables/correlation_pearson_difference}

To investigate when GNNs are better than MLP, we show the differences in the performance between GNNs and MLP in Figure \ref{tab:pearson_acc_diff}. The results indicate that, among the statistic-based homophily metrics, the structural homophily $h_S$ shows a strong correlation with the differences in the performance between GNNs and MLP. This is because $h_S$ measure the consistency of the structural information of nodes within the same class, thereby reflecting the differences in the performance between GNNs and MLP. We also observe that the $h_{KR}$ has the highest correlation with differences, confirming the effectiveness of the classifier-based metrics in measuring the differences in the performance between GNNs and MLP~\cite{when_do_graph_help}.

\subsection{Other Types of Correlation}\label{apd:exp_other_cor}

In our experiments on real-world datasets, we explore the correlation between all the metrics and model performance. In addition to the widely used Pearson correlation, we employ Kendall’s Tau rank correlation to assess these relationships. Let $\b{x}$ and $\b{y}$ be two observed variables with $\b{x},\b{y}\in\mathbb{R}^N$, the Pearson correlation $\rho$ can be calculated as

\begin{equation}
    \rho=\frac{\sum_{i=1}^N (x_i-\bar{x})(y_i-\bar{y})}{\sqrt{\sum_{i=1}^N(x_i-\bar{x})^2 \sum_{i=1}^N(y_i-\bar{y})^2}}
\end{equation}
where $\bar{x}$ and $\bar{y}$ are the means of $\b{x}$ and $\b{y}$ respectively.

Kendall’s Tau rank correlation~\cite{cor_kendall} $\tau$ can be calculated as
\begin{equation}
    \tau = \frac{\abs{\text{concordant pairs}}-\abs{\text{discordant pairs}}}{\frac{1}{2}N(N-1)}
\end{equation}
where concordant pairs occur when the ranks of both variables agree, while discordant pairs occur when they disagree.

% \yilun{comment Spearman rank correlation there...}

% Spearman rank correlation $\rho_{KT}$ can be calculated as
% \begin{equation}
%     \rho_{KT} = 1-\frac{6\sum d_i^2}{N(N^2-1)}
% \end{equation}
% where $d_o$ refers to the differences in ranks between corresponding data points.

Pearson correlation measures linear correlation between two variables, which is widely used and easy to interpret. Kendall’s Tau rank correlation measures the similarity ranking between two variables, which has no assumptions of the data distribution and are robust to outliers. 

\input{tables/correlation_kendalltau}

The correlation of all the metrics with model performance measured by Kendall’s Tau rank correlation is shown in Table \ref{tab:cor_kendalltau}, which is similar as the results in Pearson correlation. Our Tri-Hom $\mathcal{J}_h^{\mathcal{G}}$ still shows the highest correlation with GNNs performance compared with other types of metrics, which confirms the necessity of disentangling graph homophily.

\subsection{Correlation of Homophily Metrics}\label{apd:cor_all_hom}
\begin{figure}
    \centering
    \includegraphics[width=1\linewidth]{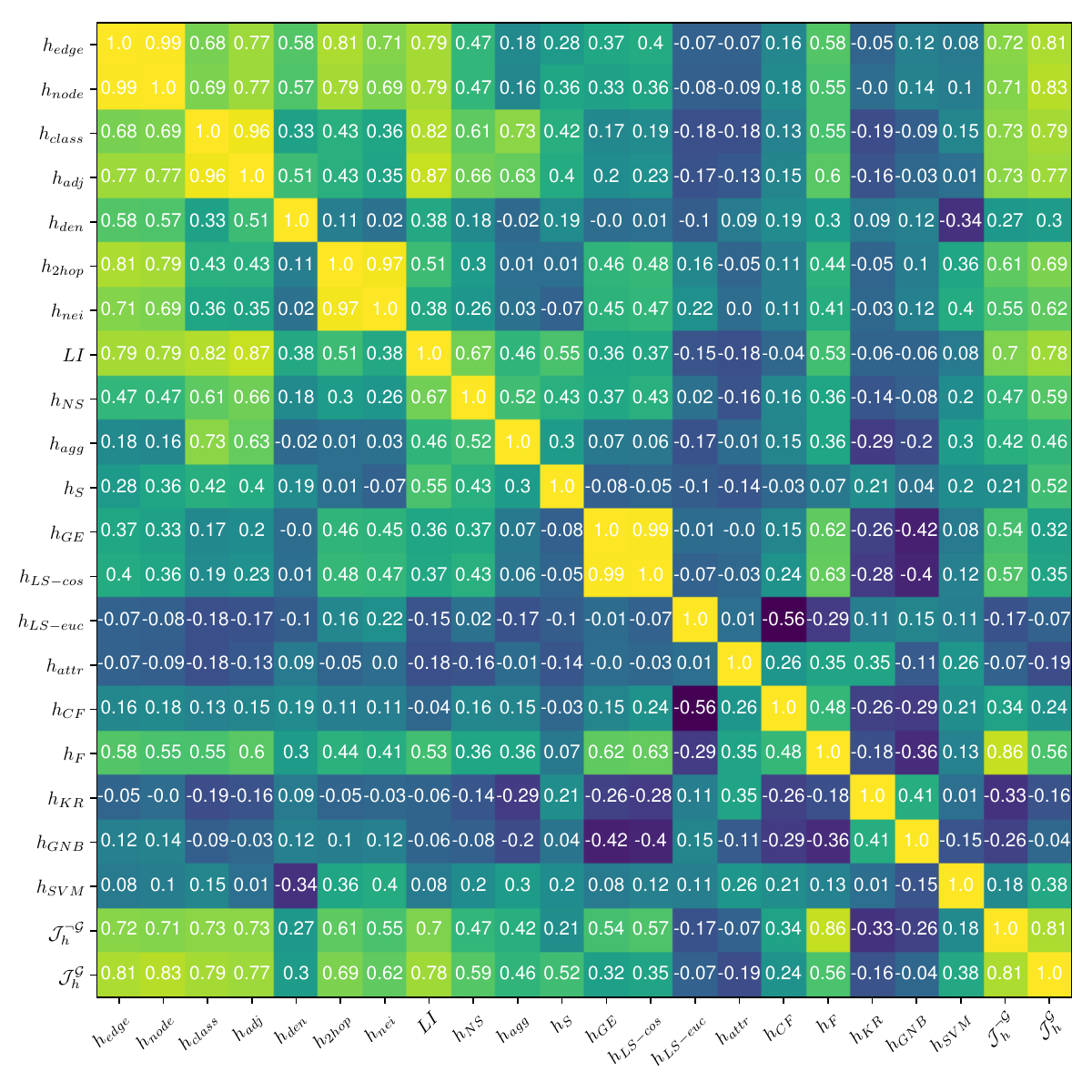}
    \caption{Pearson correlation between all the metrics. Each cell refers to the correlation between two metrics measured on 31 real-world datasets.}
    \label{fig:hom_correlation}
\end{figure}

To investigate the similarity of the information contained in graph homophily metrics and GNNs performance metrics, we show the Pearson correlation of these metrics on real-world datasets in Figure \ref{fig:hom_correlation}. We can see a high correlation among the homophily on label aspect (including $h_{edge}$, $h_{node}$, $h_{class}$, $h_{adj}$, $h_{den}$, $h_{2hop}$, and $h_{nei}$). These metrics measure the label consistency across the graph topology, sharing a similar characteristic. This also holds for the homophily on the structural aspect (including $LI$, $h_{NC}$, $h_{agg}$, and $h_S$), since these metrics measure how informative the neighbors are for the node labels. However, the Pearson correlation among the homophily on feature aspect ($h_{GE}$, $h_{LS-cos}$, $h_{LS-cos}$, $h_{attr}$, $h_{CF}$, and $h_F$) is low. We speculate that this is because different types of feature aspect ($h_{GE}$, $h_{CF}$, and $h_F$) or different similarity measurements ($h_{LS-cos}$ and $h_{LS-cos}$) could vary a lot for real-world datasets.

%% file: tables/dataset_statistics_basics.tex
\begin{table}[htbp]
  \centering
  \caption{Statistics on real-world datasets}\label{tab:real_world_datasets_statistics}
\resizebox{1\hsize}{!}{
\begin{tabular}{ccccccc}\toprule
Dataset&\#Nodes&\#Edges&\#Features&\#Classes&Average Degrees&Spectral Radius\\ \toprule
Roman-Empire&22,662&32,927&300&18&1.45&4.58\\
Amazon-Ratings&24,492&93,050&300&5&3.80&20.39\\
Minesweeper&10,000&39,402&7&2&3.94&7.99\\
Tolokers&11,758&519,000&10&2&44.14&392.36\\
Questions&48,921&153,540&301&2&3.14&95.31\\
Squirrel&2,223&46,998&2,089&5&21.14&206.02\\
Chameleon&890&8,854&2,325&5&9.95&78.05\\
Actor&7,600&26,659&932&5&3.51&37.37\\
Texas&183&279&1,703&5&1.52&10.98\\
Cornell&183&277&1,703&5&1.51&10.08\\
Wisconsin&251&450&1,703&5&1.79&11.88\\
Cora&2,708&10,556&1,433&7&3.90&14.39\\
CoraFull&19,793&126,842&8,710&70&6.41&25.63\\
CiteSeer&3,327&9,228&3,703&6&2.77&13.74\\
PubMed&19,717&88,651&500&3&4.50&23.24\\
Flickr&89,250&899,756&500&7&10.08&83.06\\
Amazon-Photo&7,650&238,162&745&8&31.13&122.54\\
Amazon-Computer&13,752&491,722&767&10&35.76&169.71\\
Coauthor-CS&18,333&163,788&6,805&15&8.93&24.60\\
Coauthor-Physics&34,493&495,924&8,415&5&14.38&51.18\\
WikiCS&11,701&431,726&300&10&36.90&149.77\\
Blog-Catalog&5,196&343,486&8,189&6&66.11&114.01\\
Ogbn-Arxiv&169,343&1,166,243&128&40&6.89&180.27\\
Genius&421,961&984,979&12&2&2.33&212.82\\
Twitch-DE&9,498&153,138&2,514&2&16.12&149.92\\
Twitch-ENGB&7,126&35,324&2,545&2&4.96&43.41\\
Twitch-ES&4,648&59,382&2,148&2&12.78&89.82\\
Twitch-FR&6,549&112,666&2,275&2&17.20&130.24\\
Twitch-PTBR&1,912&31,299&1,449&2&16.37&99.09\\
Twitch-RU&4,385&37,304&2,224&2&8.51&76.26\\
Twitch-TW&2,772&63,462&1,288&2&22.89&143.43\\
\bottomrule
\end{tabular}
}
\end{table}%

%% file: tables/dataset_statistics_hom.tex
\begin{table}[htbp]
  \centering
  \caption{Graph homophily metrics and graph performance metrics on real-world datasets}\label{tab:real_world_datasets_metrics}
\resizebox{1\hsize}{!}{
\begin{tabular}{cccccccccccc}\toprule
Dataset&$h_{edge}$&$h_{node}$&$h_{class}$&$h_{adj}$&$h_{den}$&$h_{2hop}$&$h_{nei}$&$LI$&$h_{NS}$&$h_{agg}$&$h_S$\\ \toprule
Roman-Empire&0.0469&0.0460&0.0208&-0.0497&0.4994&0.0750&0.2925&-0.6554&3.2151&0.5874&0.5271\\
Amazon-Ratings&0.3804&0.3757&0.1266&0.1386&0.5000&0.3686&0.5132&-0.1462&1.7253&0.4191&0.5256\\
Minesweeper&0.6828&0.6829&0.0094&0.0095&0.5000&0.6809&0.7999&0.0200&2.1402&0.3937&0.7070\\
Tolokers&0.5945&0.6344&0.1801&0.0944&0.4984&0.6510&0.7390&0.0047&2.1232&-0.3217&0.5451\\
Questions&0.8616&0.8980&0.0790&0.1552&0.4999&0.9097&0.9342&0.0007&23.2975&-0.5480&0.5576\\
Squirrel&0.2072&0.1905&0.0398&0.0087&0.4907&0.2239&0.3191&-0.0316&1.2125&-0.1498&0.6030\\
Chameleon&0.2361&0.2441&0.0444&0.0276&0.4917&0.2739&0.4141&-0.0620&1.1318&0.1685&0.5367\\
Actor&0.2167&0.2199&0.0064&0.0024&0.4999&0.2135&0.3091&-0.2189&1.0833&-0.1476&0.3841\\
Texas&0.0609&0.0567&0.0000&-0.2822&0.4444&0.5429&0.6946&-0.2338&4.3899&0.4863&0.5158\\
Cornell&0.1227&0.1110&0.0383&-0.2526&0.4828&0.3989&0.5819&-0.4002&2.5518&0.4098&0.3676\\
Wisconsin&0.1778&0.1552&0.0461&-0.1909&0.4838&0.4239&0.5959&-0.2842&3.0825&0.6574&0.4687\\
Cora&0.8100&0.8252&0.7657&0.7717&0.5015&0.7407&0.8118&0.3240&14.9687&0.8789&0.6164\\
CoraFull&0.5670&0.5861&0.4959&0.5552&0.5012&0.4383&0.5661&0.2895&44.7329&0.8626&0.6980\\
CiteSeer&0.7391&0.7166&0.6267&0.6673&0.5010&0.6849&0.8389&-0.0571&6.4521&0.6495&0.3909\\
PubMed&0.8024&0.7924&0.6641&0.6836&0.5002&0.7435&0.8179&0.2525&3.9050&0.6707&0.3792\\
Flickr&0.3809&0.3221&0.0698&0.1758&0.5000&0.3362&0.4657&0.0130&2.9056&0.1328&0.6086\\
Amazon-Photo&0.8272&0.8493&0.7722&0.7845&0.5036&0.6576&0.7112&0.6373&22.0358&0.9312&0.7559\\
Amazon-Computer&0.7772&0.8017&0.7002&0.6809&0.5015&0.5656&0.6439&0.4962&19.7670&0.9236&0.7628\\
Coauthor-CS&0.8081&0.8320&0.7547&0.7846&0.5008&0.6862&0.7274&0.5292&43.5786&0.9474&0.7213\\
Coauthor-Physics&0.9314&0.9153&0.8474&0.8692&0.5005&0.8369&0.8655&0.6675&26.9190&0.9424&0.7330\\
WikiCS&0.6547&0.6774&0.5675&0.5786&0.5022&0.3710&0.4306&0.3340&9.3530&0.7431&0.6366\\
Blog-Catalog&0.4011&0.3914&0.2680&0.2722&0.5029&0.2061&0.2479&0.0704&1.5237&0.5277&0.6987\\
Ogbn-Arxiv&0.6778&0.6353&0.4211&0.6158&0.5001&0.5013&0.6251&0.4535&51.7619&0.7367&0.5939\\
Genius&0.6689&0.5087&0.0229&0.1432&0.5000&0.7216&0.8078&0.0025&2.9629&0.0000&0.0931\\
Twitch-DE&0.6322&0.5958&0.1394&0.1351&0.5001&0.5431&0.6568&-0.0029&1.1644&0.2426&0.5364\\
Twitch-ENGB&0.5560&0.5452&0.0852&0.0823&0.5000&0.5235&0.6060&-0.0489&1.0328&0.1485&0.3853\\
Twitch-ES&0.5800&0.6186&0.1468&0.1067&0.4995&0.5794&0.6712&-0.0051&1.5146&-0.4148&0.5700\\
Twitch-FR&0.5595&0.5739&0.0855&0.0856&0.5000&0.5349&0.6105&-0.0081&1.1885&-0.2628&0.6009\\
Twitch-PTBR&0.5708&0.5949&0.1196&0.1082&0.4996&0.5418&0.6213&-0.0047&1.2712&-0.3086&0.5801\\
Twitch-RU&0.6176&0.6383&0.0424&0.0296&0.4998&0.6276&0.7470&-0.0086&1.7610&-0.5097&0.5698\\
Twitch-TW&0.5332&0.5500&0.0339&0.0331&0.5000&0.5212&0.5884&-0.0101&1.1133&-0.2150&0.6248\\
\bottomrule
\end{tabular}
}
\resizebox{1\hsize}{!}{
\begin{tabular}{cccccccccccc}\toprule
Dataset&$h_{GE}$&$h_{LS-cos}$&$h_{LS-euc}$&$h_{attr}$&$h_{CF}$&$h_F$&$h_{KR1}$&$h_{GNB}$&$h_{SVM}$&$\mathcal{J}^{\neg\mathcal{G}}$&$\mathcal{J}^{\mathcal{G}}$\\ \toprule
Roman-Empire&0.0257&0.0255&-8.0595&0.0585&0.7781&0.1300&0.0000&0.0000&0.0000&0.9887&0.1519\\
Amazon-Ratings&0.1146&0.1192&-24.3967&0.6159&0.5976&1.0000&0.9992&0.0253&1.0000&1.0079&0.0080\\
Minesweeper&0.3330&0.3332&-0.9430&0.1856&0.0930&0.5000&1.0000&1.0000&1.0000&1.0533&0.6042\\
Tolokers&0.8093&0.7705&-0.2957&0.1099&0.2228&1.0000&0.9982&0.0000&0.0000&1.0768&0.2814\\
Questions&0.4267&0.5544&-0.1423&0.0050&0.9060&0.7100&0.0000&0.6304&1.0000&1.1122&0.7210\\
Squirrel&0.0138&0.0219&-0.5155&0.0007&-0.4113&0.0000&1.0000&1.0000&0.0000&1.0000&0.0481\\
Chameleon&0.0141&0.0135&-0.5673&0.0006&0.0819&0.0000&0.9893&1.0000&0.0000&1.0000&0.0272\\
Actor&0.1594&0.1566&-0.6630&0.0011&0.0229&0.0000&0.0000&0.0000&0.0000&1.0000&0.0289\\
Texas&0.3487&0.3448&-0.1400&0.0008&-0.0266&0.0000&0.0000&0.0000&1.0000&1.0000&0.1415\\
Cornell&0.3322&0.3391&-0.1455&0.0009&0.1393&0.0000&0.0000&0.0000&1.0000&1.0000&0.0782\\
Wisconsin&0.3414&0.3373&-0.1444&0.0009&0.2975&0.0000&0.0000&0.0000&1.0000&1.0000&0.0553\\
Cora&0.1677&0.1780&-0.3338&0.0066&0.1410&0.1500&1.0000&1.0000&1.0000&1.0216&0.6881\\
CoraFull&0.1447&0.1604&-0.2263&0.0037&0.0052&0.0000&1.0000&1.0000&0.0000&1.0000&0.3069\\
CiteSeer&0.1906&0.2229&-0.2208&0.0067&0.1487&0.7700&0.0000&1.0000&0.0000&1.0935&0.4090\\
PubMed&0.2719&0.2658&-0.2384&0.0099&0.2326&1.0000&0.0000&0.0000&0.0000&1.1443&0.5139\\
Flickr&0.3815&0.4084&-0.0950&0.0027&0.0924&0.3000&0.0000&0.0000&0.0000&0.9982&0.0007\\
Amazon-Photo&0.4878&0.4450&-0.0843&0.0016&0.1062&0.9100&0.0000&0.0000&0.0034&1.1435&0.9500\\
Amazon-Computer&0.4897&0.4402&-0.0829&0.0015&0.0337&1.0000&0.0000&0.0000&1.0000&1.1413&0.8955\\
Coauthor-CS&0.2944&0.3194&-0.1763&0.0056&0.2355&0.4400&0.0000&0.0000&1.0000&1.0644&0.8288\\
Coauthor-Physics&0.3513&0.3723&-0.2160&0.0069&0.3741&0.8800&0.0000&0.0000&1.0000&1.1704&1.0292\\
WikiCS&0.3342&0.3916&-277.1482&-0.0209&0.9998&1.0000&0.0000&0.0000&0.0000&1.0980&0.4899\\
Blog-Catalog&0.1223&0.1228&-0.1818&0.0001&-0.0430&0.0000&0.0000&0.0000&0.0000&1.0000&0.0249\\
Ogbn-Arxiv&0.8389&0.8649&-0.0546&0.0028&0.2352&1.0000&0.0057&0.0000&0.0000&1.0974&0.4470\\
Genius&0.6656&0.5909&-0.2162&0.1585&0.1898&0.7500&0.0001&0.0000&0.0000&1.0521&0.1059\\
Twitch-DE&0.1961&0.1919&-0.2860&0.0006&0.0782&0.0000&0.0390&0.9831&0.0000&1.0000&0.3217\\
Twitch-ENGB&0.2101&0.2001&-0.2836&0.0007&0.0232&0.0000&0.0571&0.9191&0.0231&1.0000&0.1350\\
Twitch-ES&0.2039&0.1976&-0.2939&0.0007&0.0278&0.0000&0.3051&0.9999&0.0001&1.0000&0.2505\\
Twitch-FR&0.2251&0.2117&-0.2850&0.0007&0.0716&0.0000&0.9968&0.2459&0.0000&1.0000&0.2304\\
Twitch-PTBR&0.2135&0.2030&-0.2860&0.0009&0.0911&0.0000&0.3399&0.9971&0.0009&1.0000&0.2396\\
Twitch-RU&0.1925&0.1902&-0.2829&0.0006&0.0256&0.0000&0.9293&0.0307&0.4450&1.0000&0.3195\\
Twitch-TW&0.1893&0.1914&-0.3037&0.0010&0.0271&0.0000&0.3110&0.0705&0.0001&1.0000&0.1935\\
\bottomrule
\end{tabular}
}
\end{table}%

%% file: tables/node_classification_acc.tex
\begin{table}[htbp]
  \centering
  \caption{Node classification performance on real-world datasets.}\label{tab:node_cls_acc_realworld}
    \begin{tabular}{ccccc}
    \toprule
     \multirow{2}{*}{Dataset} & {MLP} &{GCN} & {GraphSage} & {GAT} \\
    \cmidrule{2-5}
    & Acc$\pm$ Std & Acc$\pm$ Std& Acc$\pm$ Std & Acc$\pm$ Std \\ \toprule
Roman-Empire&{65.47$\pm$0.66}&{78.76$\pm$0.66}&{83.74$\pm$0.52}&{83.79$\pm$0.65}\\
Amazon-Ratings&{47.58$\pm$0.30}&{50.67$\pm$0.75}&{53.21$\pm$0.66}&{51.72$\pm$0.59}\\
Minesweeper&{51.06$\pm$0.92}&{90.16$\pm$0.59}&{90.74$\pm$0.57}&{90.55$\pm$0.67}\\
Tolokers&{73.52$\pm$0.90}&{84.55$\pm$0.65}&{83.21$\pm$0.49}&{84.28$\pm$0.58}\\
Questions&{71.36$\pm$0.96}&{76.35$\pm$0.93}&{76.84$\pm$0.97}&{77.83$\pm$0.75}\\
Squirrel&{40.12$\pm$1.55}&{40.24$\pm$1.58}&{40.37$\pm$1.83}&{40.06$\pm$3.12}\\
Chameleon&{41.47$\pm$3.38}&{43.10$\pm$2.73}&{41.88$\pm$4.32}&{41.70$\pm$5.99}\\
Actor&{35.97$\pm$1.08}&{35.10$\pm$1.05}&{35.52$\pm$0.92}&{35.51$\pm$0.88}\\
texas&{75.26$\pm$4.61}&{72.28$\pm$7.41}&{79.60$\pm$5.47}&{74.72$\pm$4.27}\\
Cornell&{72.43$\pm$4.90}&{67.84$\pm$5.90}&{73.24$\pm$4.31}&{69.19$\pm$5.73}\\
Wisconsin&{80.98$\pm$4.81}&{78.24$\pm$5.43}&{83.53$\pm$3.94}&{79.41$\pm$6.49}\\
Cora&{71.57$\pm$1.56}&{86.36$\pm$0.93}&{87.10$\pm$1.43}&{87.26$\pm$1.51}\\
CoraFull&{59.18$\pm$0.68}&{68.74$\pm$0.76}&{69.06$\pm$0.71}&{70.12$\pm$0.69}\\
CiteSeer&{72.36$\pm$1.30}&{76.44$\pm$0.79}&{77.07$\pm$1.09}&{76.96$\pm$1.12}\\
PubMed&{87.26$\pm$0.24}&{89.12$\pm$0.38}&{89.30$\pm$0.28}&{89.26$\pm$0.44}\\
Flickr&{46.83$\pm$0.33}&{52.80$\pm$0.43}&{52.41$\pm$0.35}&{53.52$\pm$0.37}\\
Amazon-Photo&{91.28$\pm$0.51}&{95.16$\pm$0.54}&{95.72$\pm$0.35}&{95.59$\pm$0.24}\\
Amazon-Computer&{83.99$\pm$0.59}&{91.58$\pm$0.61}&{91.17$\pm$0.56}&{91.75$\pm$0.54}\\
Coauthor-CS&{94.54$\pm$0.32}&{95.68$\pm$0.23}&{95.53$\pm$0.29}&{95.58$\pm$0.24}\\
Coauthor-Physics&{95.18$\pm$0.27}&{96.93$\pm$0.25}&{96.83$\pm$0.30}&{96.73$\pm$0.28}\\
WikiCS&{81.38$\pm$0.51}&{85.20$\pm$0.28}&{85.71$\pm$0.47}&{85.89$\pm$0.48}\\
Blog-Catalog&{93.95$\pm$0.70}&{96.07$\pm$0.57}&{96.49$\pm$0.64}&{96.02$\pm$0.54}\\
Ogbn-Arxiv&{58.02$\pm$0.30}&{73.91$\pm$0.21}&{73.39$\pm$0.25}&{73.82$\pm$0.14}\\
Genius&{86.53$\pm$0.07}&{90.59$\pm$0.22}&{90.99$\pm$0.17}&{81.92$\pm$4.75}\\
Twitch-DE&{67.65$\pm$0.93}&{72.26$\pm$1.00}&{70.14$\pm$0.85}&{72.46$\pm$1.14}\\
Twitch-ENGB&{61.47$\pm$1.11}&{62.89$\pm$0.95}&{61.96$\pm$1.55}&{62.50$\pm$1.08}\\
Twitch-ES&{61.59$\pm$1.84}&{65.58$\pm$1.64}&{62.80$\pm$1.58}&{66.71$\pm$2.19}\\
Twitch-FR&{60.93$\pm$1.64}&{64.84$\pm$1.79}&{61.82$\pm$1.94}&{64.77$\pm$2.29}\\
Twitch-PTBR&{63.61$\pm$2.74}&{66.90$\pm$1.58}&{65.05$\pm$1.98}&{67.60$\pm$1.77}\\
Twitch-RU&{50.34$\pm$1.94}&{53.74$\pm$3.21}&{50.88$\pm$1.51}&{53.00$\pm$2.50}\\
Twitch-TW&{59.84$\pm$2.16}&{61.89$\pm$2.25}&{60.68$\pm$1.79}&{63.73$\pm$1.84}\\
    \bottomrule
    \end{tabular}%
\end{table}%

%% file: tables/correlation_pearson_difference.tex
\begin{table}[htbp]
  \centering
  \caption{Pearson correlation with p-value of all the metrics with the performance differences of GNNs with MLP of node classification on real-world datasets.}
  % \resizebox{1\hsize}{!}{
    \begin{tabular}{lccccccr}
    \toprule
     \multirow{2}{*}{Metric} & \multicolumn{2}{c}{GCN-MLP} & \multicolumn{2}{c}{GraphSage-MLP} & \multicolumn{2}{c}{GAT-MLP} & \multirow{2}{*}{Rank} \\
    \cmidrule{2-7}
    & Cor. & p-value& Cor. & p-value& Cor. & p-value & \\
    \toprule
$h_{edge}$& 0.2937& 0.1087& 0.1189& 0.5242& 0.2073& 0.2632& 8.00\\
$h_{node}$& 0.2930& 0.1097& 0.1155& 0.5362& 0.2316& 0.2099& 8.33\\
$h_{class}$& 0.0454& 0.8084& 0.0075& 0.9678& 0.0459& 0.8064& 19.67\\
$h_{adj}$& 0.1420& 0.4461& 0.0500& 0.7896& 0.1101& 0.5555& 16.00\\
$h_{den}$&\textcolor{blue}{0.3164}& 0.0829& 0.1096& 0.5572& 0.2537& 0.1685& 7.33\\
$h_{2hop}$& 0.1722& 0.3543& 0.0925& 0.6208& 0.0932& 0.6178& 15.67\\
$h_{nei}$& 0.2055& 0.2675& 0.1615& 0.3854& 0.1367& 0.4632& 11.33\\
$LI$& 0.1338& 0.4729& 0.0012& 0.9948& 0.0674& 0.7186& 18.33\\
$h_{NS}$& 0.1716& 0.3561& 0.1585& 0.3945& 0.1736& 0.3503& 11.00\\
$h_{agg}$& 0.1220& 0.5132&\textcolor{blue}{0.2497}& 0.1755& 0.1542& 0.4076& 9.67\\
$h_S$&\textcolor{purple}{0.3040}& 0.0964&\textcolor{purple}{0.2356}& 0.2020&\textcolor{red}{0.4001}& 0.0257&\textcolor{blue}{2.33}\\
$h_{GE}$& 0.2096& 0.2577& 0.2011& 0.2779& 0.1102& 0.5550& 11.00\\
$h_{LS-cos}$& 0.2130& 0.2500& 0.2080& 0.2614& 0.1299& 0.4862& 9.67\\
$h_{LS-euc}$& -0.0299& 0.8733& -0.0197& 0.9160& -0.0159& 0.9323& 21.00\\
$h_{attr}$& -0.2137& 0.2484& -0.2644& 0.1506& -0.1781& 0.3376& 22.00\\
$h_{CF}$& 0.0897& 0.6313& 0.1657& 0.3730& 0.1398& 0.4531& 13.00\\
$h_F$& 0.2338& 0.2056& 0.2224& 0.2292& 0.1701& 0.3604& 7.67\\
$h_{KR}$&\textcolor{red}{0.3582}& 0.0478&\textcolor{red}{0.2650}& 0.1497&\textcolor{blue}{0.3336}& 0.0667&\textcolor{red}{1.33}\\
$h_{GNB}$& 0.2740& 0.1357& 0.1721& 0.3546& 0.2689& 0.1435& 6.67\\
$h_{SVM}$& 0.0825& 0.6590& 0.2281& 0.2171& 0.1211& 0.5162& 12.67\\
$\mathcal{J}^{\neg\mathcal{G}}$& 0.1407& 0.4503& 0.1086& 0.5610& 0.0954& 0.6096& 16.00\\
$\mathcal{J}^{\mathcal{G}}$& 0.2903& 0.1131& 0.2333& 0.2065&\textcolor{purple}{0.2860}& 0.1188&\textcolor{purple}{4.33}\\
    \bottomrule
    \end{tabular}%
% }
  \label{tab:pearson_acc_diff}%
\end{table}%

%% file: tables/correlation_kendalltau.tex
\begin{table}[htbp]
  \centering
  \caption{Kendall’s Tau rank correlation of all the metrics with model performance of node classification on real-world datasets}
  \resizebox{1\hsize}{!}{
    \begin{tabular}{lcccccccccccccc}
    \toprule
     \multirow{2}{*}{Metric} & \multicolumn{2}{c}{MLP} & \multicolumn{2}{c}{GCN} & \multicolumn{2}{c}{GraphSage} & \multicolumn{2}{c}{GAT} & \multicolumn{2}{c}{GCN-MLP} & \multicolumn{2}{c}{GraphSage-MLP} & \multicolumn{2}{c}{GAT-MLP}\\
    \cmidrule{2-15}
    & Cor.&p-value& Cor.&p-value& Cor.&p-value& Cor.&p-value & Cor.&p-value& Cor.&p-value& Cor.&p-value \\
    \toprule
$h_{edge}$& 0.3677& 0.0033&\textcolor{blue}{0.4882}& 0.0001& 0.4452& 0.0003&\textcolor{purple}{0.4839}& 0.0001&\textcolor{blue}{0.3462}& 0.0058& 0.2129& 0.0960& 0.2731& 0.0314\\
$h_{node}$& 0.3204& 0.0110& 0.4409& 0.0004& 0.4065& 0.0011& 0.4624& 0.0002&\textcolor{purple}{0.3161}& 0.0122& 0.1828& 0.1545&\textcolor{blue}{0.3118}& 0.0135\\
$h_{class}$&\textcolor{blue}{0.3935}& 0.0016& 0.4194& 0.0007& 0.4022& 0.0012& 0.4237& 0.0006& 0.1570& 0.2231& 0.1785& 0.1647& 0.2387& 0.0611\\
$h_{adj}$&\textcolor{purple}{0.3892}& 0.0018& 0.4667& 0.0001& 0.4323& 0.0005& 0.4538& 0.0002& 0.2731& 0.0314& 0.2602& 0.0407& 0.2602& 0.0407\\
$h_{den}$& 0.2430& 0.0565& 0.3634& 0.0037& 0.3548& 0.0047& 0.3677& 0.0033& 0.2559& 0.0442& 0.2258& 0.0770& 0.2430& 0.0565\\
$h_{2hop}$&\textcolor{purple}{0.3892}& 0.0018& 0.4065& 0.0011& 0.3720& 0.0029& 0.3935& 0.0016& 0.1785& 0.1647& 0.1742& 0.1754& 0.0968& 0.4578\\
$h_{nei}$& 0.3376& 0.0073& 0.3720& 0.0029& 0.3462& 0.0058& 0.3591& 0.0042& 0.1871& 0.1448& 0.2086& 0.1031& 0.1054& 0.4178\\
$LI$& 0.3462& 0.0058&\textcolor{purple}{0.4753}& 0.0001& 0.4409& 0.0004&\textcolor{blue}{0.5054}& 0.0000&\textcolor{red}{0.3505}& 0.0052& 0.2774& 0.0287&\textcolor{red}{0.3204}& 0.0110\\
$h_{NS}$& 0.3720& 0.0029& 0.4581& 0.0002& 0.4409& 0.0004& 0.4710& 0.0001& 0.2473& 0.0521&\textcolor{red}{0.4237}& 0.0006& 0.2258& 0.0770\\
$h_{agg}$& 0.3849& 0.0020& 0.4452& 0.0003&\textcolor{blue}{0.4624}& 0.0002& 0.4495& 0.0003& 0.0022& 1.0000& 0.2129& 0.0960& 0.0323& 0.8135\\
$h_S$& 0.0667& 0.6130& 0.2387& 0.0611& 0.2129& 0.0960& 0.2774& 0.0287& 0.2172& 0.0893& 0.1613& 0.2104&\textcolor{purple}{0.3075}& 0.0149\\
$h_{GE}$& 0.3376& 0.0073& 0.3376& 0.0073& 0.3290& 0.0090& 0.3247& 0.0100& 0.1441& 0.2644& 0.2860& 0.0240& 0.0624& 0.6369\\
$h_{LS-cos}$& 0.3677& 0.0033& 0.3591& 0.0042& 0.3505& 0.0052& 0.3462& 0.0058& 0.1570& 0.2231& 0.3075& 0.0149& 0.0753& 0.5664\\
$h_{LS-euc}$& -0.2989& 0.0181& -0.2387& 0.0611& -0.2559& 0.0442& -0.2172& 0.0893& 0.0409& 0.7616& -0.1527& 0.2363& 0.0968& 0.4578\\
$h_{attr}$& -0.1183& 0.3617& -0.2903& 0.0218& -0.2817& 0.0263& -0.2688& 0.0343& -0.2688& 0.0343& -0.4366& 0.0004& -0.2387& 0.0611\\
$h_{CF}$& 0.2860& 0.0240& 0.2946& 0.0199& 0.3032& 0.0164& 0.3075& 0.0149& 0.1699& 0.1865&\textcolor{purple}{0.3118}& 0.0135& 0.1742& 0.1754\\
$h_F$& 0.2871& 0.0365& 0.3920& 0.0043& 0.3870& 0.0048& 0.3920& 0.0043& 0.3071& 0.0253&\textcolor{blue}{0.4169}& 0.0024& 0.2821& 0.0399\\
$h_{KR}$& -0.4186& 0.0010& -0.3236& 0.0107& -0.3495& 0.0059& -0.3279& 0.0097& 0.2416& 0.0568& 0.0734& 0.5631& 0.2201& 0.0828\\
$h_{GNB}$& -0.2929& 0.0222& -0.2317& 0.0704& -0.2580& 0.0440& -0.2361& 0.0653& 0.1093& 0.3935& -0.0874& 0.4948& 0.1487& 0.2458\\
$h_{SVM}$& 0.1531& 0.2364& 0.1176& 0.3631& 0.1398& 0.2797& 0.1309& 0.3114& -0.1753& 0.1752& -0.0067& 0.9589& -0.0688& 0.5948\\
$\mathcal{J}^{\neg\mathcal{G}}$&\textcolor{red}{0.4106}& 0.0024& 0.4693& 0.0005&\textcolor{purple}{0.4546}& 0.0008& 0.4790& 0.0004& 0.1369& 0.3108& 0.2004& 0.1378& 0.1027& 0.4472\\
$\mathcal{J}^{\mathcal{G}}$& 0.3290& 0.0090&\textcolor{red}{0.5097}& 0.0000&\textcolor{red}{0.4753}& 0.0001&\textcolor{red}{0.5398}& 0.0000& 0.2473& 0.0521& 0.2344& 0.0661& 0.2602& 0.0407\\
    \bottomrule
    \end{tabular}%
}
  \label{tab:cor_kendalltau}%
\end{table}%

%% file: sections/appendix_sections/ProofofTheorems.tex
\section{Proof of Theorems}\label{apd:proof_obj}
We first specify the ranges of three types of homophily and spectral radius\footnote{We consider the spectral radius from a general case in a graph. Here the specified range is further confirmed by the values of $\rho(\b{A})$ in real-world datasets as shown in Table \ref{tab:real_world_datasets_statistics}.} of an adjacency matrix, which are used for the approximation in partial derivative process:
\begin{equation}\label{eq:apd_aprrox_rho_h}
    \begin{split}
        h_L\in[0,1], \; h_S\in[0,1], \; \abs{h_F}\in(-1,1),\; \text{and} \; \rho(\b{A})\gg 1
    \end{split}
\end{equation}

\subsection{Proof of Theorem 1}\label{apd:proof_theorem1}
\begin{theorem}{1}
    In CSBM-3H, the ratio of the expectation of intra-class distance to the expectation of inter-class distance of node representations for graph-agnostic models $\mathcal{M}^{\neg\mathcal{G}}$ and graph-aware models $\mathcal{M}^{\mathcal{G}}$ is:
    \begin{equation}
        \begin{split}
            \mathcal{J}^{\neg\mathcal{G}} = (1+\mathcal{J}_{\b{N}}\mathcal{J}_h^{\neg\mathcal{G}})^{-1}
            \;\text{and}\;
            \mathcal{J}^{\mathcal{G}} = (1+\mathcal{J}_{\b{N}}\mathcal{J}_h^{\mathcal{G}})^{-1}
            \;\text{respectively,}
        \end{split}
    \end{equation}
    where $\mathcal{J}_{\b{N}}= \frac{\sum_{Y_u\neq Y_v}[2C(C-1)]^{-1}\norm{\b{\mu}_{Y_u}-\b{\mu}_{Y_v}}^2}{C^{-1}\abs{\b{\sigma}^2}}$, $\mathcal{J}_h^{\neg\mathcal{G}}=\frac{\left[1-(\frac{h_F}{\rho(\b{A})})^2(C(\frac{1-h_L}{C-1})^2+C\frac{(1-h_S)^2}{C-1}+(\frac{h_L C-1}{C-1})^2)\right]}{\left[1- (\frac{h_F}{\rho(\b{A})}) (\frac{h_L C-1}{C-1})\right]^2}$, and $\mathcal{J}_h^{\mathcal{G}}=\frac{(\frac{h_L C-1}{C-1})^2}{(C(\frac{1-h_L}{C-1})^2+\frac{(1-h_S)^2}{C-1}+(\frac{h_L C-1}{C-1})^2)}\mathcal{J}_h^{\neg\mathcal{G}}$.
\end{theorem}

\textit{Proof}. We have the objective
\begin{equation}
    \begin{split}
    \mathcal{J} &= \frac{D_{intra}(\b{H})}
                        {D_{inter}(\b{H})} 
        = \frac {\mathbb{E}_{Y_u=Y_v,\epsilon}\left[\norm{\b{H}_u-\b{H}_v}^2\right]}
                {\mathbb{E}_{Y_u\neq Y_v,\epsilon}\left[\norm{\b{H}_u-\b{H}_v}^2\right]} 
    \end{split}
\end{equation}

Then, we derive the objective for graph-agnostic models $\mathcal{M}^{\neg\mathcal{G}}$ and graph-aware models $\mathcal{M}^{\mathcal{G}}$ respectively as follows

\subsubsection{Graph-agnostic Models}\label{apd:proof_mlp_obj}
For the graph-agnostic models, we let $\b{H}_u = \b{X}_u$. The structural-agnostic feature $m$ in class $c$ is sampled from a Gaussian distribution $N_{c,m}(\mu_{c,m},\sigma^2_{c,m})$. We use $\b{N}\in\mathbb{R}^{C\times M}$ to denote the distributions of structural-aware for all the classes with mean $\b{\mu}\in\mathbb{R}^{C\times M}$ and standard deviation $\b{\sigma^2}\in\mathbb{R}^{C\times M}$ in the matrix form.

For the structural-aware feature $m$ of node $u$, we have\footnote{Instead of deriving structural-aware features from an adjacency matrix, we sample aggregated distributions of features following previous studies~\cite{when_do_graph_help,ana_nei_noise}. The binarization and symmetrization operations result in compound probability distributions, which make the analysis complex and almost unsolvable.}

\begin{equation}
    \begin{split}
        X_{u,m} &\sim \left[(\b{I}_C-\omega\b{S})^{-1}\b{N}_{:,m}\right]_{Y_u}\\
        &= \left[\sum_{k=0}^\infty (\omega\b{S})^k \b{N}_{:,m}\right]_{Y_u} \\
        &= \left[\sum_{k=0}^\infty \omega^k\left[(\frac{1-h_L}{C-1}+\epsilon)\b{1}_C+(h_L-\frac{1-h_L}{C-1})\b{I}_C\right]^k \b{N}_{:,m}\right]_{Y_u} \\
    \end{split}
\end{equation}
Next, to simplify the expression, we let $p_1=\frac{1-h_L}{C-1}+\epsilon$ and $p_0=h_L-\frac{1-h_L}{C-1}$,
\begin{equation}\label{eq:apd_result_mlp_feat_dist}
    \begin{split}
        \X_{u,m} &\sim \left[\sum_{k=0}^\infty \omega^k\left[p_1\b{1}_C+p_0\b{I}_C\right]^k \b{N}_{:,m}\right]_{Y_u} \\
        &=\left[\sum_{k=0}^\infty \omega^k \sum_{i=0}^k \binom{k}{i} (p_1\b{1}_C)^{k-i}(p_0\b{I}_C)^i\b{N}_{:,m}\right]_{Y_u} \\
        &=\left[\sum_{k=0}^\infty \omega^k \sum_{i=0}^k \binom{k}{i} (p_1)^{k-i}(p_0)^i\b{1}_C^{k-i}\b{I}_C^i\b{N}_{:,m}\right]_{Y_u} \\
        &=\left[\sum_{k=0}^\infty \omega^k \left[p_0^k\b{I}_C-p_0^kC^{-1}\b{1}_C+\sum_{i=0}^k \binom{k}{i} (p_1)^{k-i}(p_0)^iC^{k-i-1}\b{1}_C\right]\b{N}_{:,m}\right]_{Y_u} \\
        &=\left[\sum_{k=0}^\infty \omega^k \left[p_0^k\b{I}_C-p_0^kC^{-1}\b{1}_C+(Cp_1+p_0)^kC^{-1}\b{1}_C\right]\b{N}_{:,m}\right]_{Y_u} \\
        &=\left[(1-\omega p_0)^{-1}\b{I}_C\b{N}_{:,m}+\left[(1-\omega(Cp_1+p_0))^{-1}-(1-\omega p_0)^{-1}\right]C^{-1}\b{1}_C\b{N}_{:,m}\right]_{Y_u} \\
    \end{split}
\end{equation}

Then we have the feature distribution of node $u$
\begin{equation}
    \begin{split}
        X_{u,m}\sim N(\hat{\mu}_{u,m},\hat{\sigma}^2_{u,m})
    \end{split}
\end{equation}
where
\begin{equation}
    \begin{split}
        \hat{\mu}_{Y_u,m} &= (1-\omega p_0)^{-1}\mu_{Y_u,m}+C^{-1}\left[(1-\omega(Cp_1+p_0))^{-1}-(1-\omega p_0)^{-1}\right]\abs{\b{\mu}_{:,m}} \\
        \hat{\sigma}^2_{Y_u,m} &= (1-\omega^2 p_0^2)^{-1}\sigma^2_{Y_u,m}+C^{-1}\left[(1-\omega^2(Cp_1^2+p_0^2))^{-1}-(1-\omega^2 p_0^2)^{-1}\right]\abs{\b{\sigma^2}_{:,m}} \\
    \end{split}
\end{equation}

When two nodes $u$ and $v$ share same labels, \ie{$Y_u=Y_v$}, we have
\begin{equation}
    X_{u,m}-X_{v,m}\sim N(0, \hat{\sigma}^2_{Y_u,m}+ \hat{\sigma}^2_{Y_v,m})
\end{equation}

Then the expectation of intra-class distance of $\b{X}$ can be expressed as
\begin{equation}\label{eq:apd_mlp_intra_cls_d}
    \begin{split}
        \mathbb{E}_{Y_u=Y_v,\epsilon}\left[\norm{\b{X}_u-\b{X}_v}^2\right]
        &=\mathbb{E}_{Y_u=Y_v,\epsilon}\left[\sum_{m=0}^M({X}_{u,m}-{X}_{v,m})^2\right] \\
        &=\sum_{m=0}^M \mathbb{E}_{Y_u=Y_v,\epsilon}\left[({X}_{u,m}-{X}_{v,m})^2\right] \\
        &=2C^{-1}(1-\omega^2(C\mathbb{E}_{\epsilon}[p_1^2]+p_0^2))^{-1}\abs{\b{\sigma^2}}
    \end{split}
\end{equation}

When two nodes $u$ and $v$ share different labels, \ie{$Y_u\neq Y_v$}, we have
\begin{equation}
    X_{u,m}-X_{v,m}\sim N(\hat{\mu}_{Y_u,m}-\hat{\mu}_{Y_v,m}, \hat{\sigma}^2_{Y_u,m}+ \hat{\sigma}^2_{Y_v,m})
\end{equation}

Then the expectation of inter-class distance of $\b{X}$ can be expressed as
\begin{equation}\label{eq:apd_mlp_inter_cls_d}
    \begin{split}
        \mathbb{E}_{Y_u\neq Y_v,\epsilon}\left[\norm{\b{X}_u-\b{X}_v}^2\right]
        &=\mathbb{E}_{Y_u\neq Y_v,\epsilon}\left[\sum_{m=0}^M({X}_{u,m}-{X}_{v,m})^2\right] \\
        &=\sum_{m=0}^M \mathbb{E}_{Y_u\neq Y_v,\epsilon}\left[({X}_{u,m}-{X}_{v,m})^2\right] \\
        &=2C^{-1}(1-\omega^2(C\mathbb{E}_{\epsilon}[p_1^2]+p_0^2))^{-1}\abs{\b{\sigma^2}}\\
        &+[C(C-1)]^{-1}(1-\omega p_0)^{-2}\sum_{Y_u\neq Y_v}\norm{\b{\mu}_{Y_u,:}-\b{\mu}_{Y_v,:}}^2
    \end{split}
\end{equation}

Then, we come back to the objective by combining Eq. (\ref{eq:apd_mlp_intra_cls_d}) and Eq. (\ref{eq:apd_mlp_inter_cls_d})
\begin{equation}
    \begin{split}
        \mathcal{J}^{\neg\mathcal{G}} &= \frac{\mathbb{E}_{Y_u= Y_v,\epsilon}\left[\norm{\b{H}_u-\b{H}_v}^2\right]}{\mathbb{E}_{Y_u\neq Y_v,\epsilon}\left[\norm{\b{H}_u-\b{H}_v}^2\right]}\\
        &= \frac{\mathbb{E}_{Y_u= Y_v,\epsilon}\left[\norm{\b{X}_u-\b{X}_v}^2\right]}{\mathbb{E}_{Y_u\neq Y_v,\epsilon}\left[\norm{\b{X}_u-\b{X}_v}^2\right]}\\
        &=\left( 1+\frac{\sum_{Y_u\neq Y_v}[2C(C-1)]^{-1}\norm{\b{\mu}_{Y_u,:}-\b{\mu}_{Y_u,:}}^2}{C^{-1}\abs{\b{\sigma}^2}} \frac{(1-\omega^2(C\mathbb{E}_{\epsilon}[p_1^2]+p_0^2))}{(1-\omega p_0)^2} \right)^{-1}\\
    \end{split}
\end{equation}

Then we take back $p_0$, $p_1$, and $\omega$ with $h_L$, $h_S$, and $h_F$ 
\begin{equation}\label{eq:apd_p0p1w}
    \begin{split}
        p_0^2 &= \left(\frac{h_L C-1}{C-1}\right)^2\\
        \mathbb{E}_{\epsilon}[p_1^2] &= \left(\frac{1-h_L}{C-1}\right)^2+\frac{(1-h_S)^2}{C-1}\\
        \omega &= \frac{h_F}{\rho(\b{A})}
    \end{split}
\end{equation}

So we have

\begin{equation}\label{eq:apd_obj_mlp_h_same}
    \begin{split}
        \frac{(1-\omega^2(C\mathbb{E}_{\epsilon}[p_1^2]+p_0^2))}{(1-\omega p_0)^2}
        &= \frac{\left[1-(\frac{h_F}{\rho(\b{A})})^2(C(\frac{1-h_L}{C-1})^2+C\frac{(1-h_S)^2}{C-1}+(\frac{h_L C-1}{C-1})^2)\right]}{\left[1- (\frac{h_F}{\rho(\b{A})}) (\frac{h_L C-1}{C-1})\right]^2}\\
    \end{split}
\end{equation}

Then, we can rewrite the objective as
\begin{equation}
    \begin{split}
        \mathcal{J}^{\neg\mathcal{G}} = (1+\mathcal{J}_{\b{N}}\mathcal{J}_h^{\neg\mathcal{G}})^{-1}
    \end{split}
\end{equation}

where
\begin{equation}
    \begin{split}
        \mathcal{J}_{\b{N}} &= \frac{\sum_{Y_u\neq Y_v}[2C(C-1)]^{-1}\norm{\b{\mu}_{Y_u,:}-\b{\mu}_{Y_u,:}}^2}{C^{-1}\abs{\b{\sigma}^2}}, \\
        \mathcal{J}_h^{\neg\mathcal{G}}
        &=\frac{\left[1-(\frac{h_F}{\rho(\b{A})})^2(C(\frac{1-h_L}{C-1})^2+C\frac{(1-h_S)^2}{C-1}+(\frac{h_L C-1}{C-1})^2)\right]}{\left[1- (\frac{h_F}{\rho(\b{A})}) (\frac{h_L C-1}{C-1})\right]^2}
    \end{split}
\end{equation}

\subsubsection{Graph-aware Models}
For graph-aware models, we have $\b{H}_u = \frac{1}{D_u}\sum_{v\in\mathcal{N}(u)} \b{X}_v$, where $\b{X}$ is the structural-aware features as in Section \ref{apd:proof_mlp_obj}. For the representation $m$ in node $u$, we have

\begin{equation}
    \begin{split}
        \b{H}_{u,m} &\sim \left[\frac{\b{S}}{\b{I}_C-\omega\b{S}}\b{N}_{:,m}\right]_{Y_u}\\
        &=\omega^{-1}\left[\frac{\omega\b{S}-\b{I}_C+\b{I}_C}{\b{I}_C-\omega\b{S}}\b{N}_{:,m}\right]_{Y_u}\\
        &= \omega^{-1}\left[(\b{I}_C-\omega\b{S})^{-1}\b{N}_{:,m}-\b{N}_{:,m}\right]_{Y_u}\\
    \end{split}
\end{equation}
We take the result of $(\b{I}_C-\omega\b{S})^{-1}\b{N}_{:,m}$ as shown in Eq. (\ref{eq:apd_result_mlp_feat_dist}). Then we have
\begin{equation}
    \begin{split}
    \resizebox{1\hsize}{!}{$
        \b{H}_{u,m} \sim \left[((1-\omega p_0)^{-1}-1)\omega^{-1}\b{I}_C\b{N}_{:,m}+\left[(1-\omega(Cp_1+p_0))^{-1}-(1-\omega p_0)^{-1}\right]C^{-1}\omega^{-1}\b{1}_C\b{N}_{:,m}\right]_{Y_u} 
    $}
    \end{split}
\end{equation}

Then we have the distribution of the representation of node $u$
\begin{equation}
    \begin{split}
        H_{u,m}\sim N(\hat{\mu}_{u,m},\hat{\sigma}^2_{u,m})
    \end{split}
\end{equation}
where
\begin{equation}
    \begin{split}
        \hat{\mu}_{Y_u,m} &= \omega^{-1}(1-\omega p_0)^{-1}-1)\mu_{Y_u,m}+C^{-1}\omega^{-1}\left[(1-\omega(Cp_1+p_0))^{-1}-(1-\omega p_0)^{-1}\right]\abs{\b{\mu}_{:,m}} \\
        \hat{\sigma}^2_{Y_u,m} &= \omega^{-2}(1-\omega^2p_0^2)^{-1}-1){\sigma}^2_{Y_u,m}+C^{-1}\omega^{-2}\left[(1-\omega^2(Cp_1^2+p_0^2))^{-1}-(1-\omega^2 p_0^2)^{-1}\right]\abs{\b{{\sigma}^2}_{:,m}} \\
    \end{split}
\end{equation}

When two nodes $u$ and $v$ share same labels, \ie{$Y_u=Y_v$}, we have
\begin{equation}
    H_{u,m}-H_{v,m}\sim N(0, \hat{\sigma}^2_{Y_u,m}+ \hat{\sigma}^2_{Y_v,m})
\end{equation}

Then the expectation of intra-class distance of $\b{H}$ can be expressed as
\begin{equation}
    \begin{split}
        \mathbb{E}_{Y_u=Y_v,\epsilon}\left[\norm{\b{H}_u-\b{H}_v}^2\right]&=\mathbb{E}_{Y_u=Y_v,\epsilon}\left[\sum_{m=0}^M\norm{{H}_{u,m}-{H}_{v,m}}^2\right] \\
        &=2\omega^{-2}C^{-1}[(1-\omega^2(C\mathbb{E}[p_1^2]+p_0^2))^{-1}-1]\abs{\b{\sigma^2}_{:,m}}
    \end{split}
\end{equation}

When two nodes $u$ and $v$ share different labels, \ie{$Y_u\neq Y_v$}, we have
\begin{equation}
    H_{u,m}-H_{v,m}\sim N(\hat{\mu}_{Y_u,m}-\hat{\mu}_{Y_v,m}, \hat{\sigma}^2_{Y_u,m}+ \hat{\sigma}^2_{Y_v,m})
\end{equation}

Then the expectation of inter-class distance of $\b{H}$ can be expressed as
\begin{equation}
    \begin{split}
        \mathbb{E}_{Y_u\neq Y_v,\epsilon}\left[\norm{\b{H}_u-\b{H}_v}^2\right]&=\mathbb{E}_{Y_u\neq Y_v,\epsilon}\left[\sum_{m=0}^M\norm{{H}_{u,m}-{H}_{v,m}}^2\right] \\
        &=2\omega^{-2}C^{-1}[(1-\omega^2(C\mathbb{E}[p_1^2]+p_0^2))^{-1}-1]\abs{\b{\sigma^2}_{:,m}} \\
        &+\omega^{-2}[C(C-1)]^{-1}[(1-\omega p_0)^{-1}-1]^2\sum_{Y_u\neq Y_v}\norm{\b{\mu}_{Y_u,:}-\b{\mu}_{Y_v,:}}^2
    \end{split}
\end{equation}

Then, we come back to the objective
\begin{equation}
    \begin{split}
        \mathcal{J} &= \frac{\mathbb{E}_{Y_u= Y_v,\epsilon}\left[\norm{\b{H}_u-\b{H}_v}^2\right]}{\mathbb{E}_{Y_u\neq Y_v,\epsilon}\left[\norm{\b{H}_u-\b{H}_v}^2\right]}\\
        &=\left( 1+\frac{\sum_{Y_u\neq Y_v}[2C(C-1)]^{-1}\norm{\b{\mu}_{Y_u,:}-\b{\mu}_{Y_u,:}}^2}{C^{-1}\abs{\b{\sigma}^2}} \frac{[(1-\omega p_0)^{-1}-1]^2}{[(1-\omega^2(C\mathbb{E}[p_1^2]+p_0^2))^{-1}-1]} \right)^{-1}\\
    \end{split}
\end{equation}

Let's consider the term

\begin{equation}
    \begin{split}
        \frac{[(1-\omega p_0)^{-1}-1]^2}{[(1-\omega^2(C\mathbb{E}[p_1^2]+p_0^2))^{-1}-1]}=\frac{p_0^2}{(C\mathbb{E}[p_1^2]+p_0^2)}\frac{(1-\omega^2(C\mathbb{E}[p_1^2]+p_0^2))}{(1-\omega p_0)^2}\\
    \end{split}
\end{equation}

We notice the second term $\frac{(1-\omega^2(C\mathbb{E}[p_1^2]+p_0^2))}{(1-\omega p_0)^2}=\mathcal{J}_h^{\neg\mathcal{G}}$, thereby we simplify the expression as
\begin{equation}
    \begin{split}
        \frac{[(1-\omega p_0)^{-1}-1]^2}{[(1-\omega^2(C\mathbb{E}[p_1^2]+p_0^2))^{-1}-1]}
        &=\frac{p_0^2}{(C\mathbb{E}[p_1^2]+p_0^2)}\mathcal{J}_h^{\neg\mathcal{G}}\\
    \end{split}
\end{equation}

Then we take back $p_0$, $p_1$, and $\omega$ with $h_L$, $h_S$, and $h_F$ as in Eq. (\ref{eq:apd_p0p1w})

\begin{equation}
    \begin{split}
        \frac{[(1-\omega p_0)^{-1}-1]^2}{[(1-\omega^2(C\mathbb{E}[p_1^2]+p_0^2))^{-1}-1]}
        &=\frac{(\frac{h_L C-1}{C-1})^2}{(C(\frac{1-h_L}{C-1})^2+C\frac{(1-h_S)^2}{C-1}+(\frac{h_L C-1}{C-1})^2)}\mathcal{J}_h^{\neg\mathcal{G}}\\
    \end{split}
\end{equation}

Then, we can rewrite the objective as
\begin{equation}
    \begin{split}
        \mathcal{J}^{\mathcal{G}} = (1+\mathcal{J}_{\b{N}}\mathcal{J}_h^{\mathcal{G}})^{-1}
    \end{split}
\end{equation}

where
\begin{equation}
    \begin{split}
        \mathcal{J}_{\b{N}} &= \frac{\sum_{Y_u\neq Y_v}[2C(C-1)]^{-1}\norm{\b{\mu}_{Y_u,:}-\b{\mu}_{Y_u,:}}^2}{C^{-1}\abs{\b{\sigma}^2}}, \\
        \mathcal{J}_h^{\mathcal{G}}
        &=\frac{(\frac{h_L C-1}{C-1})^2}{(C(\frac{1-h_L}{C-1})^2+C\frac{(1-h_S)^2}{C-1}+(\frac{h_L C-1}{C-1})^2)}\mathcal{J}_h^{\neg\mathcal{G}}\\
    \end{split}
\end{equation}

\subsection{Proof of Theorem 3.1}\label{apd:proof_theorem31}
\begin{theorem} {3.1}
    The partial derivative of ${\mathcal{J}}_h^{\neg\mathcal{G}}$ with respect to label homophily $h_L$ satisfies,
    \begin{equation}
        \begin{split}
        \frac{\partial\mathcal{J}_h^{\neg\mathcal{G}}}{\partial h_L}
        \begin{cases}
        <0, & \text{if}\ h_F\in(-1,0)\\
        \geq 0, & \text{if}\ h_F\in[0,1)
        \end{cases}
        \end{split}
    \end{equation}
\end{theorem}
\textit{Proof.}
For the graph-agnostic models, we take the partial derivative of $\mathcal{J}_h^{\neg\mathcal{G}}$ with respect to $h_L$
\begin{equation}
    \begin{split}
        \frac{\partial\mathcal{J}_h^{\neg\mathcal{G}}}{\partial h_L}
        &=\frac{-h_F^2\Bigr(2C(1-h_L)(-1)+2C(h_L C-1)\Bigr) \mathcal{T}_{h,\text{DOWN}}^2 - (-h_F C)
 2\mathcal{T}_{h,\text{DOWN}}\mathcal{T}_{h,\text{UP}}}{\mathcal{T}_{h,\text{DOWN}}^4}\\
    \end{split}
\end{equation}

Take the approximations of $\mathcal{T}_{h,\text{UP}}$ and $\mathcal{T}_{h,\text{DOWN}}$ are given in Eq. (\ref{eq:apd_obj_up_down}) back to $\frac{\partial\mathcal{J}_h^{\neg\mathcal{G}}}{\partial h_L}$, we have
\begin{equation}
    \begin{split}
        \frac{\partial\mathcal{J}_h^{\neg\mathcal{G}}}{\partial h_L}
        &\approx\frac{-2Ch_F^2[(1-h_L)(-1)+(h_L C-1)] + 2Ch_F [\rho(\b{A})(C-1)]}{[\rho(\b{A})(C-1)]^2}\\
        &\approx\frac{2C}{[\rho(\b{A})(C-1)]}h_F\\
    \end{split}
\end{equation}

It is easy to see $\frac{2C}{[\rho(\b{A})(C-1)]}>0$, so the sign of $\frac{\partial\mathcal{J}_h^{\neg\mathcal{G}}}{\partial h_L}$ is solely depend on $h_F$. Therefore, we have 
\begin{equation}
    \begin{split}
    \frac{\partial\mathcal{J}_h^{\neg\mathcal{G}}}{\partial h_L}
    \begin{cases}
    <0, & \text{if}\ h_F\in(-1,0)\\
    \geq 0, & \text{if}\ h_F\in[0,1)
    \end{cases}
    \end{split}
\end{equation}

\subsection{Proof of Theorem 3.2}\label{apd:proof_theorem32}
\begin{theorem} {3.2}
    The partial derivative of ${\mathcal{J}}_h^{\neg\mathcal{G}}$ with respect to structural homophily $h_S$ satisfies,
    \begin{equation}
        \begin{split}
            \frac{\partial\mathcal{J}_h^{\neg\mathcal{G}}}{\partial h_S}\ge0
        \end{split}
    \end{equation}
\end{theorem}
\textit{Proof.}
We take the partial derivative of $\mathcal{J}_h^{\neg\mathcal{G}}$ with respect to $h_S$

\begin{equation}
    \begin{split}
        \frac{\partial\mathcal{J}_h^{\neg\mathcal{G}}}{\partial h_S}
        &=\frac{\frac{2C}{\rho^2(\b{A})(C-1)}}{\left[1- (\frac{h_F}{\rho(\b{A})}) (\frac{h_L C-1}{C-1})\right]^2}h_F^2(1-h_S)\\
    \end{split}
\end{equation}

From Eq. (\ref{eq:apd_aprrox_rho_h}), we have

\begin{equation}
    \frac{\frac{2C}{\rho^2(\b{A})(C-1)}}{\left[1- (\frac{h_F}{\rho(\b{A})}) (\frac{h_L C-1}{C-1})\right]^2}>0,\; h_F^2\ge 0,\; \text{and} \; (1-h_S)\ge 0
\end{equation}

Therefore, we can approximate $\frac{\partial\mathcal{J}_h^{\neg\mathcal{G}}}{\partial h_S}$ that satisfies

\begin{equation}
    \begin{split}
        \frac{\partial\mathcal{J}_h^{\neg\mathcal{G}}}{\partial h_S}\ge 0
    \end{split}
\end{equation}
\subsection{Proof of Theorem 3.3}\label{apd:proof_theorem33}
\begin{theorem} {3.3}
    The partial derivative of ${\mathcal{J}}_h^{\neg\mathcal{G}}$ with respect to feature homophily $h_F$ satisfies,
    \begin{equation}
        \begin{split}
        \frac{\partial\mathcal{J}_h^{\neg\mathcal{G}}}{\partial h_F}
            \begin{cases} 
            < 0, & \text{if}\ h_L\in(0 ,h_L^-); h_L \in( h_L^-,h_L^+)\ \text{and} \ h_F \in (\hat{h}_F,1)\\
            > 0, & \text{if}\ h_L\in(h_L^+,1]; h_L\in(h_L^-,h_L^+)\ \text{and}\ h_F\in(-1,\hat{h}_F) \\
            = 0, & \text{if}\ h_L\in(h_L^-,h_L^+)\ \text{and}\ h_F=\hat{h}_F
            \end{cases}
        \end{split}
    \end{equation}
    where $0<h_L^- < h_L^+<1$ and $-1<\hat{h}_F<1$.
\end{theorem}
\textit{Proof.}
We take the partial derivative of $\mathcal{J}_h^{\neg\mathcal{G}}$ with respect to $h_F$
\begin{equation}
    \begin{split}
    \resizebox{1\hsize}{!}{$
        \frac{\partial\mathcal{J}_h^{\neg\mathcal{G}}}{\partial h_F}
        =\frac{-2h_F\Bigr( C(1-h_L)^2+C(C-1)(1-h_S)^2 +(h_L C-1)^2 \Bigr)\mathcal{T}_{h,\text{DOWN}}^2+2(h_L C -1)\mathcal{T}_{h,\text{DOWN}}\mathcal{T}_{h,\text{UP}}}{\mathcal{T}_{h,\text{DOWN}}^4}
    $}
    \end{split}
\end{equation}

where
\begin{equation}\label{eq:apd_mlp_obj_h_f}
    \begin{split}
        \mathcal{T}_{h,\text{UP}}&=\Bigr[ \rho(\b{A})^2(C-1)^2-h_F^2\Bigr( C(1-h_L)^2+C(C-1)(1-h_S)^2 +(h_L C-1)^2 \Bigr) \Bigr] \\
        \mathcal{T}_{h,\text{DOWN}}&=\Bigr[\rho(\b{A})(C-1) - h_F(h_L C-1) \Bigr]
    \end{split}
\end{equation}

Next, we approximate $\mathcal{T}_{h,\text{UP}}$ and $\mathcal{T}_{h,\text{DOWN}}$ with Eq. (\ref{eq:apd_aprrox_rho_h})
\begin{equation}\label{eq:apd_obj_up_down}
    \begin{split}
        \mathcal{T}_{h,\text{UP}}&\approx[\rho(\b{A})(C-1)]^2 \\
        \mathcal{T}_{h,\text{DOWN}}&\approx\rho(\b{A})(C-1)
    \end{split}
\end{equation}

Then take back to Eq. (\ref{eq:apd_mlp_obj_h_f})

\begin{equation}\label{eq:apd_hf_hfstar}
    \begin{split}
        \frac{\partial\mathcal{J}_h^{\neg\mathcal{G}}}{\partial h_F}
        &\approx\frac{-2h_F\Bigr( C(\frac{1-h_L}{C-1})^2+\frac{C}{C-1}(1-h_S)^2 +(\frac{h_L C-1}{C-1})^2 \Bigr)+2\frac{h_L C -1}{C-1}\rho(\b{A})}{\rho^2(\b{A})}\\
        &\approx\frac{2(\hat{h}_F-h_F)}{\Bigr( C(\frac{1-h_L}{C-1})^2+\frac{C}{C-1}(1-h_S)^2 +(\frac{h_L C-1}{C-1})^2 \Bigr)^{-1}\rho^2(\b{A})}
    \end{split}
\end{equation}
where
\begin{equation}
    \hat{h}_F=\frac{\frac{h_L C -1}{C-1}\rho(\b{A})}{C(\frac{1-h_L}{C-1})^2+\frac{C}{C-1}(1-h_S)^2 +(\frac{h_L C-1}{C-1})^2}
\end{equation}

As a result, the sign of $\frac{\partial\mathcal{J}_h^{\neg\mathcal{G}}}{\partial h_F}$ satisfies the following conditions:
\begin{equation}
    \begin{split}
        \frac{\partial\mathcal{J}_h^{\neg\mathcal{G}}}{\partial h_F}    \begin{cases}
          >0,  & \text{if}\ h_F < \hat{h}_F \\
          <0,  & \text{if}\ h_F > \hat{h}_F \\
          =0,  & \text{if}\ h_F = \hat{h}_F
        \end{cases}
    \end{split}
\end{equation}

Then we discuss the influence of $h_L$ to $\frac{\partial\mathcal{J}_h^{\neg\mathcal{G}}}{\partial h_F}$. If $\hat{h}_F\ge 1$, then $\frac{\partial\mathcal{J}_h^{\neg\mathcal{G}}}{\partial h_F}>0$ holds $\forall h_F\in(-1,1)$. In this case, we have
\begin{equation}
    -\Bigr( C(\frac{1-h_L}{C-1})^2+\frac{C}{C-1}(1-h_S)^2 +(\frac{h_L C-1}{C-1})^2 \Bigr)+\frac{h_L C -1}{C-1}\rho(\b{A})>0
\end{equation}
The solution of this condition for $h_L\in[0,1]$ is $h_L>h_L^+$, where
\begin{equation}
    \begin{split}
    \resizebox{1\hsize}{!}{$
        h_L^+=\frac{4C+C(C-1)\rho(\b{A})-\sqrt{[4C+C(C-1)\rho(\b{A})]^2-4C(C+1)[C+1+(C-1)\rho(\b{A})+C(C-1)(1-h_S)^2]}}{2C(C+1)}
    $}
    \end{split}
\end{equation}

Conversely, If $\hat{h}_F\le -1$, then $\frac{\partial\mathcal{J}_h^{\neg\mathcal{G}}}{\partial h_F}<0$ holds $\forall h_F\in(-1,1)$. In this case, we have
\begin{equation}
    \Bigr( C(\frac{1-h_L}{C-1})^2+\frac{C}{C-1}(1-h_S)^2 +(\frac{h_L C-1}{C-1})^2 \Bigr)+\frac{h_L C -1}{C-1}\rho(\b{A})>0
\end{equation}
The solution of this condition for $h_L\in[0,1]$ is $h_L<h_L^-$, where
\begin{equation}
    \begin{split}
    \resizebox{1\hsize}{!}{$
        h_L^-=\frac{4C-C(C-1)\rho(\b{A})+\sqrt{[4C-C(C-1)\rho(\b{A})]^2-4C(C+1)[C+1-(C-1)\rho(\b{A})+C(C-1)(1-h_S)^2]}}{2C(C+1)}
    $}
    \end{split}
\end{equation}

It is easy to show that $0<h_L^-<h_L^+<1$. So far, we already know that $\frac{\partial\mathcal{J}_h^{\neg\mathcal{G}}}{\partial h_F}>0$ when  $h_L>h_L^+$ and $\frac{\partial\mathcal{J}_h^{\neg\mathcal{G}}}{\partial h_F}<0$ when  $h_L<h_L^-$ for all $h_F\in(-1,1)$. Then, for $h_L^-<h_L<h_L^+$, the sign of $\frac{\partial\mathcal{J}_h^{\neg\mathcal{G}}}{\partial h_F}$ is dependent on $h_F$ as shown in Eq. (\ref{eq:apd_hf_hfstar}). In conclusion, we have
\begin{equation}
    \begin{split}
    \frac{\partial\mathcal{J}_h^{\neg\mathcal{G}}}{\partial h_F}
        \begin{cases} 
        < 0, & \text{if}\ h_L\in(0 ,h_L^-); h_L \in( h_L^-,h_L^+)\ \text{and} \ h_F \in (\hat{h}_F,1)\\
        > 0, & \text{if}\ h_L\in(h_L^+,1]; h_L\in(h_L^-,h_L^+)\ \text{and}\ h_F\in(-1,\hat{h}_F) \\
        = 0, & \text{if}\ h_L\in(h_L^-,h_L^+)\ \text{and}\ h_F=\hat{h}_F
        \end{cases}
    \end{split}
\end{equation}

\subsection{Proof of Theorem 2.1}\label{apd:proof_theorem21}
\begin{theorem} {2.1}
    The partial derivative of ${\mathcal{J}}_h^{\mathcal{G}}$ with respect to label homophily $h_L$ satisfies,
    \begin{equation}
        \begin{split}
        \frac{\partial\mathcal{J}_h^{\mathcal{G}}}{\partial h_L}
        \begin{cases}
        <0, & \text{if}\ h_L\in[0,\frac{1}{C})\\
        \geq 0, & \text{if}\ h_L\in[\frac{1}{C},1]
        \end{cases}
        \end{split}
    \end{equation}
\end{theorem}
\textit{Proof.}
For the graph-aware models, we approximate the $\mathcal{J}_h^{\mathcal{G}}$ with Eq. (\ref{eq:apd_aprrox_rho_h})
\begin{equation}
    \begin{split}
        \mathcal{J}_h^{\mathcal{G}}
        &=\frac{(\frac{h_L C-1}{C-1})^2\left[1-(\frac{h_F}{\rho(\b{A})})^2(C(\frac{1-h_L}{C-1})^2+C\frac{(1-h_S)^2}{C-1}+(\frac{h_L C-1}{C-1})^2)\right]}{(C(\frac{1-h_L}{C-1})^2+C\frac{(1-h_S)^2}{C-1}+(\frac{h_L C-1}{C-1})^2)\left[1- (\frac{h_F}{\rho(\b{A})}) (\frac{h_L C-1}{C-1})\right]^2}\\
        &\approx\frac{(\frac{h_L C-1}{C-1})^2}{(C(\frac{1-h_L}{C-1})^2+C\frac{(1-h_S)^2}{C-1}+(\frac{h_L C-1}{C-1})^2)}
    \end{split}
\end{equation}

Then we take the partial derivative with respect to $h_L$
\begin{equation}
    \begin{split}
        \frac{\partial\mathcal{J}_h^{\mathcal{G}}}{\partial h_L}
        &=(h_L C-1)\Bigr[ \frac{2C^2(C-1)(C+1)h_L^2}{(C-1)^4}+\frac{8C^2(C-1)+2C(3C+1)}{(C-1)^4}h_L \Bigr. \\
        &\Bigr. + \frac{2C(C+1)(C-1)-4C+2C^2(C-1)^2(1-h_S)^2}{(C-1)^4} \Bigr]
    \end{split}
\end{equation}

We can see that the first term monotonically increases with respect to $h_L$ and the second term is always positive for $h_L\in[0,1]$. So we have
\begin{equation}
    \begin{split}
    \frac{\partial\mathcal{J}_h^{\mathcal{G}}}{\partial h_L}
    \begin{cases}
    <0, & \text{if}\ h_L\in[0,\frac{1}{C})\\
    \geq 0, & \text{if}\ h_L\in[\frac{1}{C},1]
    \end{cases}
    \end{split}
\end{equation}
\subsection{Proof of Theorem 2.2}\label{apd:proof_theorem22}
\begin{theorem} {2.2}
    The partial derivative of ${\mathcal{J}}_h^{\mathcal{G}}$ with respect to structural homophily $h_S$ satisfies,
    \begin{equation}
        \begin{split}
            \frac{\partial\mathcal{J}_h^{\mathcal{G}}}{\partial h_S}\ge0
        \end{split}
    \end{equation}
\end{theorem}
\textit{Proof.}
We take the partial derivative of $\mathcal{J}_h^{\mathcal{G}}$ with respect to $h_F$

\begin{equation}
    \begin{split}
        \frac{\partial\mathcal{J}_h^{\mathcal{G}}}{\partial h_S}
        &=\frac{(\frac{h_L C-1}{C-1})^2\frac{2C}{C-1}(1-h_S)}{(C(\frac{1-h_L}{C-1})^2+C\frac{(1-h_S)^2}{C-1}+(\frac{h_L C-1}{C-1})^2)^2}\mathcal{J}_h^{\neg\mathcal{G}}\\
        &+\frac{(\frac{h_L C-1}{C-1})^2}{(C(\frac{1-h_L}{C-1})^2+C\frac{(1-h_S)^2}{C-1}+(\frac{h_L C-1}{C-1})^2)}\frac{\partial\mathcal{J}_h^{\neg\mathcal{G}}}{\partial h_S}
    \end{split}
\end{equation}

From Eq. (\ref{eq:apd_aprrox_rho_h}), we have

\begin{equation}
    \frac{(\frac{h_L C-1}{C-1})^2\frac{2C}{C-1}(1-h_S)}{(C(\frac{1-h_L}{C-1})^2+C\frac{(1-h_S)^2}{C-1}+(\frac{h_L C-1}{C-1})^2)^2}\ge0 \; \text{and} \;
    \frac{(\frac{h_L C-1}{C-1})^2}{(C(\frac{1-h_L}{C-1})^2+C\frac{(1-h_S)^2}{C-1}+(\frac{h_L C-1}{C-1})^2)}\ge 0
\end{equation}

Together with $\mathcal{J}_h^{\neg\mathcal{G}}>0$ and $\frac{\partial\mathcal{J}_h^{\neg\mathcal{G}}}{\partial h_S}\ge 0$, we can approximate $\frac{\partial\mathcal{J}_h^{\mathcal{G}}}{\partial h_S}$ as
\begin{equation}
    \begin{split}
        \frac{\partial\mathcal{J}_h^{\mathcal{G}}}{\partial h_S}\ge 0
    \end{split}
\end{equation}

\subsection{Proof of Theorem 2.3}\label{apd:proof_theorem23}
\begin{theorem} {2.3}
    The partial derivative of ${\mathcal{J}}_h^{\mathcal{G}}$ with respect to feature homophily $h_F$ satisfies,
    \begin{equation}
        \begin{split}
            \begin{cases} 
            \frac{\partial\mathcal{J}_h^{\mathcal{G}}}{\partial h_F}< 0, & \text{if}\ h_L\in(0 ,h_L^-); h_L \in( h_L^-,h_L^+)\ \text{and} \ h_F \in (\hat{h}_F,1)\\
            \frac{\partial\mathcal{J}_h^{\mathcal{G}}}{\partial h_F}> 0, & \text{if}\ h_L\in(h_L^+,1]; h_L\in(h_L^-,h_L^+)\ \text{and}\ h_F\in(-1,\hat{h}_F) \\
            \frac{\partial\mathcal{J}_h^{\mathcal{G}}}{\partial h_F} = 0, & \text{if}\ h_L = \frac{1}{C}; h_L\in(h_L^-,h_L^+)\ \text{and}\ h_F=\hat{h}_F
            \end{cases}
        \end{split}
    \end{equation}
    where $0<h_L^- < h_L^+<1$ and $-1<\hat{h}_F<1$. The expressions and detailed calculation of $h_L^-$, $h_L^+$, and $\hat{h}_F$ are shown in Appendix \ref{apd:proof_theorem33}.
\end{theorem}
\textit{Proof.}
For the graph-aware models, we take the partial derivative of $\mathcal{J}_h^{\mathcal{G}}$ with respect to $h_F$
\begin{equation}
    \begin{split}
        \frac{\partial\mathcal{J}_h^{\mathcal{G}}}{\partial h_F}
        &=\frac{(\frac{h_L C-1}{C-1})^2}{(C(\frac{1-h_L}{C-1})^2+C\frac{(1-h_S)^2}{C-1}+(\frac{h_L C-1}{C-1})^2)}\frac{\partial\mathcal{J}_h^{\neg\mathcal{G}}}{\partial h_F}\\
    \end{split}
\end{equation}
Since the term $\frac{(\frac{h_L C-1}{C-1})^2}{(C(\frac{1-h_L}{C-1})^2+C\frac{(1-h_S)^2}{C-1}+(\frac{h_L C-1}{C-1})^2)}\ge 0$, the sign of $\frac{\partial\mathcal{J}_h^{\mathcal{G}}}{\partial h_F}$ is the determined by $\frac{\partial\mathcal{J}_h^{\neg\mathcal{G}}}{\partial h_F}$ except for $h_L=\frac{1}{C}$, which leads to the similar results as graph-aware models
\begin{equation}
    \begin{split}
        \begin{cases} 
        \frac{\partial\mathcal{J}_h^{\mathcal{G}}}{\partial h_F}< 0, & \text{if}\ h_L\in(0 ,h_L^-); h_L \in( h_L^-,h_L^+)\ \text{and} \ h_F \in (\hat{h}_F,1)\\
        \frac{\partial\mathcal{J}_h^{\mathcal{G}}}{\partial h_F}> 0, & \text{if}\ h_L\in(h_L^+,1]; h_L\in(h_L^-,h_L^+)\ \text{and}\ h_F\in(-1,\hat{h}_F) \\
        \frac{\partial\mathcal{J}_h^{\mathcal{G}}}{\partial h_F} = 0, & \text{if}\ h_L = \frac{1}{C}; h_L\in(h_L^-,h_L^+)\ \text{and}\ h_F=\hat{h}_F
        \end{cases}
    \end{split}
\end{equation}